\RequirePackage{fix-cm}
\documentclass[natbib,twocolumn]{svjour3}          %
\smartqed  %
\usepackage{graphicx}
\usepackage{xspace}
\usepackage{multirow}
\usepackage{rotating}
\usepackage{subfigure}
\usepackage{amsmath}
\usepackage{url}
\usepackage{amssymb}
\usepackage{epsfig}
\usepackage{rotating} %
\usepackage{color}
\usepackage{booktabs} %
\usepackage{tabularx} %
\usepackage{url}

\graphicspath{{./}}
\usepackage[english,american]{babel}

\newcommand{\redtext}[1]{\emph{\textcolor{red}{#1}}}

\newcommand{\myparagraph}[1]{ \paragraph{#1}} 

\newcommand{\invisible}[1]{}%

\graphicspath{{./fig/}{./fig/plots/}}

\newcommand{\MPIIMD}{MPII-MD\xspace}

\newcommand{\LSMDC}{LSMDC\xspace}
\newcommand{\LaScMoDeCh}{Large Scale Movie Description Challenge\xspace}

\newcommand{\MoViAnDa}{Montreal Video Annotation Dataset\xspace}
\newcommand{\MVAD}{M-VAD\xspace}

\newcommand{\ApproachVisualLabels}{Visual-Labels\xspace}
 \makeatletter
 \DeclareRobustCommand\onedot{\futurelet\@let@token\@onedot}
 \def\@onedot{\ifx\@let@token.\else.\null\fi\xspace}
 \def\eg{e.g\onedot} 
 \def\ie{i.e\onedot}

 \makeatother

\DeclareRobustCommand{\figref}[1]{Figure~\ref{#1}}

\DeclareRobustCommand{\Figref}[1]{Figure~\ref{#1}}

\DeclareRobustCommand{\Secref}[1]{Section~\ref{#1}}
\DeclareRobustCommand{\secref}[1]{Section~\ref{#1}}

\DeclareRobustCommand{\Tableref}[1]{Table~\ref{#1}}
\DeclareRobustCommand{\tableref}[1]{Table~\ref{#1}}

\newcommand{\nLSMDCMovies}{202\xspace}
\newcommand{\nLSMDCSentences}{118,114\xspace}
\newcommand{\nLSMDCClips}{118,081\xspace}

\newcommand{\nMovies}{94\xspace}
\newcommand{\nSentences}{68,375\xspace}
\newcommand{\nClips}{68,337\xspace}

\newcommand{\nMoviesScript}{50\xspace}
\newcommand{\nMoviesOnlyScript}{39\xspace}
\newcommand{\nMoviesOverlap}{11\xspace}

\journalname{IJCV}
\begin{document}

\title{Movie Description}

\subtitle{}

\author{Anna Rohrbach \and Atousa Torabi \and Marcus Rohrbach \and Niket Tandon \and Christopher Pal \and Hugo Larochelle \and Aaron Courville \and Bernt Schiele}

\authorrunning{A. Rohrbach, A. Torabi, M. Rohrbach, N. Tandon, C. Pal, H. Larochelle, A. Courville, B. Schiele} 

\institute{Anna Rohrbach \textsuperscript{1} \and Atousa Torabi\textsuperscript{3} \and Marcus Rohrbach\textsuperscript{2} \and Niket Tandon \textsuperscript{1} \and Christopher Pal\textsuperscript{4} \and Hugo Larochelle\textsuperscript{5,6
} \and Aaron Courville\textsuperscript{7} \and  Bernt Schiele\textsuperscript{1}\at
            \textsuperscript{1} Max Planck Institute for Informatics, Saarbr\"ucken, Germany\\
            \textsuperscript{2} ICSI and EECS, UC Berkeley, United States\\
            \textsuperscript{3} Disney Research, Pittsburgh, United States\\
            \textsuperscript{4} \'Ecole Polytechnique de Montr\'eal, Montr\'eal, Canada\\
            \textsuperscript{5} Universit\'{e} de Sherbrooke, Sherbrooke, Canada\\
            \textsuperscript{6} Twitter, Cambridge, United States\\
            \textsuperscript{7} Universit\'{e} de Montr\'{e}al, Montr\'eal, Canada\\
}

\date{}

\maketitle

\begin{abstract}
Audio Description (AD) provides linguistic descriptions of movies and allows visually impaired people to follow a movie along with their peers. Such descriptions are by design mainly visual and thus naturally form an interesting data source for computer vision and computational linguistics. In this work we propose a novel dataset which contains transcribed ADs, which are temporally aligned to full length movies. In addition we also collected and aligned movie scripts used in prior work and compare the two sources of descriptions. In total the \emph{Large Scale Movie Description Challenge} (LSMDC) contains a parallel corpus of \nLSMDCSentences sentences and video clips from \nLSMDCMovies movies. First we characterize the dataset by benchmarking different approaches for generating video descriptions. Comparing ADs to scripts, we find that ADs are indeed more visual and describe precisely what \emph{is shown} rather than what \emph{should happen} according to the scripts created prior to movie production.
Furthermore, we present and compare the results of several teams who participated in a challenge organized in the context of the workshop ``Describing and Understanding Video \& The Large Scale Movie Description Challenge (LSMDC)'', at ICCV 2015.
\end{abstract}

\section{Introduction}
\label{sec:intro}
\begin{figure}[t]
\scriptsize
\begin{center}
\begin{tabular}{@{}p{2.5cm}p{2.5cm}p{2.5cm}}
\includegraphics[width=\linewidth]{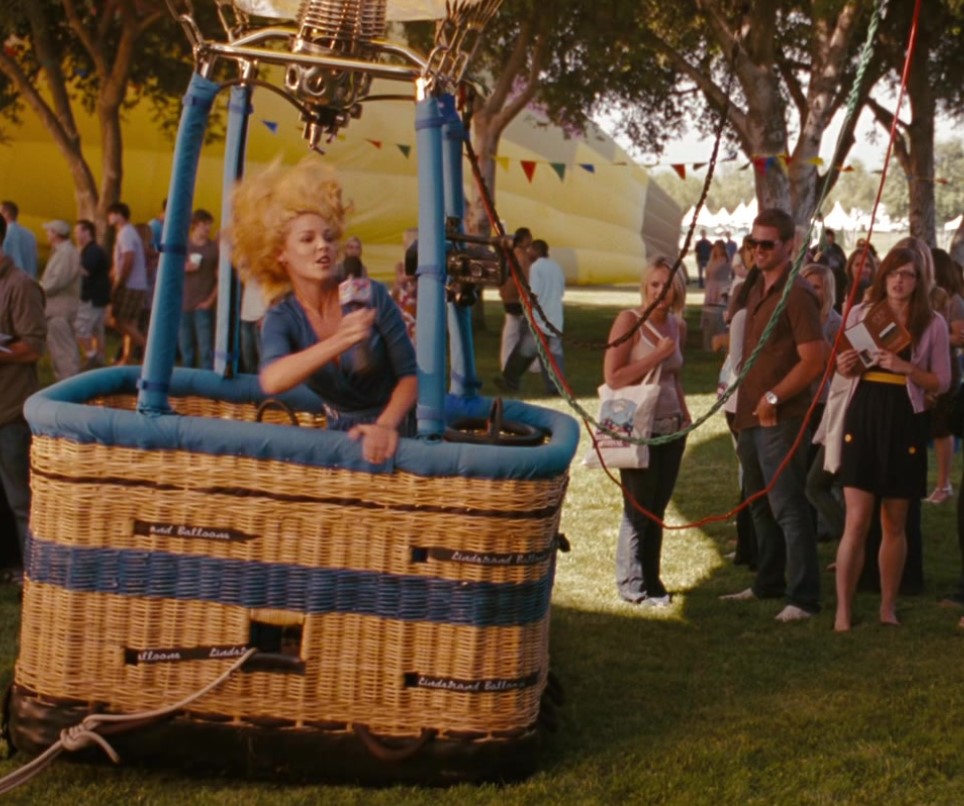} & \includegraphics[width=\linewidth]{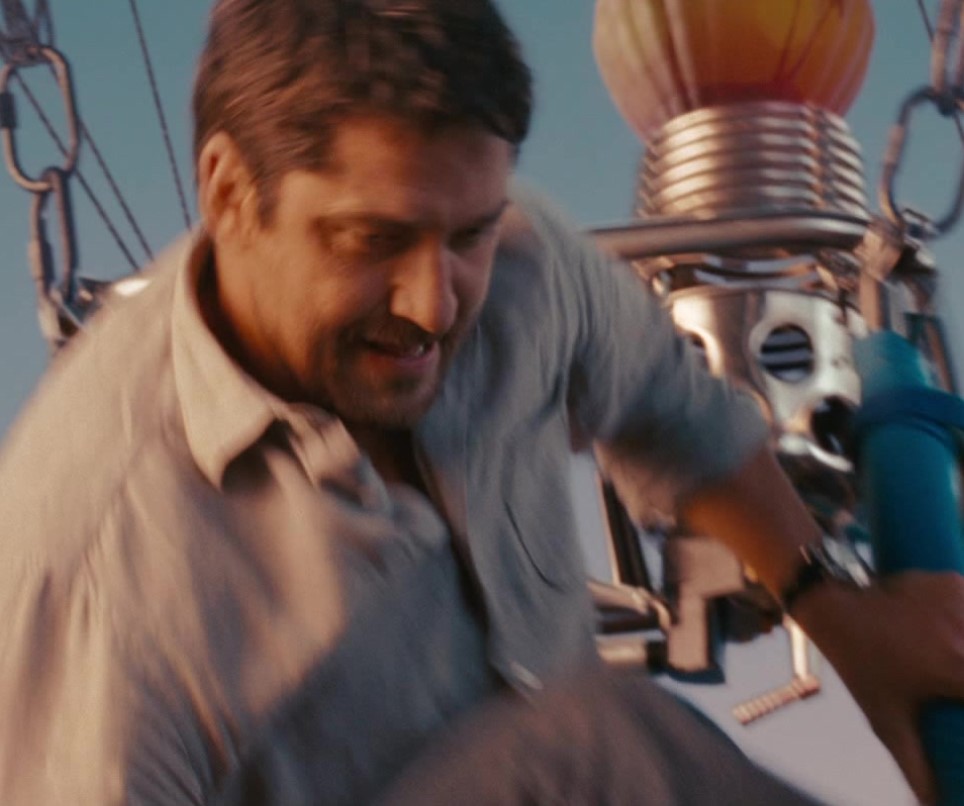} & \includegraphics[width=\linewidth]{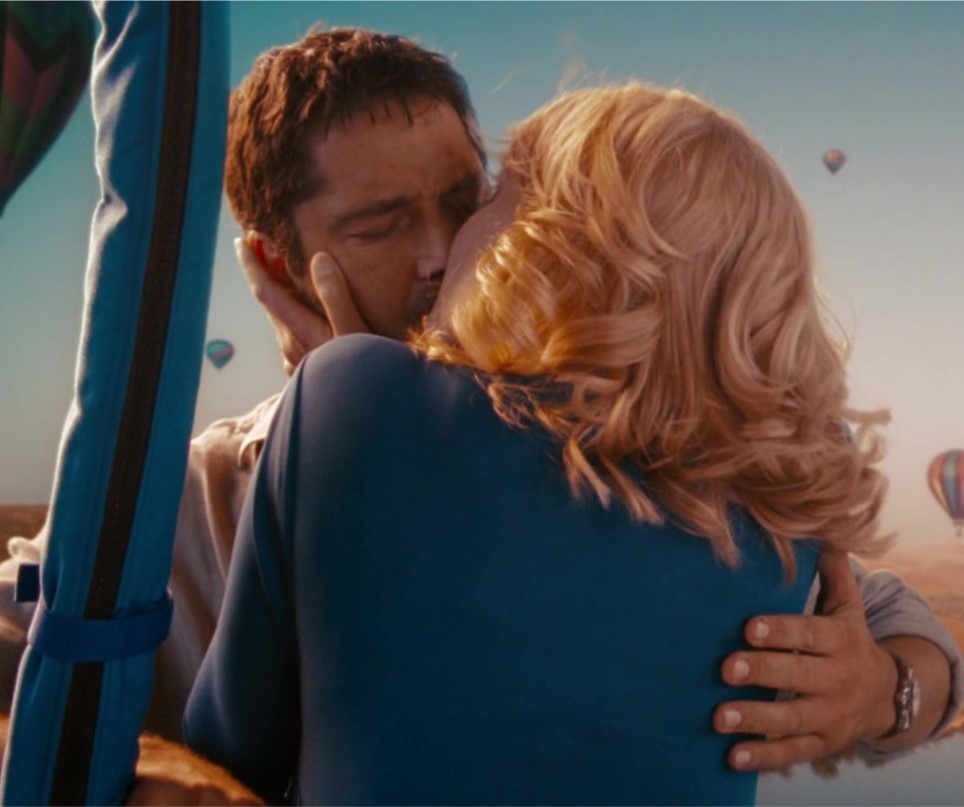} \\
\textbf{AD}: Abby gets in the basket. & Mike leans over and sees how high they are. & Abby clasps her hands around his face and kisses him passionately. \\
\textbf{Script}: After a moment a frazzled Abby pops up in his place. & Mike looks down to see -- they are now fifteen feet above the ground. & For the first time in her life, she stops thinking and grabs Mike and kisses the hell out of him. \\
\end{tabular}
\caption{Audio description (AD) and movie script samples from the movie ``Ugly Truth''.}
\label{fig:teaser1}
\end{center}
\end{figure}

Audio descriptions (ADs) make movies accessible to millions of blind or visually impaired people\footnote{\label{fn:blind} In this work we refer for simplicity to ``the blind'' to account for all blind and visually impaired people which benefit from AD, knowing of the variety of visually impaired and that AD is not accessible to all.}. AD --- sometimes also referred to as Descriptive Video Service (DVS) --- provides an audio narrative of the ``most important aspects of the visual information'' \citep{salway07corpus}, namely actions, gestures, scenes, and character appearance as can be seen in Figures \ref{fig:teaser1} and \ref{fig:teaser}. AD is prepared by trained describers and read by professional narrators. While more and more movies are audio transcribed,  it may take up to 60 person-hours to describe a 2-hour movie \citep{lakritz06tr}, resulting in the fact that today only a small subset of movies and TV programs are available for the blind. Consequently, automating this process has the potential to greatly increase accessibility to this media.

\renewcommand{\bottomfraction}{0.8}
\setcounter{dbltopnumber}{2}
\renewcommand{\textfraction}{0.07}
\newcommand{\colwidth}{3.1cm}
\begin{figure*}[t]
\scriptsize
\begin{center}
\begin{tabular}{@{}p{\colwidth}p{\colwidth}p{\colwidth}p{\colwidth}p{\colwidth}}
 \includegraphics[width=\linewidth]{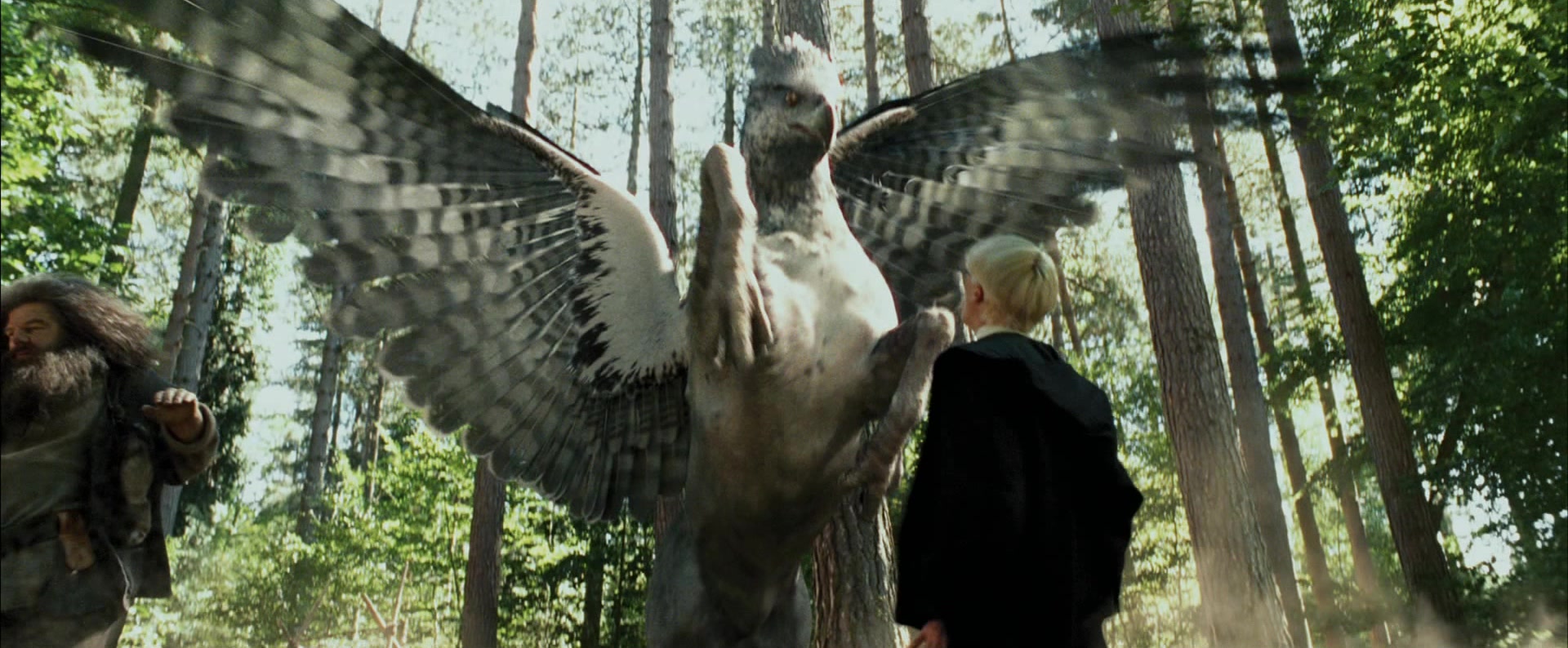} & \includegraphics[width=\linewidth]{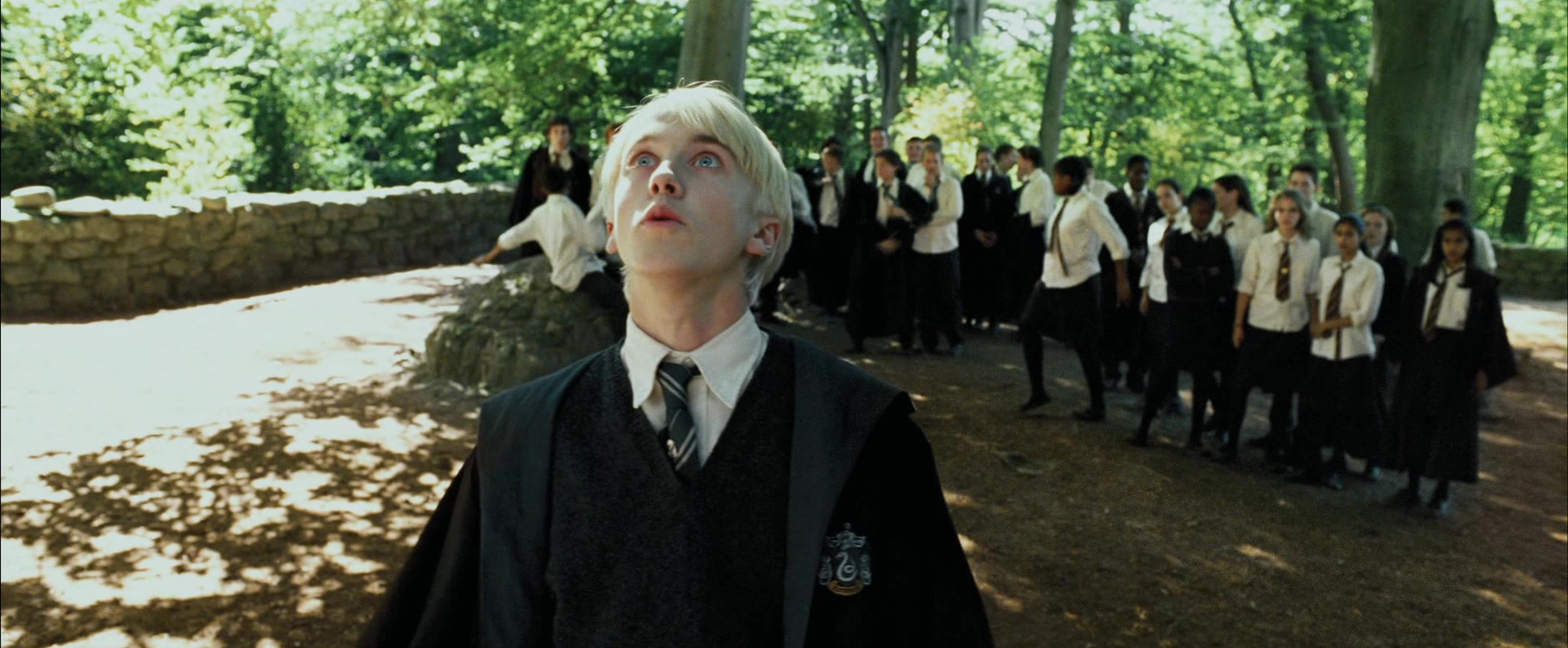} & \includegraphics[width=\linewidth]{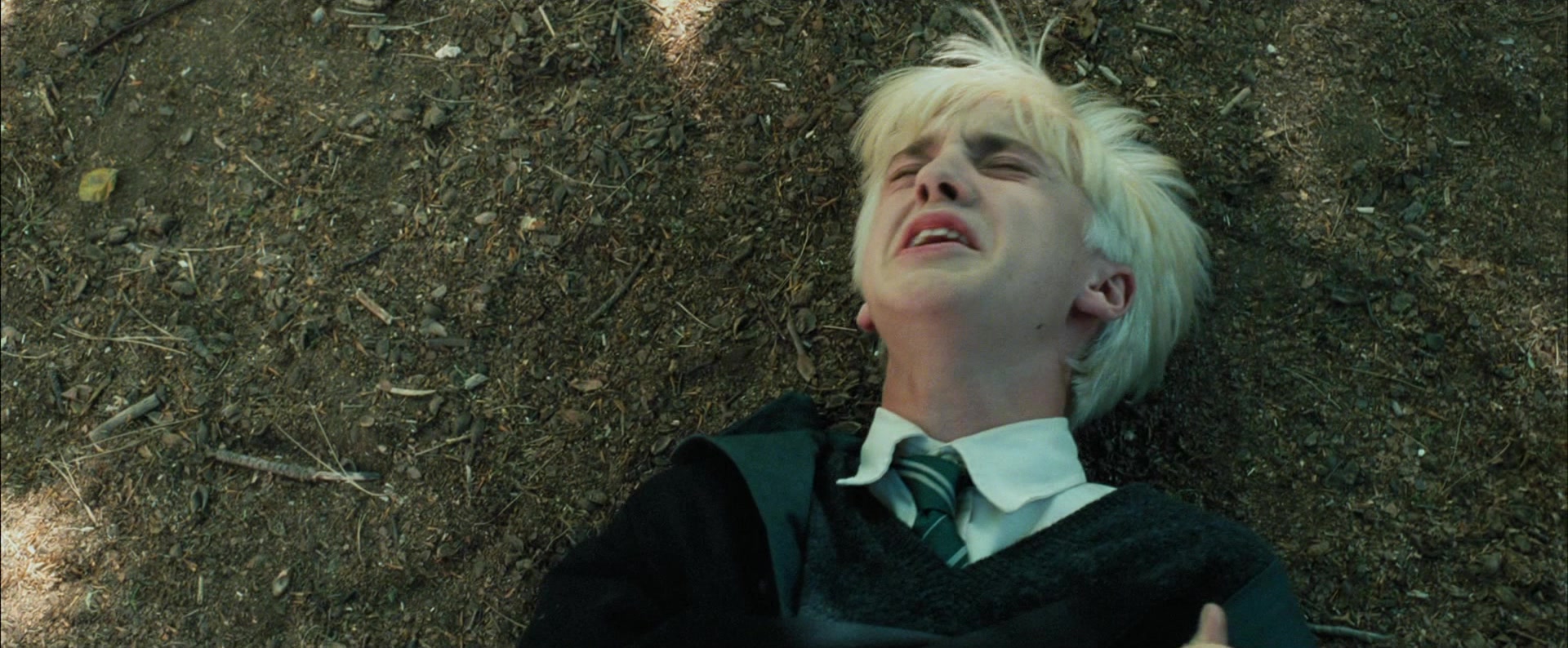} & \includegraphics[width=\linewidth]{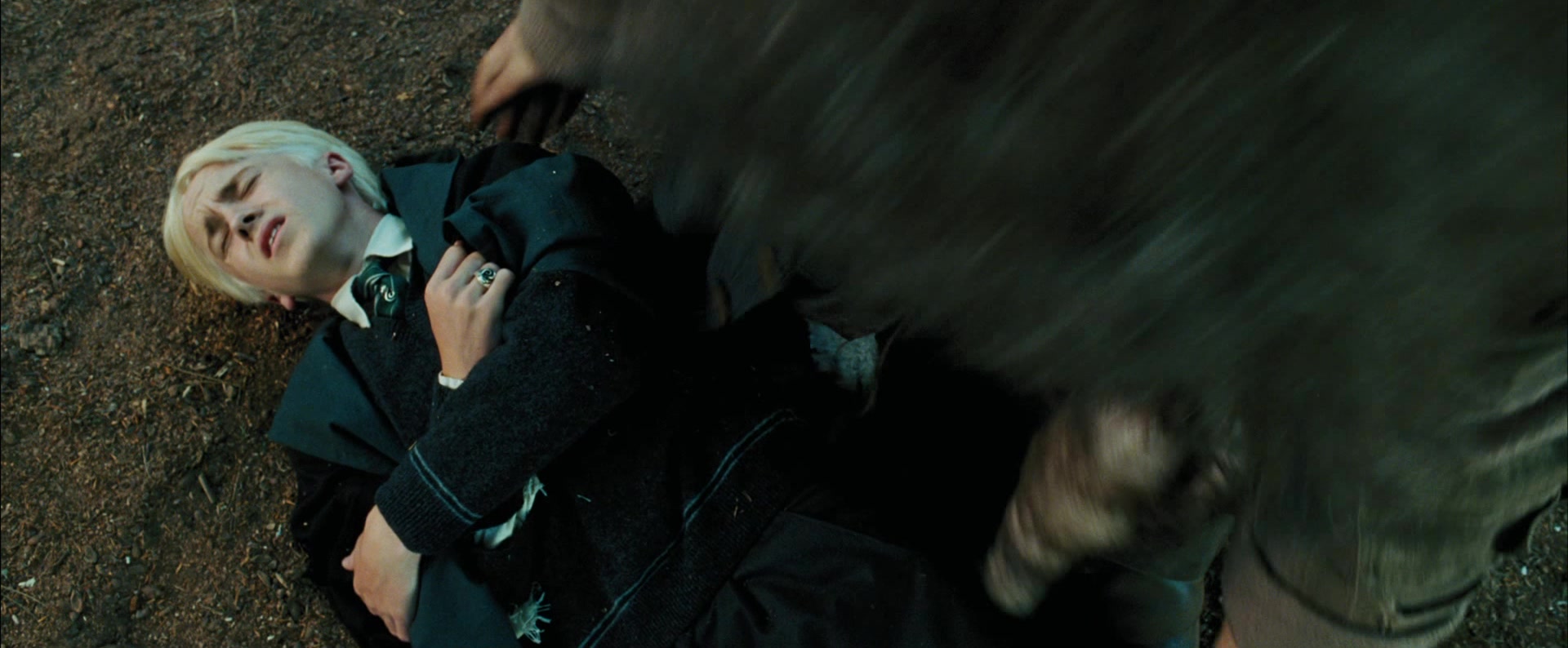} & \includegraphics[width=\linewidth]{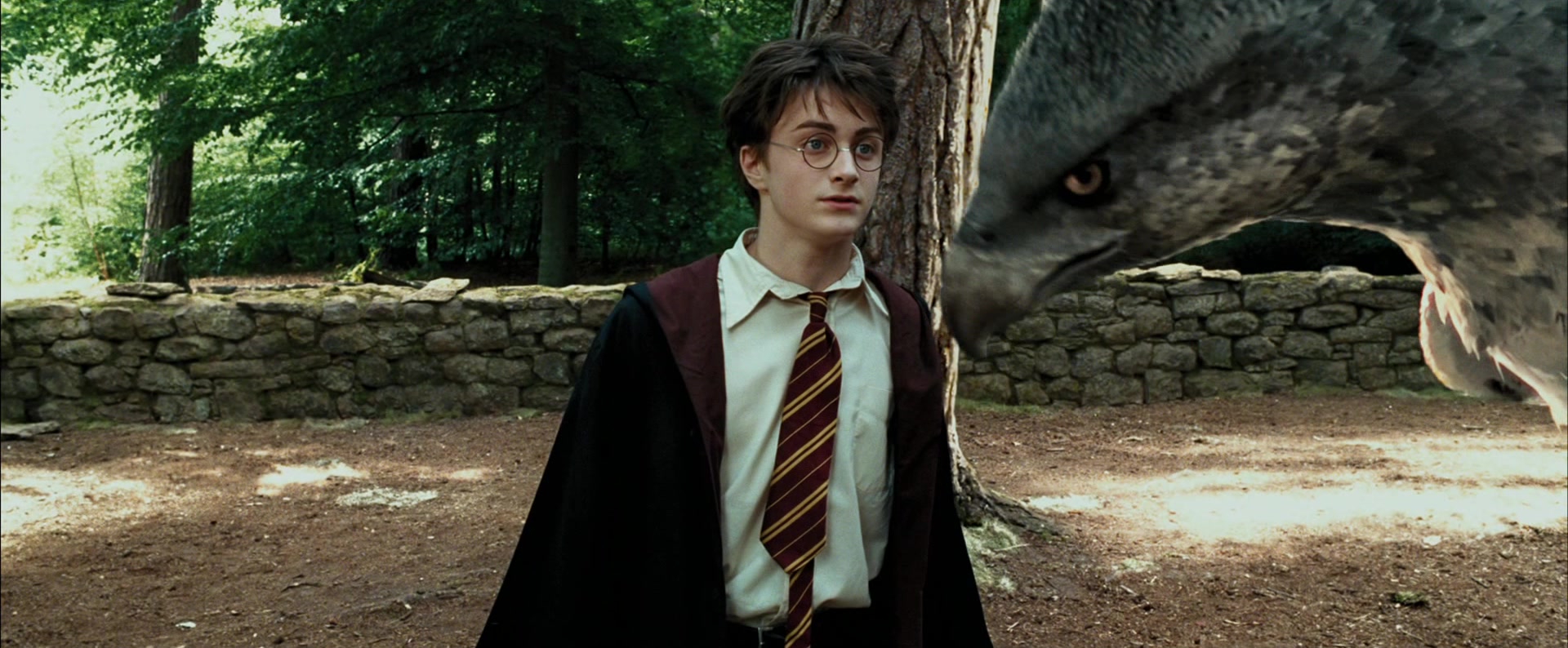}\\
 \textbf{AD}: Buckbeak rears and attacks Malfoy. &  & & Hagrid lifts Malfoy up. & As Hagrid carries Malfoy away, the hippogriff gently nudges Harry. \\
 \textbf{Script}: In a flash, Buckbeak's steely talons slash down. & Malfoy freezes. & \redtext{Looks down at the blood blossoming on his robes.} & & \redtext{Buckbeak whips around, raises its talons and - seeing Harry - lowers them.}\\
 \includegraphics[width=\linewidth]{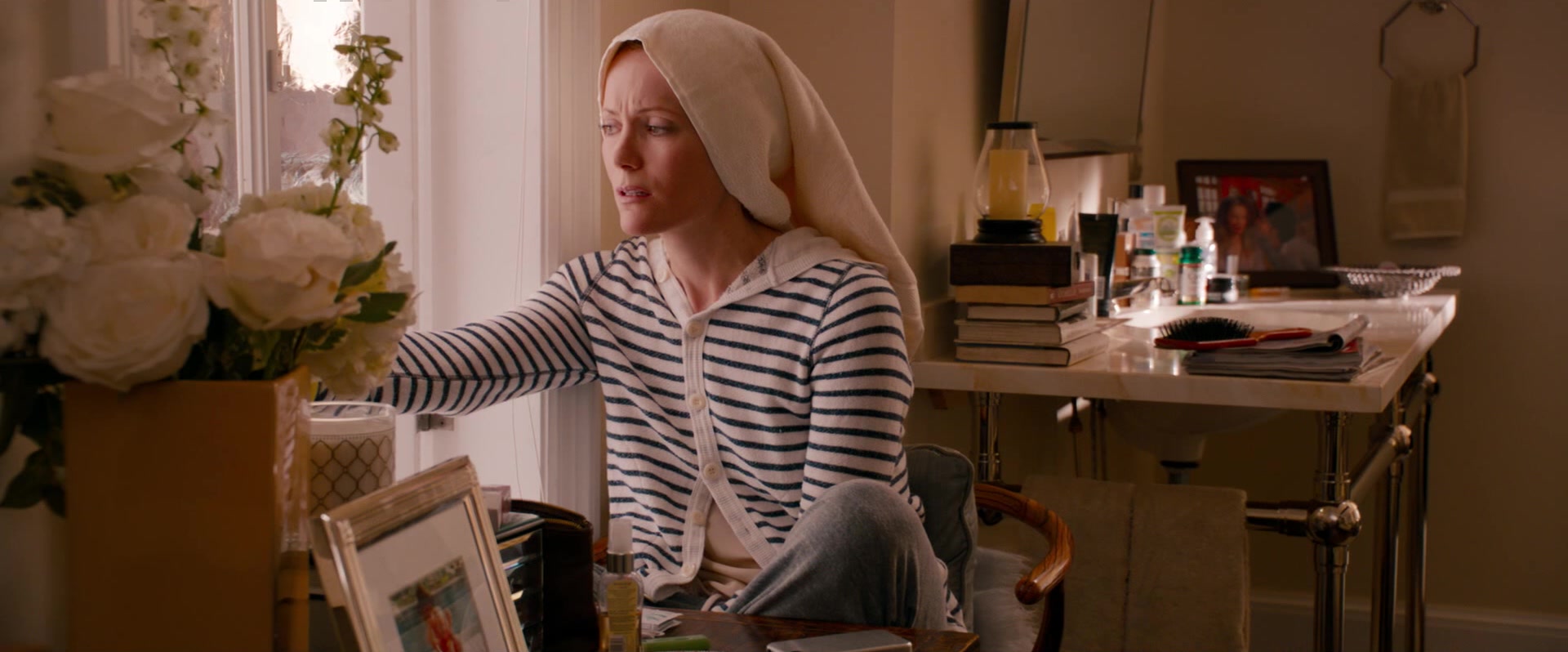} & \includegraphics[width=\linewidth]{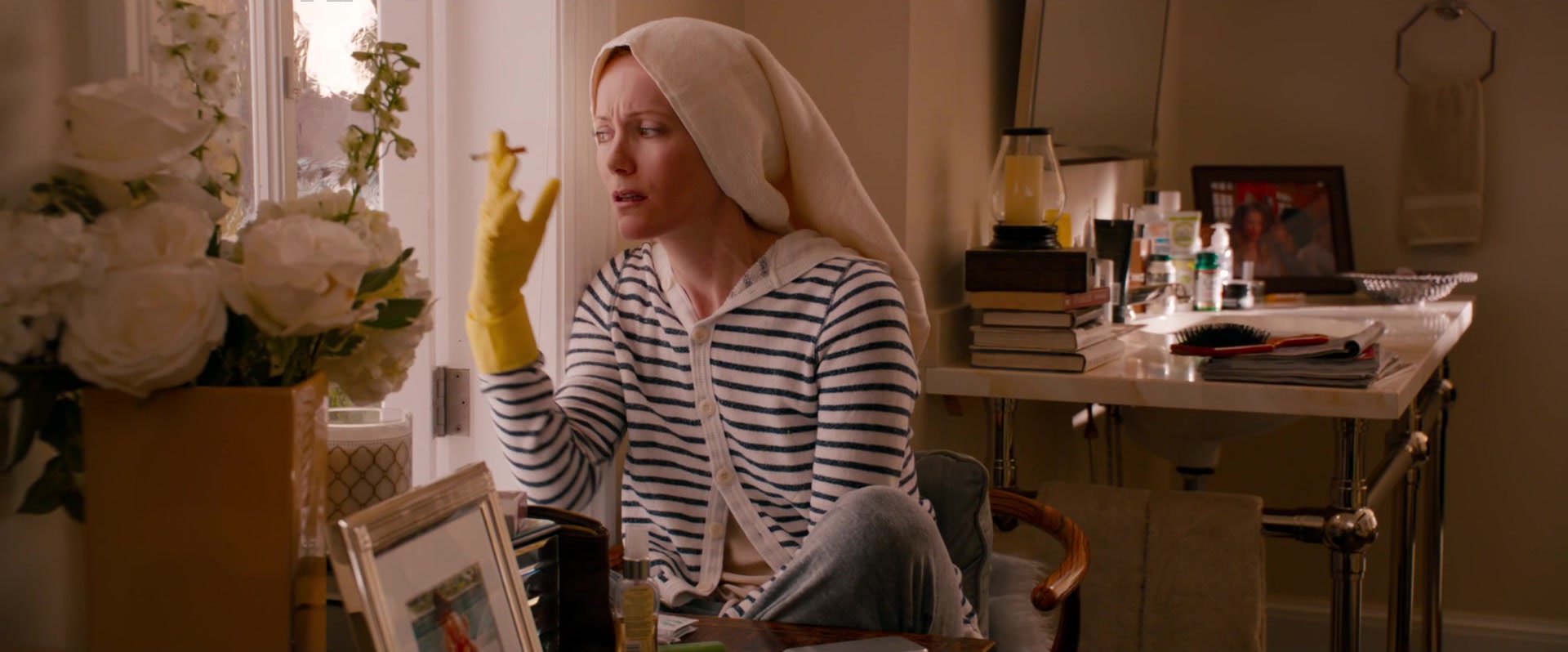} & \includegraphics[width=\linewidth]{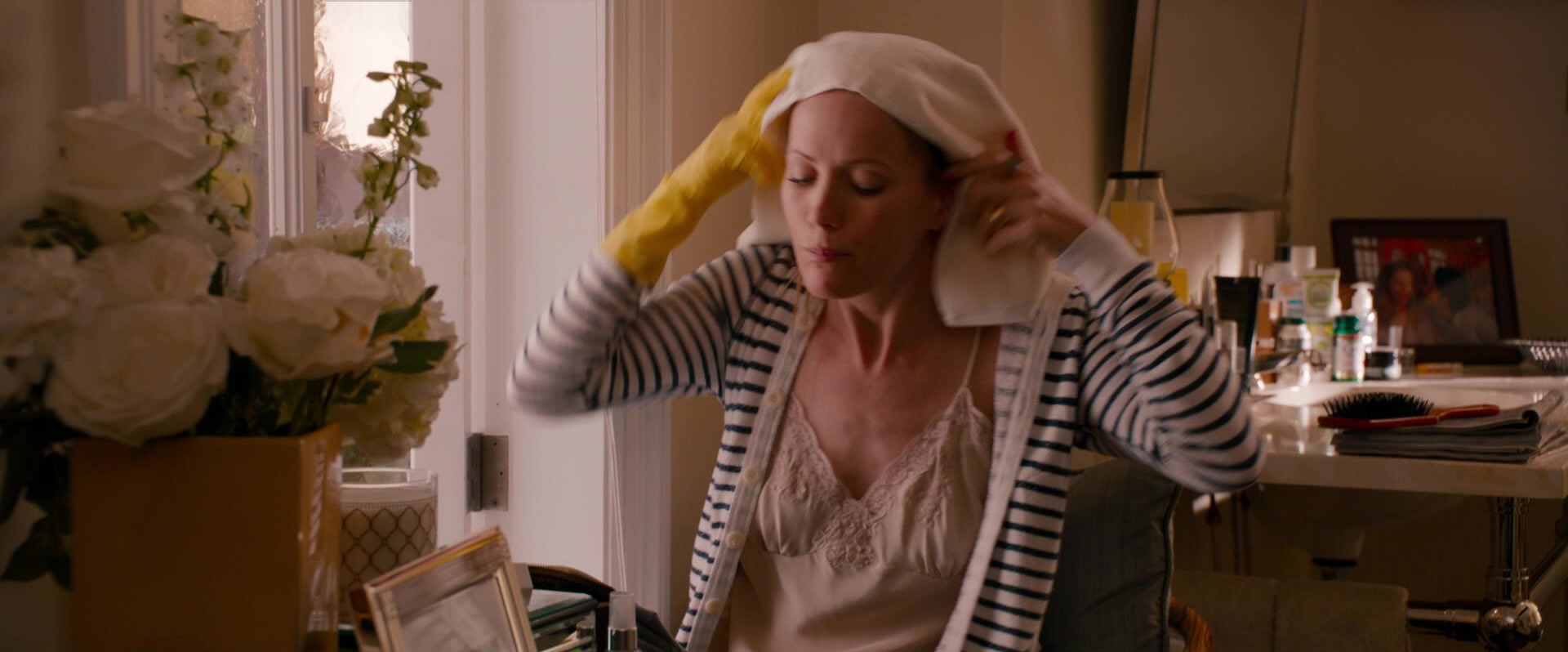} & \includegraphics[width=\linewidth]{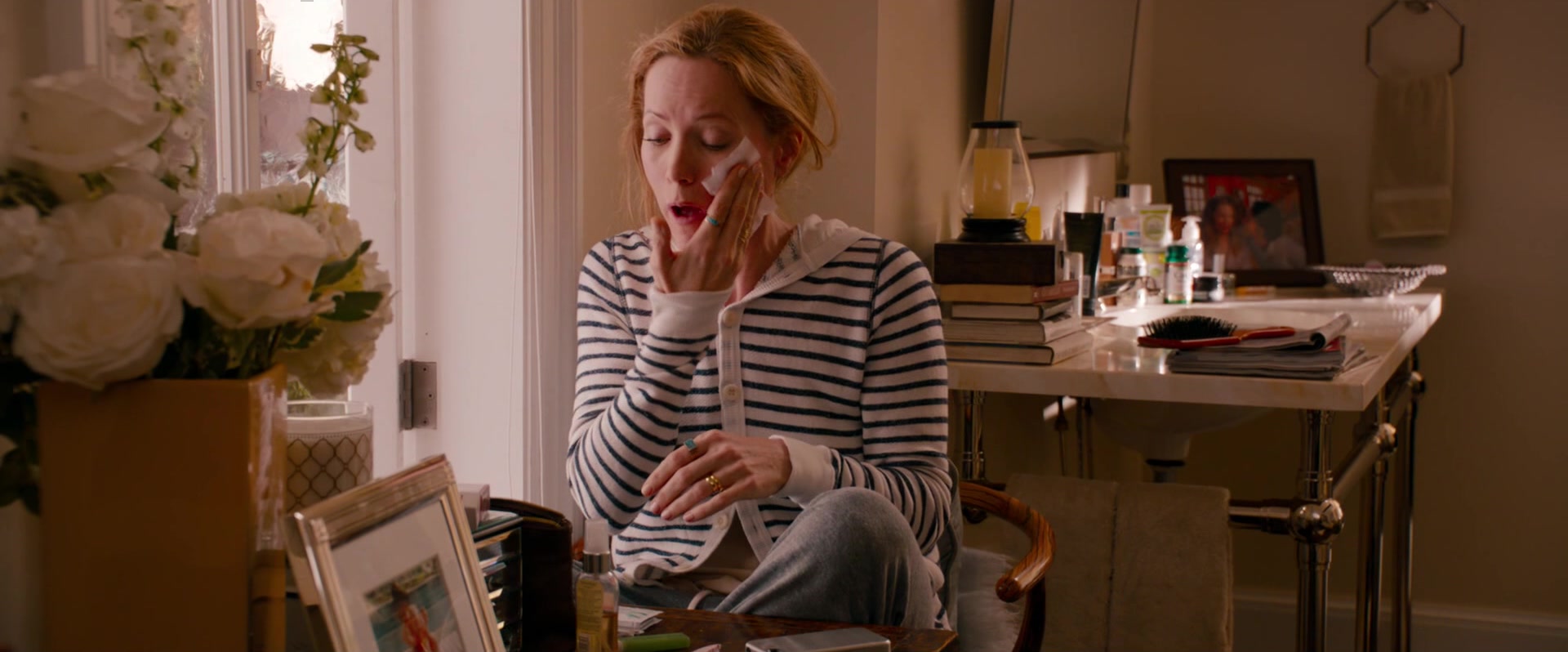} & \includegraphics[width=\linewidth]{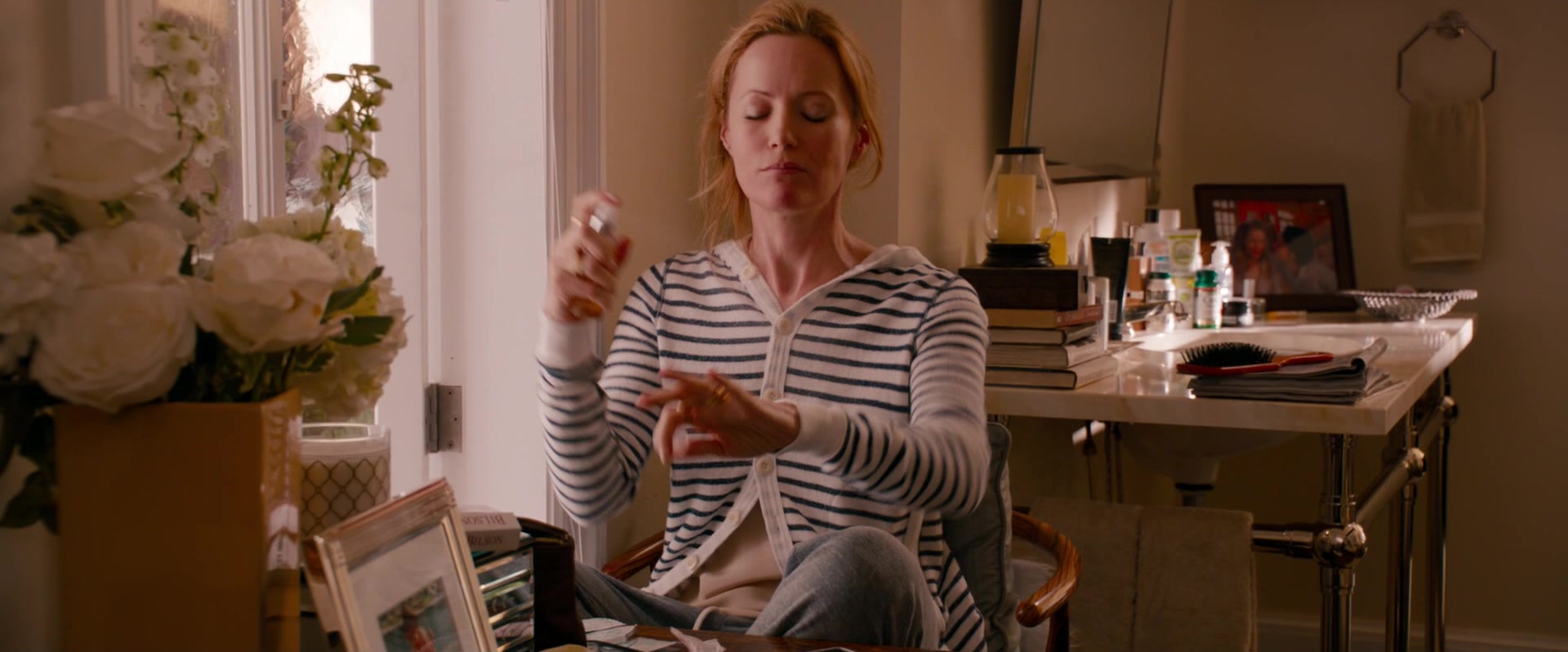}\\
 \textbf{AD}: Another room, the wife and mother sits at a window with a towel over her hair. & She smokes a cigarette with a latex-gloved hand. & Putting the cigarette out, she uncovers her hair, removes the glove and pops gum in her mouth. & She pats her face and hands with a wipe, then sprays herself with perfume. & She pats her face and hands with a wipe, then sprays herself with perfume. \\
 \textbf{Script}: Debbie opens a window and sneaks a cigarette. & She holds her cigarette with a yellow dish washing glove. &  She puts out the cigarette and goes through an elaborate routine of hiding the smell of smoke. & She \redtext{puts some weird oil in her hair} and uses a wet nap on her neck \redtext{and clothes and brushes her teeth}. & She sprays cologne \redtext{and walks through it.}\\
\\
 \includegraphics[width=\linewidth]{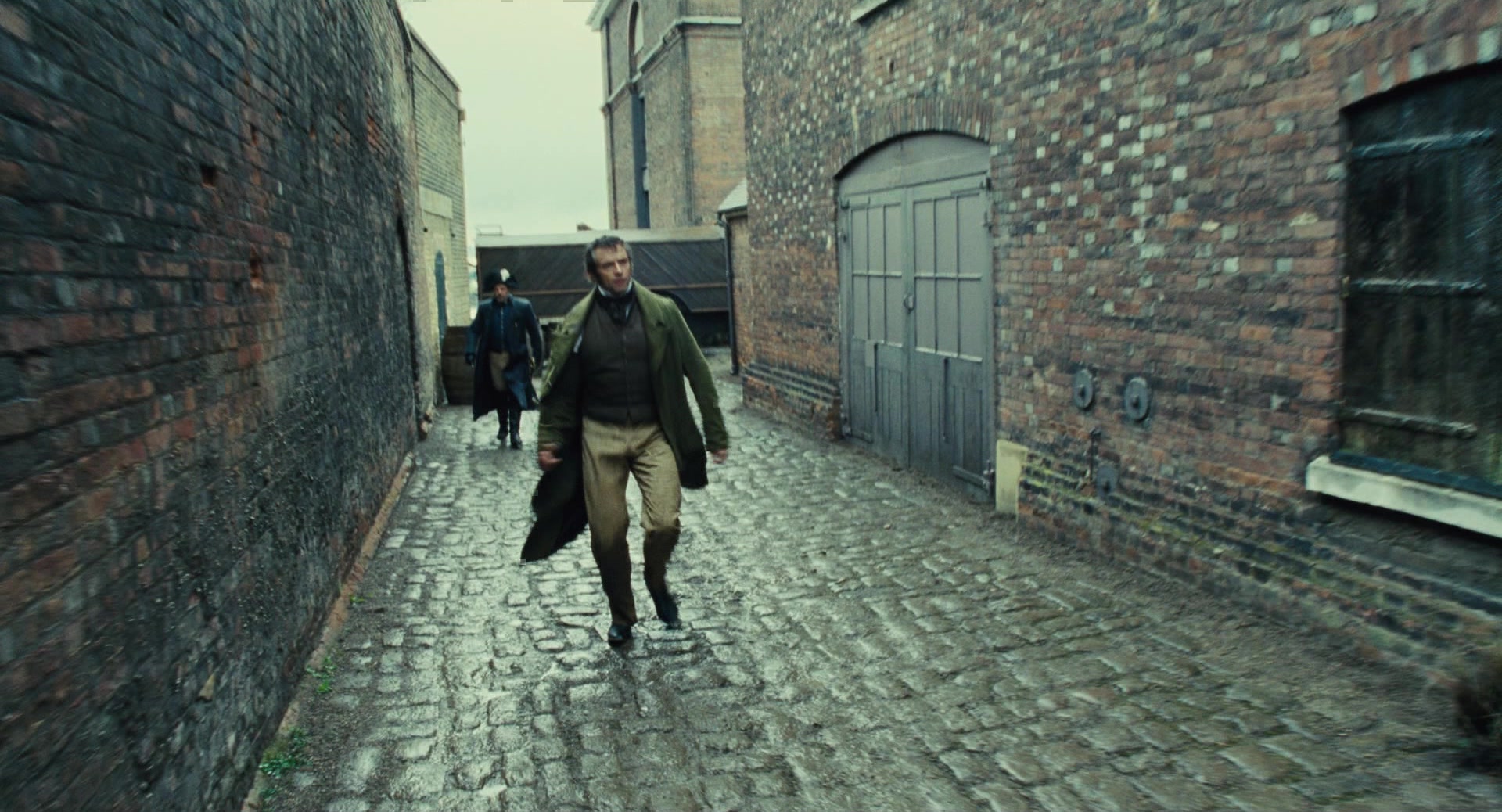} & \includegraphics[width=\linewidth]{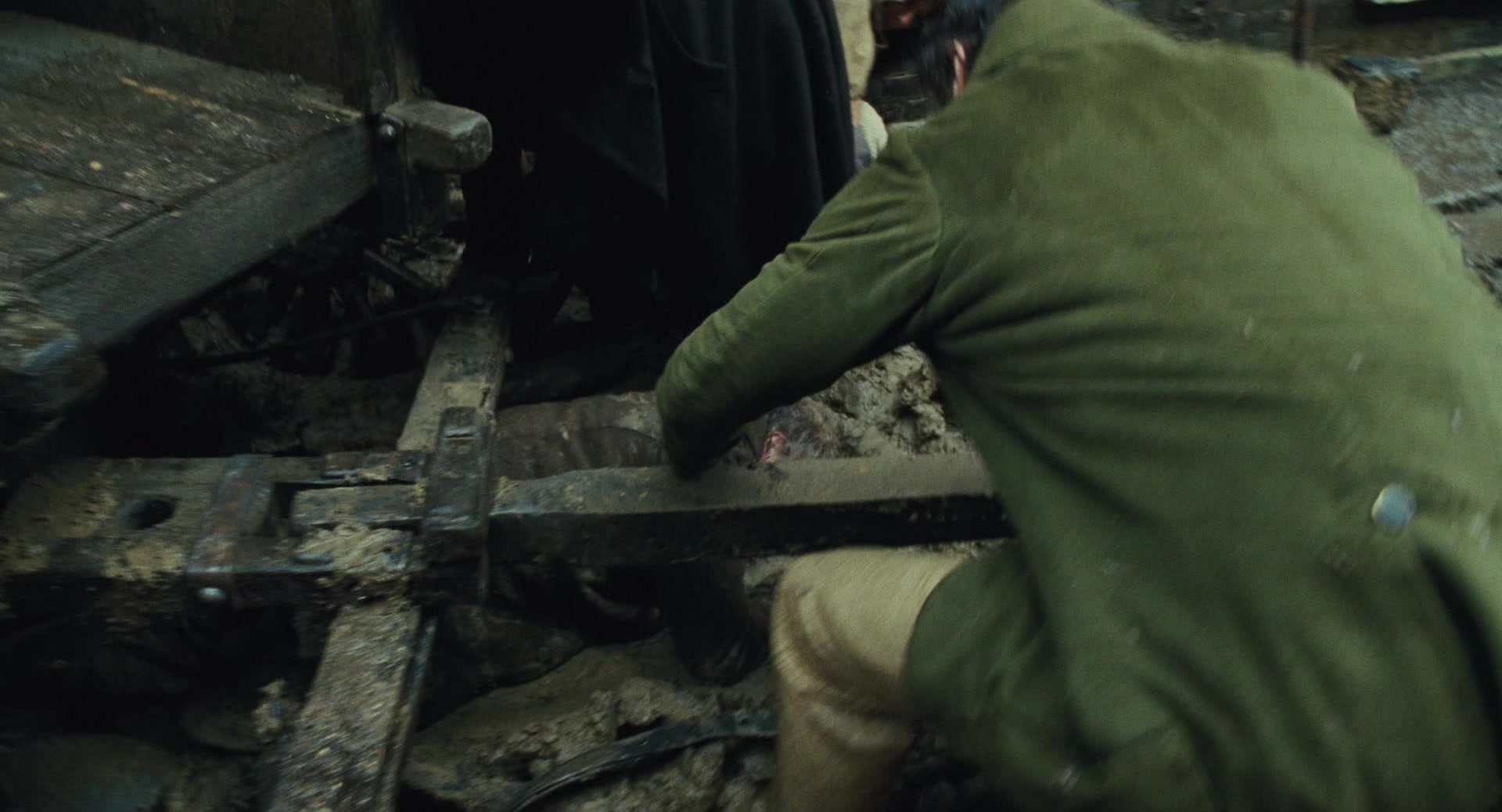} & \includegraphics[width=\linewidth]{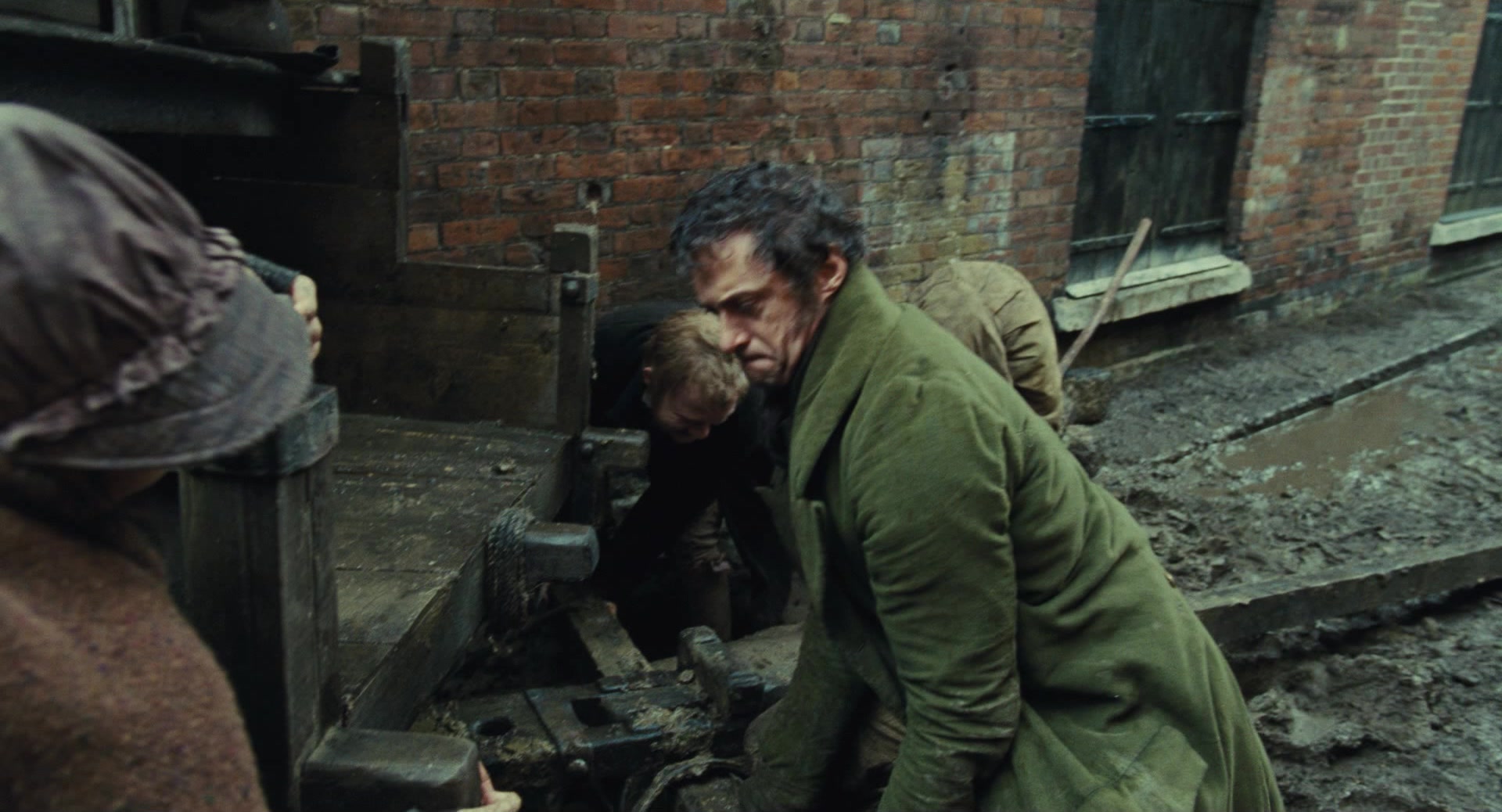} & \includegraphics[width=\linewidth]{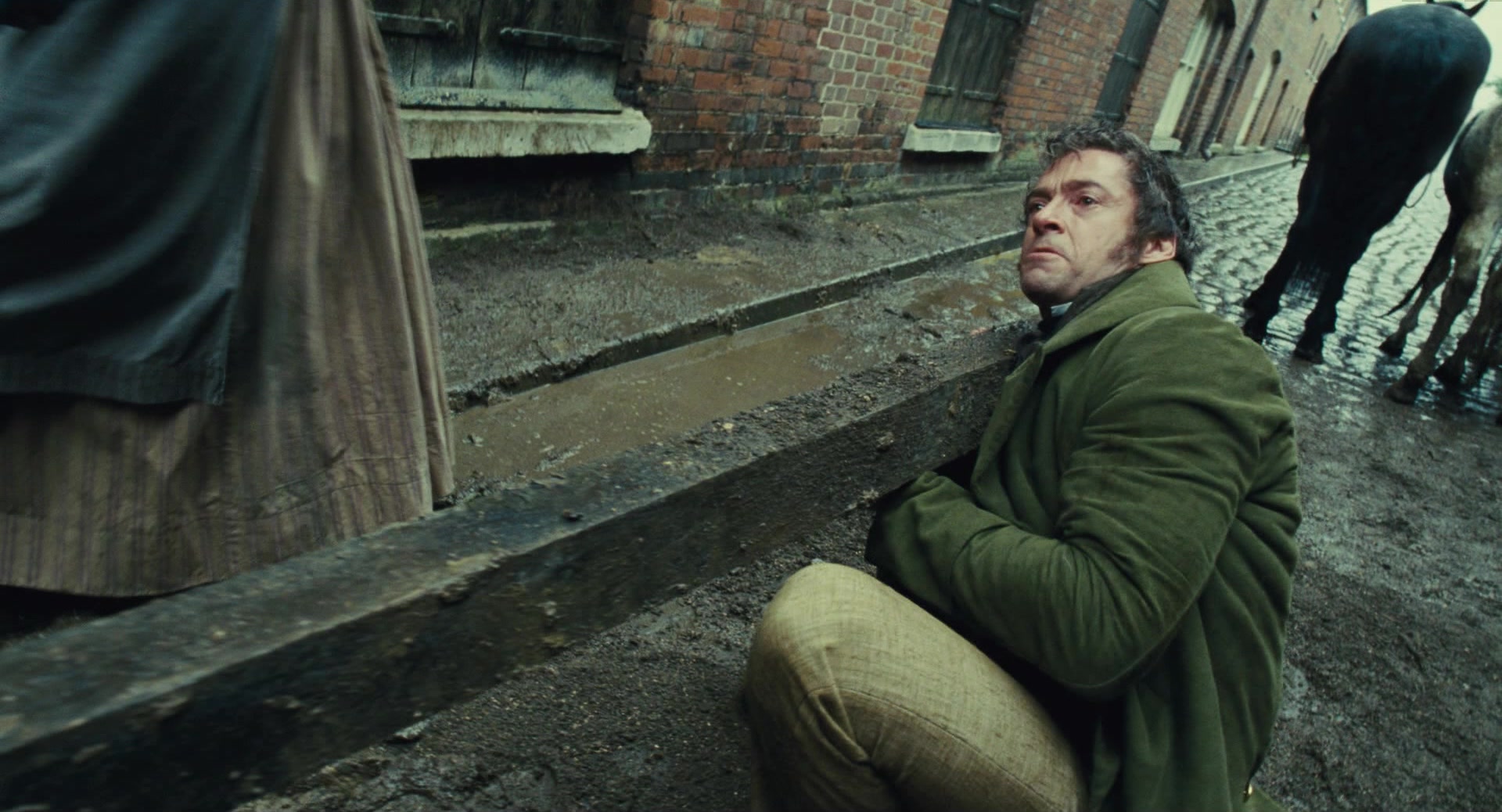} & \includegraphics[width=\linewidth]{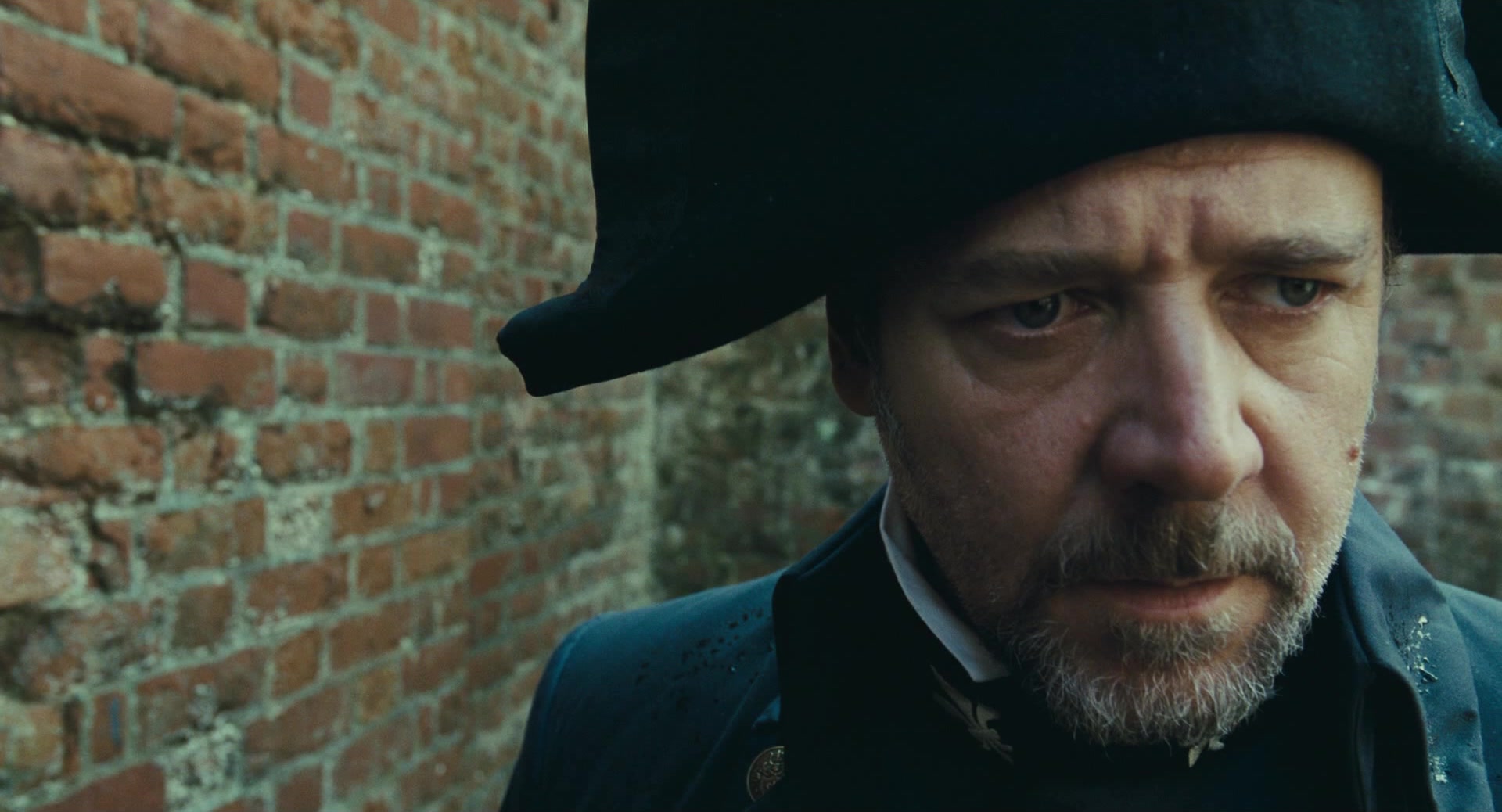}\\
 \textbf{AD}: They rush out onto the street. & A man is trapped under a cart. &  Valjean is crouched down beside him. & Javert watches as Valjean places his shoulder under the shaft. & Javert's eyes narrow. \\
 \textbf{Script}: Valjean and Javert hurry out across the factory yard and down the muddy track beyond to discover - & A heavily laden cart has toppled onto the cart driver. &  Valjean, \redtext{Javert and Javert's assistant} all hurry to help, but they can't get a proper purchase in the spongy ground. & He throws himself under the cart at this higher end, and braces himself to lift it from beneath. & Javert stands back and looks on.\\

\end{tabular}
  \caption{Audio description (AD) and  movie script samples from the movies ``Harry Potter and the Prisoner of Azkaban'', ``This is 40'', and ``Les Miserables''. Typical mistakes contained in scripts marked in \redtext{red italic}.}
  \label{fig:teaser}
\end{center}
\end{figure*}

In addition to the benefits for the blind, generating descriptions for video is an interesting task in itself, requiring the combination of core techniques from computer vision and computational linguistics. To understand the visual input one has to reliably recognize scenes, human activities, and participating objects. To generate a good description one has to decide what part of the visual information to verbalize, \ie recognize what is salient.
 
Large datasets of objects \citep{deng09cvpr} and scenes \citep{xiao10cvpr,zhou14nips} have had an important impact in computer vision and have significantly improved our ability to recognize objects and scenes. The combination of large datasets and convolutional neural networks (CNNs) has been particularly potent \citep{krizhevsky12nips}.
To be able to learn how to generate descriptions of visual content, parallel datasets of visual content paired with descriptions are indispensable~\citep{rohrbach13iccv}. While recently several large datasets have been released which provide images with descriptions \citep{flickr30k,coco2014,ordonez11nips}, video description datasets focus on short video clips with single sentence descriptions and have a limited number of video clips \citep{xu16cvpr,chen11acl} or are not publicly available \citep{over12tv}.
TACoS Multi-Level \citep{rohrbach14gcpr} and YouCook \citep{das13cvpr} are exceptions as they provide multiple sentence descriptions and longer videos. While these corpora pose challenges in terms of fine-grained recognition, they are restricted to the cooking scenario.
In contrast, movies are open domain and realistic, even though, as any other video source (\eg YouTube or surveillance videos), they have their specific characteristics. ADs and scripts associated with movies provide rich multiple sentence descriptions. They even go beyond this by telling a story which means they facilitate the study of how to extract plots, the understanding of long term semantic dependencies and human interactions from both visual and textual data. 

\begin{figure*}[t]
\scriptsize
\begin{center}
\includegraphics[width=\linewidth]{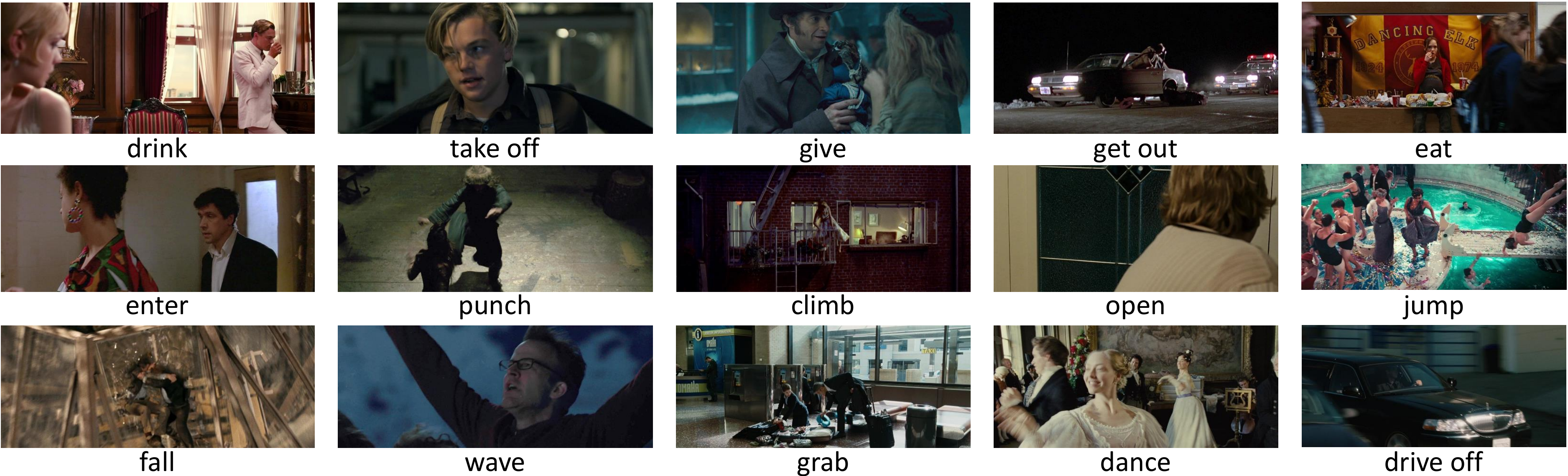}
\caption{Some of the diverse verbs / actions present in our \LaScMoDeCh (\LSMDC).}
\label{fig:actions}
\end{center}
\end{figure*}

Figures \ref{fig:teaser1} and \ref{fig:teaser} show examples of ADs and compare them to movie scripts. Scripts have been used for various tasks \citep{cour08eccv,duchenne09iccv,laptev08cvpr,liang11cvpr,marszalek09cvpr}, but so far not for video description. The main reason for this is that automatic alignment frequently fails due to the discrepancy between the movie and the script.
As scripts are typically produced prior to the shooting of the movie they are frequently not as precise as the AD (\Figref{fig:teaser} shows such mistakes marked in red). A typical case is that part of the sentence is correct, while another part contains irrelevant information.
As can be seen in the examples, AD narrations describe key visual elements of the video such as changes in the scene, people's appearance, gestures, actions, and their interaction with each other and the scene's objects in concise and precise language. \figref{fig:actions} shows the variability of AD data w.r.t. to verbs (actions) and corresponding scenes from the movies.

In this work we present a dataset which provides transcribed ADs, aligned to full length movies. 
AD soundtracks are carefully positioned within movies to fit in natural pauses in the dialogue and are mixed with the original movie soundtrack by professional post-production. For this we retrieve audio streams from DVDs and Blu-ray disks, segment out the sections of the AD audio and transcribe them via a crowd-sourced transcription service. The ADs provide an initial temporal alignment, which however does not always cover the full activity in the video.  We discuss a way to fully automate both audio-segmentation and temporal alignment, but also manually align each sentence to the movie for the majority of the data.
Therefore, in contrast to \citet{salway07corpus} and \citet{salway07civr}, our dataset provides  alignment to the actions in the video, rather than just to the audio track of the description. 
In addition we also mine existing movie scripts, pre-align them automatically, similar to \citet{cour08eccv} and \citet{laptev08cvpr}, and then manually align the sentences to the movie.

As a first study on our dataset we benchmark several approaches for movie description. We first examine nearest neighbour retrieval using diverse visual features which do not require any additional labels, but retrieve sentences from the training data. Second, we adapt the approach of \citet{rohrbach13iccv} by automatically extracting the semantic representation from the sentences using semantic parsing. %
Third, based on the success of Long Short-Term Memory networks (LSTMs) \citep{hochreiter97nc} for the image captioning problem \citep{donahue15cvpr,karpathy15cvpr,kiros15tacl,vinyals15cvpr} we propose our approach\emph{\ApproachVisualLabels}. It first builds robust visual classifiers which distinguish verbs, objects, and places extracted from weak sentence annotations. Then the visual classifiers form the input to an LSTM for generating movie descriptions. 

The main contribution of this work is the \LaScMoDeCh (\LSMDC)\footnote{\label{fn:website}{https://sites.google.com/site/describingmovies/}} which provides transcribed and aligned AD and script data sentences. 
The \LSMDC 2015 has been presented at the Workshop ``Describing and Understanding Video \& The Large Scale Movie Description Challenge (LSMDC)'', collocated with ICCV 2015. We are in progress of setting up the \LSMDC 2016 which will be presented at the ``Joint Workshop on Storytelling with Images and Videos and Large Scale Movie Description and Understanding Challenge'', collocated with ECCV 2016.
The challenge includes a public and blind test set and evaluation server\footnote{\label{fn:codalab}{https://competitions.codalab.org/competitions/6121}} for automatic evaluation. \LSMDC is based on
 the MPII Movie Description dataset (MPII-MD) and the \MoViAnDa (\MVAD) which were initially collected independently but are presented jointly in this work. We detail the data collection and dataset properties in \secref{sec:datasets}, which includes our approach to  automatically collect and align AD data. 
In \secref{sec:approaches} we present several benchmark approaches for movie description, including our \emph{\ApproachVisualLabels} approach which learns robust visual classifiers and generates description using an LSTM. 
In \secref{sec:eval} we present an evaluation of the benchmark approaches on the \MVAD and \MPIIMD datasets, analyzing the influence of the different design choices. Using automatic and human evaluation, we also show that our \ApproachVisualLabels approach outperforms prior work on both datasets. In \secref{sec:analysis} we perform an analysis of prior work and our approach to understand the challenges of the movie description task.
In \secref{sec:eval-lsmdc} we present and discuss the results of the \LaScMoDeCh 2015, which we held in conjunction with ICCV 2015.

This work is partially based on the original publications from \citet{rohrbach15cvpr,rohrbach15gcpr} and the technical report from \citet{torabi15arxiv}. \citet{torabi15arxiv} collected \MVAD, \citet{rohrbach15cvpr} collected the \MPIIMD dataset and presented the SMT-based description approach. \citet{rohrbach15gcpr} proposed the \ApproachVisualLabels approach.

\section{Related work}
We  discuss recent approaches to image and video description including existing work using movie scripts and ADs. We also discuss works which build on our dataset. We compare our proposed dataset to related video description datasets in \tableref{tbl:datasets} (\secref{sec:datasetsComparison}).

\subsection{Image description}
Prior work on image description includes \cite{farhadi10eccv,kulkarni11cvpr,kuznetsova12acl,li11acl,kuznetsova14tacl,mitchell12eacl,socher14tacl}. Recently image description has gained increased attention with work such as that of \cite{chen15cvpr,donahue15cvpr,fang15cvpr,karpathy15cvpr,kiros14icml,kiros15tacl,mao15iclr,vinyals15cvpr,xu15icml}. 
Much of the recent work has relied 
on Recurrent Neural Networks (RNNs) and in particular on Long Short-Term Memory networks (LSTMs). New datasets have been released, such as the Flickr30k \citep{young2014image} and MS COCO Captions \citep{chen15arxiv1504.00325}, where \citet{chen15arxiv1504.00325} also presents a standardized protocol for image captioning evaluation. %
Other work has analyzed the performance of recent methods, \eg \cite{devlin15acl} compare them with respect to the novelty of generated descriptions, %
while also exploring a nearest neighbor baseline that improves over recent methods.

\subsection{Video description}
In the past video description has been addressed in controlled settings \citep{barbu12uai,kojima02ijcv}, on a small scale \citep{das13cvpr,guadarrama13iccv,thomason14coling} or in single domains like cooking \citep{rohrbach14gcpr,rohrbach13iccv,donahue15cvpr}. \cite{donahue15cvpr} first proposed to describe videos using an LSTM, relying on precomputed CRF scores from \cite{rohrbach14gcpr}. Later \cite{venugopalan15naacl} extended this work to extract CNN features from frames which are max-pooled over time. %
\citet{yingweipan16cvpr} propose a framework that consists of a 2-/3-D CNN and LSTM trained jointly 
with a visual-semantic embedding to ensure better coherence between video and text.  \citet{xu2015aaai} jointly address the language generation and video/language retrieval tasks by learning a joint embedding for a deep video model and a compositional semantic language model.
\cite{li15acm} study the problem of summarizing a long video to a single concise description by using ranking based summarization of multiple generated candidate sentences. 

\paragraph{Concurrent and consequent work.}
To handle the challenging scenario of movie description,
\citet{yao2015iccv} propose a soft-attention based model which selects the most relevant temporal segments in a video, incorporates 3-D CNN and generates a sentence using an LSTM. \citet{venugopalan15iccv} propose S2VT, an encoder-decoder framework, where a single LSTM encodes the input video frame by frame and decodes it into a sentence. %
\citet{pan16cvpr} extend the video encoding idea by introducing a second LSTM layer which receives input of the first layer, but skips several frames, reducing its temporal depth.
\cite{venugopalan16arxiv} explore the benefit of pre-trained word embeddings and language models for generation on large external text corpora. \citet{shetty15arxiv} evaluate different visual features as input for an LSTM generation frame-work. Specifically they use dense trajectory features \cite{wang13ijcv} extracted for the clips and CNN features extracted at center frames of the clip. They find that training concept classifiers on MS COCO with the CNN features, combined with dense trajectories provides the best input for the LSTM. \cite{ballas16iclr} leverages multiple convolutional maps from different CNN layers to improve the visual representation for activity and video description. To model multi-sentence description, \cite{yu16cvpr} propose to use two stacked RNNs where the first one models words within a sentence and the second one, sentences within a paragraph. \citet{yao16iclr} has conducted an interesting study on performance upper bounds for both image and video description tasks on available datasets, including the LSMDC dataset.

\subsection{Movie scripts and audio descriptions}
Movie scripts have been used for automatic discovery and annotation of scenes and human actions in videos \citep{duchenne09iccv,laptev08cvpr,marszalek09cvpr}, as well as a resource to construct activity knowledge base \citep{tandon2015knowlywood,tandon2016multimodalkb}. We rely on the approach presented by \citet{laptev08cvpr} to align movie scripts using subtitles.

\citet{bojanowski13iccv} approach the problem of learning a joint model of actors and actions in movies using weak supervision provided by scripts. They rely on the semantic parser SEMAFOR \citep{das2012acl} trained on the FrameNet database \citep{Baker98acl}, however, they limit the recognition only to two frames.
\citet{bojanowski14eccv} aim to localize individual short actions in longer clips by exploiting the ordering constraints as weak supervision. \cite{bojanowski13iccv, bojanowski14eccv, duchenne09iccv, laptev08cvpr, marszalek09cvpr} proposed datasets focused on extracting several activities from movies. Most of them are part of the ``Hollywood2'' dataset \citep{marszalek09cvpr} which contains 69 movies and 3669 clips. Another line of work \citep{cour09cvpr, Everingham06bmvc, ramanathan13eccv, sivic09cvpr, tapaswi12cvpr} proposed datasets for character identification targeting TV shows. All the mentioned datasets rely on alignments to movie/TV scripts and none uses ADs.

ADs have also been used to understand which characters interact with each other \citep{salway07civr}.
 Other prior work has looked at supporting
 AD production using scripts as an information source \citep{lakritz06tr} and automatically finding scene boundaries \citep{gagnon10cvprw}.
\cite{salway07corpus} analyses the linguistic properties on a non-public corpus of ADs from 91 movies. Their corpus is based on the original sources to create the ADs and contains different kinds of artifacts not present in actual description, such as dialogs and production notes. In contrast, our text corpus is much cleaner as it consists only of the actual ADs.

\subsection{Works building on our dataset}
Interestingly, other works, datasets, and challenges are already building upon our data.
\citet{zhu15iccv} learn a visual-semantic embedding from our clips and ADs to relate movies to books. 
\citet{tapaswi16cvpr} used our AD transcripts for building their MovieQA dataset, which asks natural language questions about movies, requiring an understanding of visual and textual information, such as Dialogue and AD, to answer the question.
\citet{zhu15arxiv1511.04670} present a fill-in-the-blank challenge for audio description of the current, previous, and next sentence description for a given clip, requiring to understand the temporal context of the clips.

\section{Datasets for movie description}
\label{sec:datasets}

In the following, we present how we collected our data for movie description and discuss its properties. The \LaScMoDeCh (\LSMDC) is based on two datasets which were originally collected independently. The MPII Movie Description Dataset (MPII-MD), initially presented by \citet{rohrbach15cvpr}, was collected from Blu-ray movie data. It consists of AD and script data and uses sentence-level manual alignment of transcribed audio to the actions in the video (\secref{sec:datasets:mpiimd}). In \Secref{sec:datasets:mvad} we discuss how to fully automate AD audio segmentation and alignment for the \MoViAnDa (\MVAD), initially presented by \citet{torabi15arxiv}. \MVAD was collected with DVD data quality and only relies on AD. \secref{sec:datasets:lsmdc} details the \LaScMoDeCh (\LSMDC) which is based on \MVAD and MPII-MD, but also contains additional movies, and was set up as a challenge and includes a submission server using a public and blind test sets. In \secref{sec:datasetsStatistics} we present the detailed statistics of our datasets, also see \tableref{tab:AD-scripts-numbers}. In \secref{sec:datasetsComparison} we compare our movie description data to other video description datasets.

\subsection{The MPII Movie Description (MPII-MD) dataset}
\label{sec:datasets:mpiimd}
In the following we describe the approach behind the collection of ADs (\secref{sec:datasets:mpiimd:ads}) and script data (\secref{subsec:scripts}). Then we discuss how to manually align them to the video (\secref{sec:datasets:Manualalign}) and which visual features we extracted from the video (\secref{subsec:visual_features}).

\subsubsection{Collection of ADs}
\label{sec:datasets:mpiimd:ads}
We search for Blu-ray movies with ADs in the ``Audio Description'' section of the British Amazon\footnote{www.amazon.co.uk} and select a set of movies of diverse genres. %
As ADs are only available in audio format, we first retrieve the audio stream from the Blu-ray HD disks. We use MakeMKV\footnote{www.makemkv.com/} to extract a Blu-ray in the .mkv file format, then XMediaRecode\footnote{www.xmedia-recode.de/} to select and extract the audio streams from it. Then we semi-automatically segment out the sections of the AD audio (which is mixed with the original audio stream) with the approach described below.
The audio segments are then transcribed by a crowd-sourced transcription service\footnote{CastingWords transcription service, http://castingwords.com/} that also provides us the time-stamps for each spoken sentence. 

\paragraph{Semi-automatic segmentation of ADs.} 
We are given two audio streams: the original audio and the one mixed with the AD. We first estimate the temporal alignment between the two as there might be a few time frames difference. The precise alignment is important to compute the similarity of both streams. Both steps (alignment and similarity) are estimated using the spectograms of the audio stream, which is computed using a Fast Fourier Transform (FFT). If the difference between the two audio streams is larger than a given threshold we assume the mixed stream contains AD at that point in time. We smooth this decision over time using a minimum segment length of 1 second. The threshold was picked on a few sample movies, but had to be adjusted for each movie due to different mixing of the AD stream, different narrator voice level, and movie sound. While we found this semi-automatic approach sufficient when using a further manual alignment, we describe a fully automatic procedure in \Secref{sec:datasets:mvad}.

\subsubsection{Collection of script data}
\label{subsec:scripts}
In addition to the ADs we mine script web resources\footnote{http://www.weeklyscript.com, http://www.simplyscripts.com, http://www.dailyscript.com, http://www.imsdb.com} and select \nMoviesOnlyScript movie scripts. %
As starting point we use the movie scripts from ``Hollywood2'' \citep{marszalek09cvpr} that have highest alignment scores to their movie. We are also interested in comparing the two sources (movie scripts and ADs), so we are looking for the scripts labeled as ``Final'', ``Shooting'', or ``Production Draft'' where ADs are also available. We found that the ``overlap" is quite narrow, so we analyze \nMoviesOverlap such movies %
in our dataset. This way we end up with \nMoviesScript movie scripts in total.
We follow existing approaches \citep{cour08eccv,laptev08cvpr} to automatically align scripts to movies. First we parse the scripts, extending the method of \citep{laptev08cvpr} to handle scripts which deviate from the default format. Second, we extract the subtitles from the Blu-ray disks with SubtitleEdit\footnote{www.nikse.dk/SubtitleEdit/}. It also allows for subtitle alignment and spellchecking. Then we use the dynamic programming method of \citep{laptev08cvpr} to align scripts to subtitles and infer the time-stamps for the description sentences.
We select the sentences with a reliable alignment score (the ratio of matched words in the near-by monologues) of at least {0.5}. The obtained sentences are then manually aligned to video in-house.

\subsubsection{Manual sentence-video alignment}
\label{sec:datasets:Manualalign}
\label{sec:datasetStats}
As the AD is added to the original audio stream between the dialogs, there might be a small misalignment between the time of speech and the corresponding visual content. Therefore, we manually align each sentence from ADs and scripts to the movie in-house. 
During the manual alignment we also filter out: a) sentences describing movie introduction/ending (production logo, cast, etc); b) texts read from the screen; c) irrelevant sentences describing something not present in the video; d) sentences related to audio/sounds/music. For the movie scripts, the reduction in number of words is about {19\%}, while for ADs it is under {4\%}. In the case of ADs, filtering mainly happens due to initial/ending movie intervals and transcribed dialogs (when shown as text). For the scripts, it is mainly attributed to irrelevant sentences. Note that we retain the sentences that are ``alignable'' but contain minor mistakes. 

\subsubsection{Visual features}
\label{subsec:visual_features}
We extract video clips from the full movie based on the aligned sentence intervals. We also uniformly extract 10 frames from each video clip.
As discussed earlier, ADs and scripts describe activities, objects and scenes (as well as emotions which we do not explicitly handle with these features, but they might still be captured, \eg by the context or activities). 
In the following we briefly introduce the visual features computed on our data which are publicly available\footnote{\label{fn:mpiimd}mpii.de/movie-description}.

\textbf{IDT}
We extract the improved dense trajectories compensated for camera motion \citep{wang13iccv}. For each feature (Trajectory, HOG, HOF, MBH) we create a codebook with 4,000 clusters and compute the corresponding histograms. We apply L1 normalization to the obtained histograms and use them as features. 

\textbf{LSDA}
We use the recent large scale object detection CNN \citep{hoffman14nips} which distinguishes 7,604 ImageNet \citep{deng09cvpr} classes. We run the detector on every second extracted frame (due to computational constraints). Within each frame we max-pool the network responses for all classes, then do mean-pooling over the frames within a video clip and use the result as a feature.

\textbf{PLACES and HYBRID}
Finally, we use the recent scene classification CNNs \citep{zhou14nips} featuring 205 scene classes. We use both available networks, \emph{Places-CNN} and \emph{Hybrid-CNN}, where the first is trained on the Places dataset \citep{zhou14nips} only, while the second is additionally trained on the 1.2 million images of ImageNet (ILSVRC 2012) \citep{ILSVRCarxiv14}. We run the classifiers on all the extracted frames of our dataset. 
We mean-pool over the frames of each video clip, using the result as a feature.

\begin{figure*}[t]
\center
\includegraphics[width=13cm]{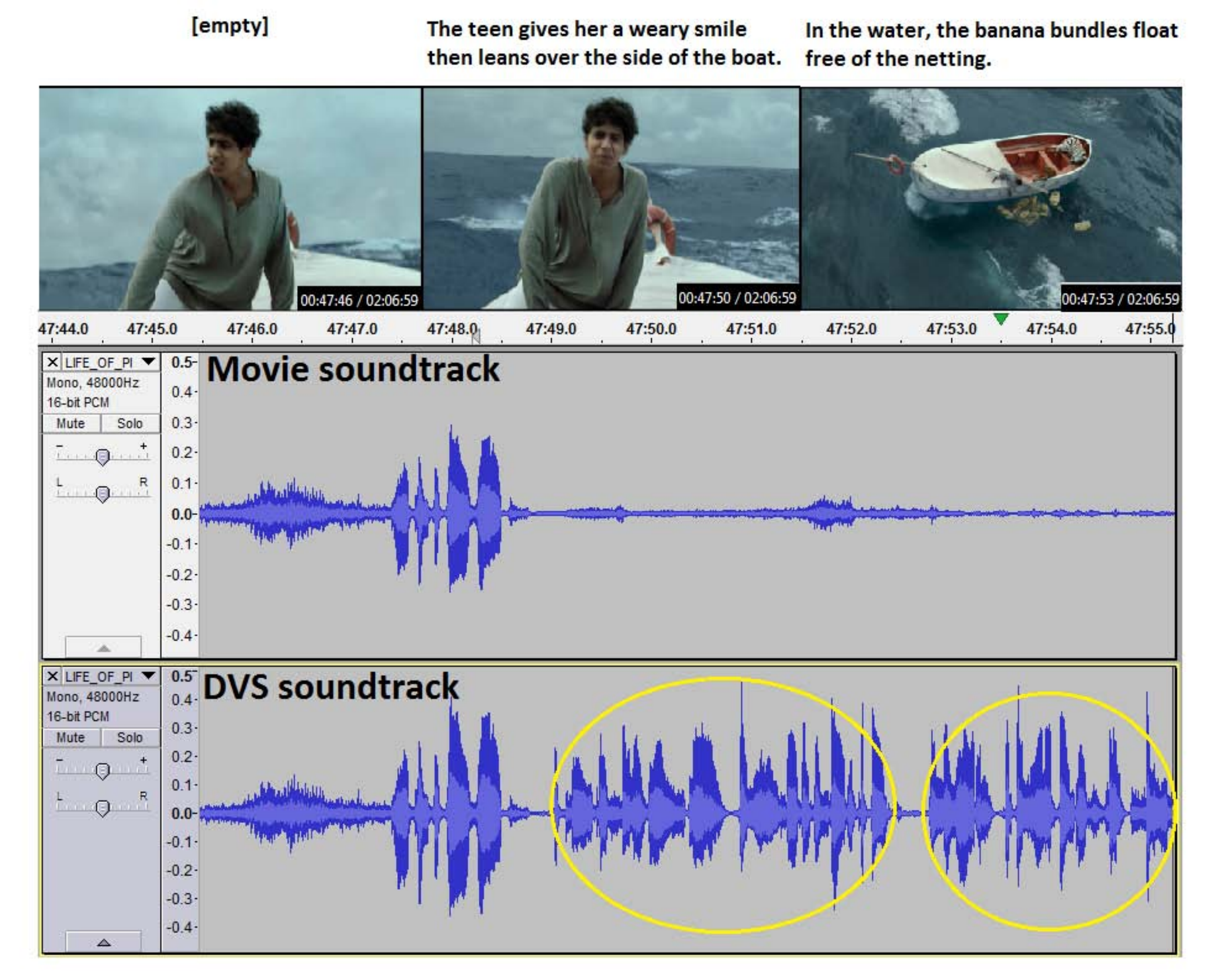}            
\caption{AD dataset collection. From the movie "Life of Pi". Line 2 and 3: Vocal isolation of movie and AD soundtrack. Second and third rows shows movie and AD audio signals after voice isolation. The two circles show the AD segments on the AD mono channel track. A pause (flat signal) between two AD narration parts shows the natural AD narration segmentation while the narrator stops and then continues describing the movie. We automatically segment AD audio based on these natural pauses. At first row, you can also see the transcription related to first and second AD narration parts on top of second and third image shots. }
\label{fig:thirdlayers}
\end{figure*}

\subsection{The \MoViAnDa (\MVAD)}
\label{sec:datasets:mvad}
One of the main challenges in automating the construction of a video annotation dataset derived from AD audio is accurately segmenting the AD output, which is mixed with the original movie soundtrack.
In this section we describe our methods for AD narration isolation and video alignment. AD narrations are typically carefully placed within key locations of a movie and edited by a post-production supervisor for continuity. For example, when a scene changes rapidly, the narrator will speak multiple sentences without pauses. Such content should be kept together when describing that part of the movie. If a scene changes slowly, the narrator will instead describe the scene in one sentence, then pause for a moment, and later continue the description. By detecting those short pauses, we are able to align a movie with video descriptions automatically. 

In the following we detail our automatic approach to AD segmentation (\secref{sec:mvad_vocal_isolation}). In \secref{sec:mvad_ad_alignment} we discuss how to align AD to the video and obtain high quality AD transcripts.

\subsubsection{AD narrations segmentation using vocal isolation}
\label{sec:mvad_vocal_isolation}
Despite the advantages offered by AD, creating a completely automated approach for extracting the relevant narration or annotation from the audio track and refining the alignment of the annotation with the scene still poses some challenges. In the following, we discuss our automatic solution for AD narrations segmentation. We use two audio tracks included in DVDs: 1) the standard movie audio signal and 2) the standard movie audio mixed with AD narrations signal.

Vocal isolation techniques boost vocals, including dialogues and AD narrations while suppressing background movie sound in stereo signals. This technique is used widely in karaoke machines for stereo signals to remove the vocal track by reversing the phase of one channel to cancel out any signal perceived to come from the center while leaving the signals that are perceived as coming from the left or the right. The main reason for using vocal isolation for AD segmentation is based on the fact that AD narration is mixed in natural pauses in the dialogue. Hence, AD narration can only be present when there is no dialogue. 
 In vocal isolated signals, whenever the narrator speaks, the movie signal is almost a flat line relative to the AD signal, allowing us to cleanly separate the narration from other dialogue by comparing the two signals. \Figref{fig:thirdlayers} illustrates an example from the movie ``Life of Pi'', where in the original movie soundtrack there are sounds of ocean waves in the background.

Our approach has three main steps. First we isolate vocals, including dialogues and AD narrations. Second, we separate the AD narrations from dialogues. Finally, we apply a simple thresholding method to extract AD segment audio tracks.  

We isolate vocals using Adobe Audition's center channel extractor\footnote{creative.adobe.com/products/audition} implementation to boost AD narrations and movie dialogues while suppressing movie background sounds on both AD and movie audio signals. We align the movie and AD audio signals by taking an FFT of the two audio signals, compute the cross-correlation, measure similarity for different offsets and select the offset which corresponds to peak cross-correlation. After alignment, we apply Least Mean Square (LMS) noise cancellation and subtract the AD mono squared signal from the movie mono squared signal in order to suppress dialogue in the AD signal. For the majority of movies on the market (among the 104 movies that we purchased, 12 movies have been mixed to the center of the audio signal, therefore we were not able to automatically align them), applying LMS results in cleaned AD narrations for the AD audio signal. Even in cases where the shapes of the standard movie audio signal and standard movie audio mixed with AD signal are very different - due to the AD mixing process - our procedure is sufficient for the automatic segmentation of AD narration.

Finally, we extract the AD audio tracks by detecting the beginning and end of AD narration segments in the AD audio signal (\ie where the narrator starts and stops speaking) using a simple thresholding method that we applied to all DVDs without changing the threshold value. 

\subsubsection{Movie/AD alignment and professional transcription}
\label{sec:mvad_ad_alignment}
AD audio narration segments are time-stamped based on our automatic AD narration segmentation. In order to compensate for the potential 1-2 seconds misalignment between the AD narrator speaking and the corresponding scene in the movie, we automatically added two seconds to the end of each video clip. Also we have discard all the transcriptions related to movie introduction/ending which are located at the beginning and the end of movies. Unlike movie scripts, in AD less than 0.5\% might be related to non-visual descriptions such as a text message on a cellphone, or street name in the movie, which we keep in our corpus.   
 
In order to obtain high quality text descriptions, the AD audio segments were transcribed with more than 98\% transcription accuracy, using a professional transcription service\footnote{TranscribeMe professional transcription, http://transcribeme.com}. These services use a combination of automatic speech recognition techniques and human transcription to produce a high quality transcription. Our audio narration isolation technique allows us to process the audio into small, well defined time segments and reduce the overall transcription effort and cost.

\begin{table*}[t]
\newcommand{\midruleADScripts}{\cmidrule(lr){1-9}}
\center
\begin{tabular}{lrrrrrrrr}
\toprule
             &Unique &       &           &       & \multicolumn{2}{l}{Average} & \multicolumn{2}{l}{Total} \\
             &Movies & Words & Sentences & Clips & \multicolumn{2}{l}{length, sec.}  & \multicolumn{2}{l}{length, h.}\\
\midruleADScripts
MPII-MD (AD)	       & 55	& 330,086 & 37,272  & 37,266  & 4.2 &(4.1) & 44.0 &(42.5) \\
MPII-MD (Movie script) & 50	& 317,728 & 31,103  & 31,071  & 3.9 &(3.6) & 33.8 &(31.1) \\
MPII-MD (Total)        & 94	& 647,814 & 68,375  & 68,337  & 4.1 &(3.9) & 77.8 &(73.6) \\
\MVAD (AD)             & 92 & 502,926 & 55,904  & 46,589  & 6.2 & - & 84.6 & - \\
\midruleADScripts
LSMDC Training        & 153 & 913,841   & 91,941 & 91,908 & 4.9 &(4.8) & 124.9 &(121.4) \\
LSMDC Validation      & 12  & 63,789    & 6,542 & 6,542 & 5.2 &(5.2) & 9.5 &(9.4) \\
LSMDC Public Test     & 17  & 87,147    & 10,053 & 10,053 & 4.2 &(4.1) & 11.6 &(11.3) \\
LSMDC Blind Test      & 20  & 83,766    & 9,578 & 9,578 & 4.5 &(4.4) & 12.0 &(11.8) \\
LSMDC (Total)    & \nLSMDCMovies & 1,148,543 & \nLSMDCSentences & \nLSMDCClips & 4.8 &(4.7) & 158.0 &(153.9) \\
\bottomrule
\end{tabular}
\caption{Movie description dataset statistics, see discussion in \secref{sec:datasetsStatistics}.}
\label{tab:AD-scripts-numbers}
\end{table*}

\subsection{The \LaScMoDeCh (\LSMDC)}
\label{sec:datasets:lsmdc}

For our Large Scale Movie Description Challenge 
(LSMDC), we combined the \MVAD and MPII-MD datasets. We first identified the overlap between the two, so that the same movie does not appear in the training and test set of the joined dataset. We also excluded script-based movie alignments from the validation and test sets of MPII-MD. The datasets were then joined by combining the corresponding training, validation and test sets, see Table \ref{tab:AD-scripts-numbers} for detailed statistics. The combined test set is used as a public test set of the challenge.  We additionally added 20 more movies where we only released the video clips, but not the aligned sentences. They form the blind test set of the challenge and are only used for evaluation. We rely on the respective best aspects of \MVAD and \MPIIMD for the public and blind test sets: we provide Blu-ray quality for them, use the automatic alignment and transcription described in \Secref{sec:datasets:mvad} and clean them using a manual alignment as in \Secref{sec:datasets:Manualalign}. We are also in the process to manually align the \MVAD validation and training sets and will release them with Blu-ray quality for \LSMDC 2016. 
We set up the evaluation server\textsuperscript{\ref{fn:codalab}} for the challenge using the Codalab\footnote{https://codalab.org/} platform. The challenge data is available online\textsuperscript{\ref{fn:website}}. We provide more information about the challenge setup and results in \secref{sec:eval-lsmdc}.

\newcommand{\midrulePOS}{\cmidrule(lr){1-1} \cmidrule(lr){2-6}}
\begin{table}[t]
\center
\begin{tabular}{p{1.5cm}r@{\ \ \ }r@{\ \ \ }r@{\ \ }r@{\ \ }r}
\toprule
Dataset & Vocab.     & Nouns & Verbs & Adjec- & Adverbs\\
        & size &       &       & tives  & \\
\midrulePOS
\MPIIMD & 18,871 & 10,558 & 2,933 & 4,239 & 1,141 \\
\MVAD & 17,609 & 9,512	& 2,571 & 3,560 & 857 \\
\LSMDC & 23,442 & 12,983 & 3,461 & 5,710 & 1,288 \\
\bottomrule
\end{tabular}
\caption{Vocabulary and POS statistics (after word stemming) for our movie description datasets, see discussion in \secref{sec:datasetsStatistics}.}
\label{tab3}
\end{table}

\begin{table*}[t]
\center
\begin{tabular}{lrrrrrr}
\toprule
Dataset& multi-sentence & domain & sentence source  & videos & clips & sentences  \\
\midrule
YouCook \citep{das13cvpr} & x & cooking & crowd & 88 & - & 2,668 \\
TACoS \citep{regneri13tacl} & x & cooking & crowd & 127 & 7,206 & 18,227  \\
TACoS Multi-Level \citep{rohrbach14gcpr}& x & cooking & crowd & 185 & 14,105 & 52,593 \\ 
MSVD \citep{chen11acl} & & open & crowd & - & 1,970 & 70,028 \\
TGIF \citep{li16cvpr} & & open & crowd & - & 100,000 & 125,781 \\
MSR-VTT \citep{xu16cvpr} & & open & crowd & 7,180 & 10,000  & 200,000 \\
\midrule
\MVAD (ours) & x & open & professional & 92 & 48,986 & 55,904 \\
MPII-MD (ours) & x & open & professional & \nMovies & \nClips & \nSentences \\
LSMDC (ours)  & x & open & professional & \nLSMDCMovies & \nLSMDCClips & \nLSMDCSentences\\
\bottomrule
\end{tabular}
\caption{Comparison of video description datasets. Discussion see Section \ref{sec:datasetsComparison}.}
\label{tbl:datasets}
\end{table*}

\subsection{Movie description dataset statistics}
\label{sec:datasetsStatistics}
Table \ref{tab:AD-scripts-numbers} presents statistics for the number of words, sentences and clips in our movie description corpora. We also report the average/total length of the annotated time intervals. If the manually aligned video clip was shorter than 2 seconds, we symmetrically expanded it (from beginning and end) to be exactly 2 seconds long. In the table we include both, the final as well as the original (precise) clip length (in brackets). In total \MPIIMD contains \nClips clips and \nSentences sentences (sometimes multiple sentences migh refer to the same video clip), while \MVAD includes 46,589 clips and 55,904 sentences. Our combined \LSMDC dataset contains over 118K sentence-clips pairs and 158 hours of video.

For \LSMDC the training/validation/public-/blind-test split consist of 91,908, 6,542, 10,053 and 9,578 video clips respectively. This split balances movie genres within each set, which is motivated by the fact that the vocabulary used to describe, say, an action movie could be very different from the vocabulary used in a comedy movie.

Table~\ref{tab3} illustrates the vocabulary size, number of nouns, verbs, adjectives, and adverbs in each respective dataset. To compute the part of speech statistics for our corpora we tag and stem all words in the datasets with the Standford Part-Of-Speech (POS) tagger and stemmer toolbox~\citep{pos}, then we compute the frequency of stemmed words in the corpora. It is important to notice that in our computation each word and its variations in corpora is counted once since we applied stemmer. Interesting observation on statistics is that \eg the number of adjectives is larger than the number of verbs, which shows that the AD is describing the characteristics of visual elements in the movie in high detail.

\subsection{Comparison to other video description datasets}
\label{sec:datasetsComparison}
We compare our corpus to other existing parallel video corpora in Table~\ref{tbl:datasets}.
The main limitations of prior datasets include the coverage of a single domain \citep{das13cvpr,regneri13tacl,rohrbach14gcpr} and having a limited number of video clips \citep{chen11acl}. Recently, two video description datasets have been proposed, namely MSR-VTT \citep{xu16cvpr} and TGIF \citep{li16cvpr}. Similar to MSVD dataset \citep{chen11acl}, MSR-VTT is based on YouTube clips. While it has a large number of sentence descriptions (200K) it is still rather small in terms of the number of video clips (10K). TGIF is a large dataset of 100k image sequences (GIFs) with associated descriptions. Both datasets are similar in that they represent web-videos, while our proposed datasets focus on movies.

\section{Approaches for movie description}
\label{sec:approaches}
\begin{figure*}[t]
\scriptsize
\begin{center}
\includegraphics[width=15cm]{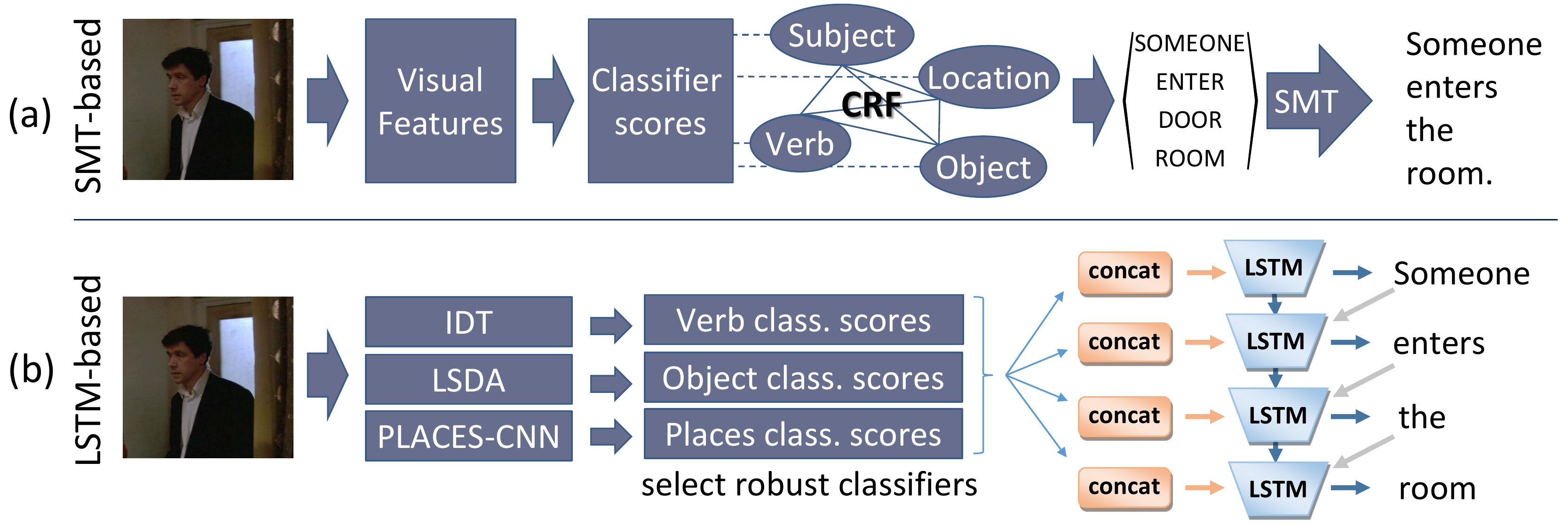}
\caption{Overview of our movie description approaches: (a) SMT-based approach, adapted from \citep{rohrbach13iccv}; (b) our proposed LSTM-based approach.}
\label{fig:smt_lstm}
\end{center}
\end{figure*}

Given a training corpus of aligned videos and sentences we want to describe a new unseen test video. In this section we discuss two approaches to the video description task that we benchmark on our proposed datasets. Our first approach in \secref{sec:smt-semantic-parsing} is based on the statistical machine translation (SMT) approach of \citep{rohrbach13iccv}. Our second approach (\secref{sec:visual_labels}) learns to generate descriptions using  Long Short-Term Memory network (LSTM).  
For the first step both approaches rely on visual classifiers learned on annotations (labels) extracted from natural language descriptions using our semantic parser (\secref{sec:semantic-parsing}). While the first approach does not differentiate which features to use for different labels, our second approach defines different semantic groups of labels and uses most relevant visual features for each group. For this reason we refer to this approach as \emph{\ApproachVisualLabels}. In the second step, the SMT-based approach uses a CRF to predict single most likely tuple (Subject, Verb, Object, Location) and translates it into a sentence. On the other hand, the LSTM-based approach takes the complete score vectors from the classifiers as input and generates a sentence based on them. \figref{fig:smt_lstm} provides an overview of the two discussed approaches.

\subsection{Semantic parsing + SMT}
\label{sec:smt-semantic-parsing}
As our first approach we adapt the two-step translation approach of \citep{rohrbach13iccv} which uses an intermediate semantic representation (SR), modeled as a tuple, \eg $\langle cut,knife,tomato \rangle$ and Statistical Machine Translation (SMT)~\citep{koehn07acl} (see \figref{fig:smt_lstm}(a)). While we cannot rely on an annotated SR as in \citep{rohrbach13iccv}, we automatically mine the SR from sentences using semantic parsing which we introduce in this section.

\newcommand{\q}[1] {``\textit{#1}''}
\newcommand{\qp}[1] {\textit{#1}}
\newcommand{\lbl}[1] {\texttt{\small #1}}
\newcommand{\ignore}[1] {}

\subsubsection{Semantic parsing}
\label{sec:semantic-parsing}
Learning from a parallel corpus of videos and natural language sentences is challenging when no annotated intermediate representation is available. In this section we introduce our approach to exploit the sentences using semantic parsing. The proposed method extracts intermediate semantic representations (SRs) from the natural sentences.

\begin{table}
\center
\small
\begin{tabular}{p{1.8cm} p{1cm} p{1.7cm} p{2.1cm}}
\toprule
Phrase        &       WordNet            & VerbNet                  & Expected \\
              &       Mapping            & Mapping                  & Frame \\
\midrule
the man       &         man\#1           & Agent.animate            & Agent: man\#1\\
\cmidrule(lr){1-4}
begin to shoot&         shoot\#4         & shoot\#vn\#3             & Action: shoot\#4\\
\cmidrule(lr){1-4}
a video       &         video\#1         & Patient.solid            & Patient: video\#1\\
\cmidrule(lr){1-4}
in            &         in               & PP.in                    & \\
\cmidrule(lr){1-4}
the moving bus&         bus\#1           & NP.Location. solid        & Location: moving bus\#1\\
\bottomrule
\end{tabular}
\caption{Semantic parse for \q{He began to shoot a video in the moving bus}. Discussion see Section \ref{sec:semantic-parsing}}
\label{tab:semantic-parse-expected-output}
\end{table}

\myparagraph{Approach.} We lift the words in a sentence to a semantic space of roles and WordNet \citep{Fellbaum1998,pedersen2004wordnet} senses by performing SRL (Semantic Role Labeling) and WSD (Word Sense Disambiguation). For an example, refer to Table \ref{tab:semantic-parse-expected-output}, the expected outcome of semantic parsing on the input sentence \q{He shot a video in the moving bus} is ``\lbl{Agent: man, Action: shoot, Patient: video, Location: bus}''. Additionally, the role fillers are disambiguated.  

We use the ClausIE tool \citep{clauseIE} to decompose sentences into their respective clauses. For example, \q{he shot and modified the video} is split into two phrases \q{he shot the video} and \q{he modified the video}). We then use the OpenNLP tool suite\footnote{http://opennlp.sourceforge.net/} for chunking the text of each clause. In order to provide the linking of words in the sentence to their WordNet sense mappings, we rely on a state-of-the-art WSD system, IMS \citep{ims-wsd}. The WSD system, however, works at a word level. We enable it to work at a phrase level. For every noun phrase, we identify and disambiguate its head word (\eg \lbl{the moving bus} to \q{bus\#1}, where \q{bus\#1} refers to the first sense of the word \lbl{bus}). We link verb phrases to the proper sense of its head word in WordNet (\eg \lbl{begin to shoot} to \q{shoot\#4}). 
The phrasal verbs such as \eg \emph{``pick up''} or \emph{``turn off''} are preserved as long as they exist in WordNet.

In order to obtain word role labels, we link verbs to VerbNet \citep{verbnet-2006,verbnet-2009}, a manually curated high-quality linguistic resource for English verbs. VerbNet is already mapped to WordNet, thus we map to VerbNet via WordNet. We perform two levels of matches in order to obtain role labels. First is the syntactic match. Every VerbNet verb sense comes with a syntactic frame \eg for \lbl{shoot}, the syntactic frame is \lbl{NP V NP}. We first match the sentence's verb against the VerbNet frames. These become candidates for the next step.
Second we perform the semantic match: VerbNet also provides a role restriction on the arguments of the roles \eg for \lbl{shoot} (sense killing), the role restriction is \lbl{Agent.animate V Patient.\textbf{animate} PP Instrument.solid}. For the other sense for \lbl{shoot} (sense snap), the semantic restriction is \lbl{Agent.animate V \break Patient.\textbf{solid}}. We only accept candidates from the syntactic match that satisfy the semantic restriction.

\myparagraph{Semantic representation.} VerbNet contains over 20 roles and not all of them are general or can be recognized reliably. Therefore, we group them to get the SUBJECT, VERB, OBJECT and LOCATION roles.
We explore two approaches to obtain the labels based on the output of the semantic parser. First is to use the extracted text chunks directly as labels. Second is to use the corresponding senses as labels (and therefore group multiple text labels). In the following we refer to these as \emph{text-} and \emph{sense-labels}.
Thus from each sentence we extract a semantic representation in a form of (SUBJECT, VERB, OBJECT, LOCATION).

\subsubsection{SMT}
For the sentence generation we build on the two-step translation approach of \citep{rohrbach13iccv}.
As the first step it learns a mapping from the visual input to the semantic representation (SR), modeling pairwise dependencies in a CRF using visual classifiers as unaries. The unaries are trained using an SVM on dense trajectories \citep{wang13iccv}. In the second step it translates the SR to a sentence using Statistical Machine Translation (SMT)~\citep{koehn07acl}. For this the approach uses a concatenated SR as input language, \eg \emph{cut knife tomato}, and natural sentence as output language, \eg \emph{The person slices the tomato.} We obtain the SR automatically from the semantic parser, as described above, \secref{sec:semantic-parsing}. In addition to dense trajectories we use the features described in \secref{subsec:visual_features}.

\begin{figure*}[t]
\begin{center}
\includegraphics[width=12cm]{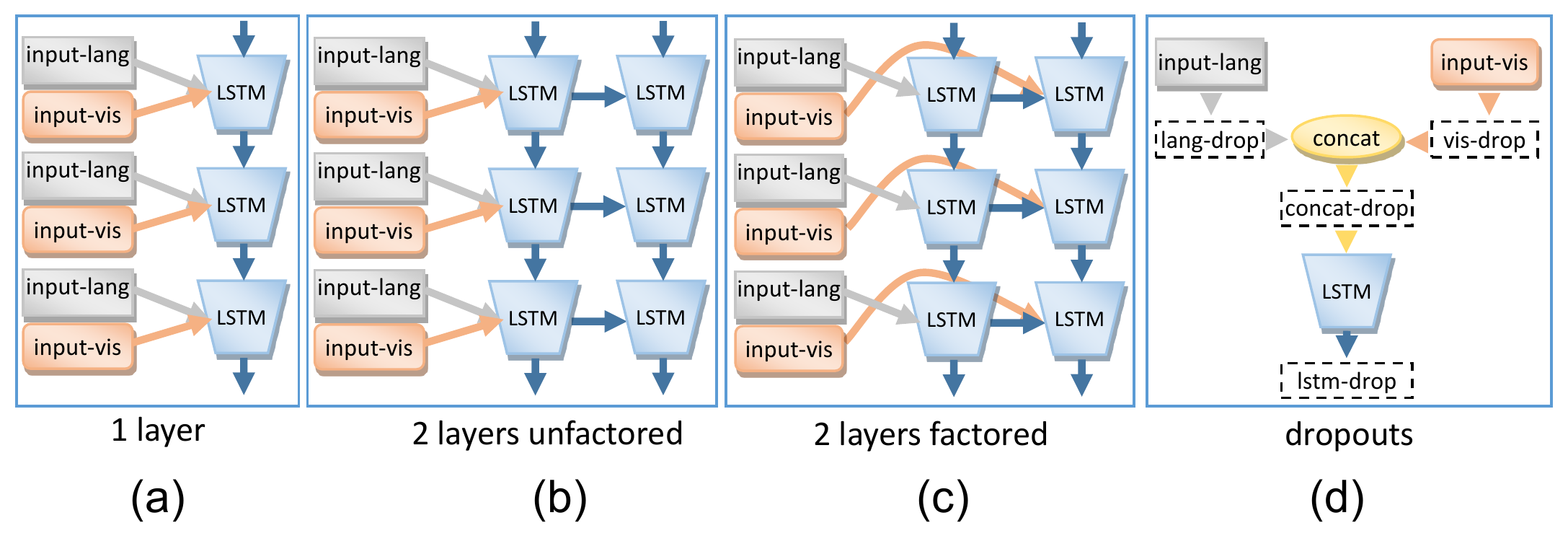}
\end{center}
\caption{(a-c) LSTM architectures. (d) Variants of placing the dropout layer.}
\label{fig:lstm_arch}
\end{figure*}

\subsection{Visual labels + LSTM}
\label{sec:visual_labels}
Next we present our two-step LSTM-based approach. The first step performs visual recognition using the visual classifiers which we train according to labels' semantics and ``visuality''. The second step generates textual descriptions using an LSTM network (see \figref{fig:smt_lstm}(b)). We explore various design choices for building and training the LSTM.

\subsubsection{Robust visual classifiers}
For training we rely on a parallel corpus of videos and weak sentence annotations. As before (see \secref{sec:smt-semantic-parsing}) we parse the sentences to obtain a set of labels (single words or short phrases, \eg \emph{look up}) to train visual classifiers. However, this time we aim to select the most visual labels which can be robustly recognized. In order to do that we take three steps. 
 
\myparagraph{Avoiding parser failure.}
Not all sentences can be parsed successfully, as \eg some sentences are incomplete or grammatically incorrect. To avoid loosing the potential labels in these sentences, we match our set of initial labels to the sentences which the parser failed to process.

\myparagraph{Semantic groups.}
Our labels correspond to different semantic groups. In this work we consider three most important groups: verbs, objects and places. We propose to treat each label group independently. First, we rely on a different representation for each semantic group, which is targeted to the specific group. Namely we use the activity recognition features Improved Dense Trajectories (DT) for verbs, LSDA scores for objects and PLACES-CNN scores for places. Second, we train one-vs-all SVM classifiers for each group separately. The intuition behind this is to avoid ``wrong negatives'' (\eg using \emph{object} ``bed'' as negative for \emph{place} ``bedroom''). 

\myparagraph{Visual labels.}
Now, how do we  select \emph{visual} labels for our semantic groups? In order to find the verbs among the labels we rely on our semantic parser (\secref{sec:semantic-parsing}). Next, we look up the list of ``places'' used in \citep{zhou14nips} and search for corresponding words among our labels. We look up the object classes used in \citep{hoffman14nips} and search for these ``objects'', as well as their base forms (\eg ``domestic cat'' and ``cat''). We discard all the labels that do not belong to any of our three groups of interest as we assume that they are likely not visual and thus are difficult to recognize. Finally, we discard labels which the classifiers could not learn reliably, as these are likely noisy or not visual. For this we require the classifiers to have certain minimum area under the  ROC-curve (Receiver Operating Characteristic).

\subsubsection{LSTM for sentence generation}
\label{sec:lstm}
We rely on the basic LSTM architecture proposed in \citep{donahue15cvpr} for video description. At each time step an LSTM generates a word and receives the visual classifiers (\emph{input-vis}) as well as as the previous generated word (\emph{input-lang}) as input (see \Figref{fig:lstm_arch}(a)). We encode each word with a one-hot-vector according to its index in a dictionary and project it in a lower dimensional embedding. The embedding is jointly learned during training of the LSTM.
We feed in the classifier scores as input to the LSTM which is equivalent to the best variant proposed in \citep{donahue15cvpr}. %
We analyze the following aspects for this architecture:

\myparagraph{Layer structure.} We compare a 1-layer architecture with a 2-layer architecture. In the 2-layer architecture, the output of the first layer is used as input for the second layer (\Figref{fig:lstm_arch}b) and was used by \citep{donahue15cvpr} for video description. Additionally we also compare to a 2-layer factored architecture of \citep{donahue15cvpr}, where the first layer only gets the language as input and the second layer gets the output of the first as well as the visual input.

\myparagraph{Dropout placement.} To learn a more robust network which is less likely to overfit we rely on a dropout \citep{hinton2012improving}, \ie a ratio $r$ of randomly selected units is set to 0 during training (while all others are multiplied with $1/r$). We explore different ways to place dropout in the network, \ie either for language input (\emph{lang-drop}) or visual (\emph{vis-drop}) input only, for both inputs (\emph{concat-drop}) or for the LSTM output (\emph{lstm-drop}), see \Figref{fig:lstm_arch}(d). %

\section{Evaluation}
\label{sec:eval}
In this section we provide more insights about our movie description datasets. First we compare ADs to movie scripts and then benchmark the approaches to video description introduced in Section \ref{sec:approaches} as well as other related work.

\subsection{Comparison of AD vs script data}
\label{sec:comparisionAD}
We compare the AD and script data using {11} movies from the MPII-MD dataset where both are available (see Section \ref{subsec:scripts}). For these movies we select the overlapping time intervals with an intersection over union overlap of at least {75\%}, which results in 279 sentence pairs, we remove 2 pairs which have idendical sentences. We ask humans via Amazon Mechanical Turk (AMT) to compare the sentences with respect to their correctness and relevance to the video, using both video intervals as a reference (one at a time). Each task was completed by 5 different human subjects, covering 2770 tasks done in total. Table \ref{tab:AD-scripts} presents the results of this evaluation. AD is ranked as more correct and relevant in about 2/3 of the cases, which supports our intuition that scrips contain mistakes and irrelevant content even after being cleaned up and manually aligned. 

\begin{table}[t]
\center
\begin{tabular}{ll@{ (}l@{) \ \ }ll@{ (}l@{)}}
\toprule
                & \multicolumn{2}{c}{Correctness}  & & \multicolumn{2}{c}{Relevance} \\
\midrule
Movie scripts	& 33.9	& 11.2   & & 33.4 &	16.8\\
ADs	            & 66.1	& 35.7   & & 66.6 &	44.9\\
\bottomrule
\end{tabular}
\caption{Human evaluation of movie scripts and ADs: which sentence is more correct/relevant with respect to the video (forced choice). Majority vote of 5 judges in \%. In brackets: at least 4 of 5 judges agree. See also Section \ref{sec:comparisionAD}.}
\label{tab:AD-scripts}
\end{table}

\subsection{Semantic parser evaluation}
\label{sec:semanticParserEval}
Table \ref{tab:semantic-parse-accuracy-per-source-detailed} reports the accuracy of individual components of the semantic parsing pipeline. The components are clause splitting (Clause), POS tagging and chunking (NLP), semantic role labeling (Roles) and word sense disambiguation (WSD). We manually evaluate the correctness on a randomly sampled set of sentences using human judges. %
It is evident that the poorest performing parts are the NLP and the WSD components. Some of the NLP mistakes arise due to incorrect POS tagging. WSD is considered a hard problem and when the dataset contains less frequent words, the performance is severely affected. %

\subsection{Evaluation metrics for description}
\label{sec:eval-metrics}
In this section we describe how we evaluate the generated descriptions using automatic and human evaluation.

\begin{table}
\centering
\begin{tabular}{@{\ }ll@{\ \ }l@{\ \ \ }l@{\ \ }l@{\ }}
\toprule
Corpus &Clause&NLP&Roles&WSD \\
\midrule
MPII-MD & 0.89  & 0.62  & 0.86  & 0.7  \\
\bottomrule
\end{tabular}
\caption{Semantic parser accuracy on MPII-MD. Discussion in Section \ref{sec:semanticParserEval}.}
\label{tab:semantic-parse-accuracy-per-source-detailed}
\end{table}

\subsubsection{Automatic metrics}
For automatic evaluation we rely on the MS COCO Caption Evaluation API\footnote{https://github.com/tylin/coco-caption}. The automatic evaluation measures include BLEU-1,-2,-3,-4 \citep{papineni02acl}, METEOR \citep{lavie2014meteor}, ROUGE-L \citep{lin2004rouge}, and CIDEr \citep{vedantam2014cider}. 
While we report all measures for the final evaluation in the \LSMDC (\secref{sec:eval-lsmdc}), we focus our discussion on METEOR score in the preliminary evaluations in this section. According to \citep{elliott2013image,vedantam2014cider}, METEOR supersedes previously used measures such as BLEU  or ROUGE in terms of agreement with human judgments. METEOR also outperforms CIDEr when the number of references is small and in the case of the movie description data we have only a single reference.
\subsubsection{Human evaluation}
For the human evaluation we rely on a ranking approach, \ie human judges are given multiple descriptions from different systems, and are asked to rank them with respect to the following criteria: correctness, relevance, and grammar, motivated by prior work \citet{rohrbach13iccv} and on the other hand we asked human judges to rank sentences for ``how helpful they would be for a blind person to understand what is happening in the movie''. The AMT workers are given randomized sentences, and, in addition to some general instruction, the following definitions:
\paragraph{Grammar.}
``Rank grammatical correctness of sentences: Judge the fluency and readability of the sentence (independently of the correctness with respect to the video).''
\paragraph{Correctness.}
``Rank correctness of sentences:
For which sentence is the content more correct with respect to the video (independent if it is complete, \ie describes everything), independent of the grammatical correctness.''
\paragraph{Relevance.} ``Rank relevance of sentences:
Which sentence contains the more salient (\ie relevant, important) events/objects of the video?''

\paragraph{Helpful for the blind.}
In the \LSMDC evaluation we introduced a new measure, which should capture how useful a description would be for blind people:

``Rank the sentences according to how useful they would be for a blind person which would like to understand/follow the movie without seeing it.''

\subsection{Movie description evaluation}
\label{sec:moviedescription}
As the collected text data comes from the movie context, it contains a lot of information specific to the plot, such as names of the characters. We pre-process each sentence in the corpus, transforming the names to ``Someone'' or ``people'' (in case of plural). %

We first analyze the performance of the proposed approaches on the MPII-MD dataset, and then evaluate the best version on the \MVAD dataset. For  MPII-MD we split the {11} movies with associated scripts and ADs (in total {22} alignments, see Section \ref{subsec:scripts}) into validation set ({8}) and test set ({14}). The other {83} movies are used for training. On \MVAD we use {10} movies for testing, {10} for validation and {72} for training.

\begin{table}[t]
\center
\begin{tabular}{lr}
\toprule
\multicolumn{2}{r}{METEOR} \\
\midrule
\multicolumn{2}{l}{SMT with our sense-labels}\\
IDT 30	&	4.93\\
IDT 100	&	5.12\\
Combi 100	&	5.19\\
\midrule
\multicolumn{2}{l}{SMT with our text-labels}\\
IDT 30	&	5.59\\
IDT 100	&	5.51\\
Combi 100	&	5.42\\
\bottomrule
\end{tabular}
\caption{Video description performance of different SMT versions on MPII-MD. Discussion in Section \ref{sec:VideoDescription}.}
\label{tab:automaticevalsmt}
\end{table}

\newcommand{\midruleValLong}{\cmidrule(rr){1-1} \cmidrule(rr){2-4}}
\begin{table*}[t]
\begin{center}
\begin{tabular}{l@{\ \ \ }r@{\ \ \ }r@{\ \ \ }r}
\toprule
         &        & \multicolumn{2}{c}{Classifiers (\scriptsize{METEOR} in \%)} \\
Approach & Labels & Retrieved   & Trained \\
\midruleValLong
\multicolumn{4}{l}{\textbf{Baseline: all labels treated the same way}} \\
(1) IDT                 & 1263 & - & 6.73 \\
(2) LSDA               & 1263 & - & 7.07 \\
(3) PLACES             & 1263 & - & 7.10 \\
(4) IDT+LSDA+PLACES     & 1263 & - & 7.24 \\
\multicolumn{4}{l}{\textbf{Visual labels}} \\
(5) Verbs(IDT), Others(LSDA)                        & 1328 & 7.08 & 7.27 \\
(6) Verbs(IDT), Places(PLACES), Others(LSDA)        & 1328 & 7.09 & 7.39 \\
(7) Verbs(IDT), Places(PLACES), Objects(LSDA)       & 913 & 7.10 & 7.48 \\
(8) \ \ + restriction to labels~with~$ROC~\ge~0.7$  & 263 & 7.41 & \textbf{7.54} \\
\multicolumn{4}{l}{\textbf{Baseline: all labels treated the same way, labels from (8)}} \\
(9) IDT+LSDA+PLACES                                 & 263 & 7.16 & 7.20 \\
\bottomrule
\end{tabular}
\end{center}
\caption{Comparison of different choices of labels and visual classifiers. All results reported on the validation set of MPII-MD. For discussion see \secref{sec:eval:visuallabelsLSTM}.}
\label{tbl:valset_labels_viscls}
\end{table*}

\newcommand{\midruleValShort}{\cmidrule(lr){1-1} \cmidrule(lr){2-2}}
\begin{table*}[t]
\begin{center}
\begin{tabular}{@{}ccc@{}}
\begin{tabular}{lr}
\toprule
Architecture & \scriptsize{$METEOR$} \\
\midruleValShort
{1 layer}             & \textbf{7.54} \\
{2 layers unfact.} & \textbf{7.54} \\
{2 layers fact.}   & {7.41} \\
\bottomrule
\end{tabular}
&
\begin{tabular}{lr}
\toprule
Dropout & \scriptsize{$METEOR$} \\
\midruleValShort
{no dropout}               & 7.19 \\
{lang-drop} & 7.13 \\
{vis-drop}   & 7.34 \\
{concat-drop}           & 7.29 \\
{lstm-drop}   & \textbf{7.54} \\
\bottomrule
\end{tabular}
&
\begin{tabular}{lr}
\toprule
Dropout ratio & \scriptsize{$METEOR$} \\
\midruleValShort
{r=0.1}         & 7.22 \\
{r=0.25}        & 7.42 \\
{r=0.5}         & \textbf{7.54} \\
{r=0.75}        & 7.46 \\
\bottomrule 
\end{tabular}\\
(a)&(b)&(c)\\
 LSTM architectures (lstm-drop 0.5)&Dropout strategies (1-layer, dropout 0.5) &Dropout ratios (1-layer,lstm-drop)\\
\end{tabular}
\caption{LSTM architectures, MPII-MD val set. Labels, classifiers as Table \ref{tbl:valset_labels_viscls}, line (8). For discussion see \secref{sec:eval:visuallabelsLSTM}.}
\label{tbl:valset_configs}
\end{center}
\end{table*}

\subsubsection{Semantic parsing + SMT}
\label{sec:VideoDescription}

Table \ref{tab:automaticevalsmt} summarizes results of multiple variants of the SMT approach when using the SR from our semantic parser. ``Combi'' refers to combining IDT, HYBRID, and PLACES as unaries in the CRF. We did not add LSDA as we found that it reduces the performance of the CRF. After extracting the labels we select the ones which appear at least 30 or 100 times as our visual attributes. %
Overall, we observe similar performance in all cases, with slightly better results for text-labels than sense-labels. This can be attributed to sense disambiguation errors of the semantic parser. In the following we use the ``IDT 30'' model, which achieves the highest score of {5.59}, and denote it as ``SMT-Best''\footnote{We also evaluated the ``Semantic parsing+SMT'' approach on a corpus where annotated SRs are available, namely TACoS Multi-Level \citep{rohrbach14gcpr}, and showed the comparable performance to manually annotated SRs, see \citep{rohrbach15cvpr}.}.

\subsubsection{Visual labels + LSTM}
\label{sec:eval:visuallabelsLSTM}
We start with exploring different design choices of our approach. We build on the labels discovered by the semantic parser. To learn classifiers we select the labels that appear at least 30 times, resulting in 1,263 labels. The parser additionally tells us whether the label is a verb. %
The LSTM output/hidden unit as well as memory cell have each 500 dimensions. 

\myparagraph{Robust visual classifiers.}
We first analyze our proposal to consider groups of labels to learn different classifiers and also to use different visual representations for these groups (see \secref{sec:visual_labels}). In \Tableref{tbl:valset_labels_viscls} we evaluate our generated sentences using different input features to the LSTM on the validation set of MPII-MD. In our baseline, in the top part of \Tableref{tbl:valset_labels_viscls}, we use the same visual descriptors for all labels. The PLACES feature is best with 7.10 METEOR. Combination by stacking all features (IDT + LSDA + PLACES) improves further to 7.24 METEOR. The second part of the table demonstrates the effect of introducing different semantic label groups. We first split the labels into ``Verbs'' and all others. Given that some labels appear in both roles, the total number of labels increases to 1328 (line 5). We compare two settings of training the classifiers: ``Retrieved'' (we retrieve the classifier scores from the classifiers trained in the previous step), ``Trained'' (we train the SVMs specifically for each label type, \eg ``Verbs''). Next, we further divide the non-"Verb'' labels into ``Places'' and ``Others''(line 6), and finally into ``Places'' and ``Objects''(line 7). We discard the unused labels and end up with 913 labels. Out of these labels, we select the labels where the classifier obtains a ROC higher or equal to 0.7 (threshold selected on the validation set). After this we obtain 263 labels and the best performance in the ``Trained'' setting (line 8). To support our intuition about the importance of the label discrimination (\ie using different features for different semantic groups of labels), we propose another baseline (line 9). Here we use the same set of 263 labels but provide the same feature for all of them, namely the best performing combination IDT + LSDA + PLACES. As we see, this results in an inferior performance.

We make several observations from \Tableref{tbl:valset_labels_viscls} which lead to robust visual classifiers from the weak sentence annotations. a) It is beneficial to select features based on the label semantics. b) Training one-vs-all SVMs for specific label groups consistently improves the performance as it avoids ``wrong'' negatives. c)~Focusing on more ``visual'' labels helps: we reduce the LSTM input dimensionality to 263 while improving the performance.

\newcommand{\rot}[1]{\multicolumn{1}{c}{\begin{turn}{90}\hspace{-2pt}#1\end{turn}}}

\newcommand{\midruleTestLongMVAD}{\cmidrule(rr){1-1} \cmidrule(rr){2-2}}

\newcommand{\midruleTestLong}{\cmidrule(rr){1-1} \cmidrule(rr){2-2} \cmidrule(rr){3-5}}
\begin{table*}[t]
\begin{center}
\begin{tabular}{cc}
\begin{tabular}{l@{\ \ \ }c@{\ \ \ }ccc}
\toprule
         &\scriptsize{$METEOR$}           & \multicolumn{3}{c}{Human evaluation: rank} \\
Approach  & in \% & {Correct.} & {Grammar} & {Relev.} \\
\midruleTestLong
\multicolumn{2}{l}{NN baselines}	\\
\ \ IDT	    &	4.87 &  -& -& -\\
\ \ LSDA	&	4.45 &  -& -& -\\
\ \ PLACES	&	4.28 &  -& -& -\\
\ \ HYBRID	&	4.34 &  -& -& -\\
\midruleTestLong
SMT-Best (ours) & 5.59 & 2.11 & 2.39 & 2.08 \\
S2VT & 6.27 & 2.02 & \textbf{1.67} & 2.06 \\
Visual-Labels (ours) & \textbf{7.03} & \textbf{1.87} & 1.94 & \textbf{1.86} \\
\midruleTestLong
NN upperbound                          & 19.43 & -& -& -\\
\bottomrule 
\end{tabular}
&
\begin{tabular}{l@{\ \ \ }c}
\toprule
  & \scriptsize{$METEOR$} \\
Approach& in \%\\
\midruleTestLongMVAD
Temporal attention& 4.33 \\
S2VT & 5.62 \\
Visual-Labels (ours)                            & \textbf{6.36} \\
\bottomrule 
\end{tabular}\\
\\
(a) Test Set of MPII-MD.& (b) Test Set of \MVAD.\\
\end{tabular}
\caption{Comparison of our proposed methods to prior work: S2VT  \citep{venugopalan15arxiv1505.00487v2}, Temporal attention \citep{yao2015iccv}. Human eval ranked 1 to 3, lower is better. For discussion see \secref{sec:comparison-related-work}.}
\label{tbl:testset}
\end{center}
\end{table*}

\myparagraph{LSTM architectures.}
Now, as described in \secref{sec:lstm}, we look at different LSTM architectures and training configurations. In the following we use the best performing ``Visual Labels'' approach,  \Tableref{tbl:valset_labels_viscls}, line (8).

We start with examining the architecture, where we explore different configurations of LSTM and dropout layers. \Tableref{tbl:valset_configs}(a) shows the performance of three different networks: ``1 layer'', ``2 layers unfactored'' and ``2 layers factored'' introduced in \secref{sec:lstm}. As we see, the ``1 layer'' and ``2 layers unfactored'' perform equally well, while ``2 layers factored'' is inferior to them. In the following experiments we use the simpler ``1 layer'' network. We then compare different dropout placements as illustrated in (\Tableref{tbl:valset_configs}(b)). We obtain the best result when applying dropout after the LSTM layer (``lstm-drop''), while having no dropout or applying it only to language leads to stronger over-fitting to the visual features. Putting dropout after the LSTM (and prior to a final prediction layer) makes the entire system more robust. As for the best dropout ratio, we find that 0.5 works best with lstm-dropout (\Tableref{tbl:valset_configs}(c)).

In most of the experiments we trained our networks for 25,000 iterations. After looking at the METEOR scores for intermediate iterations we found that at iteration 15,000 we achieve best performance overall. Additionally we train multiple LSTMs with different random orderings of the training data. In our experiments we combine three in an ensemble, averaging the resulting word predictions. %

To summarize, the most important aspects that decrease over-fitting and lead to better sentence generation are: (a) a correct learning rate and step size, (b) dropout after the LSTM layer, (c) choosing the training iteration based on METEOR score as opposed to only looking at the LSTM accuracy/loss which can be misleading, and (d) building ensembles of multiple networks with different random initializations\footnote{\label{fn:supplemental}More details can be found in our corresponding arXiv version \cite{rohrbach15arxiv}}. %

\subsubsection{Comparison to related work}
\label{sec:comparison-related-work}
\myparagraph{Experimental setup.}
In this section we evaluate on the test set of the MPII-MD dataset (6,578 clips) and  \MVAD dataset (4,951 clips). We use METEOR for automatic evaluation and  we perform a human evaluation on a random subset of 1,300 video clips, see \secref{sec:eval-metrics} for details.  %
We train our method on \MVAD and use the same LSTM architecture and parameters as for MPII-MD, but select the number of iterations on the \MVAD validation set.

\begin{figure*}[t]

\center
\begin{tabular}{l@{\ \ \ }l@{\ \ \ }l}
\toprule
& Approach &  Sentence\\
\midrule
\multirow{6}{*}{\includegraphics[width=3cm]{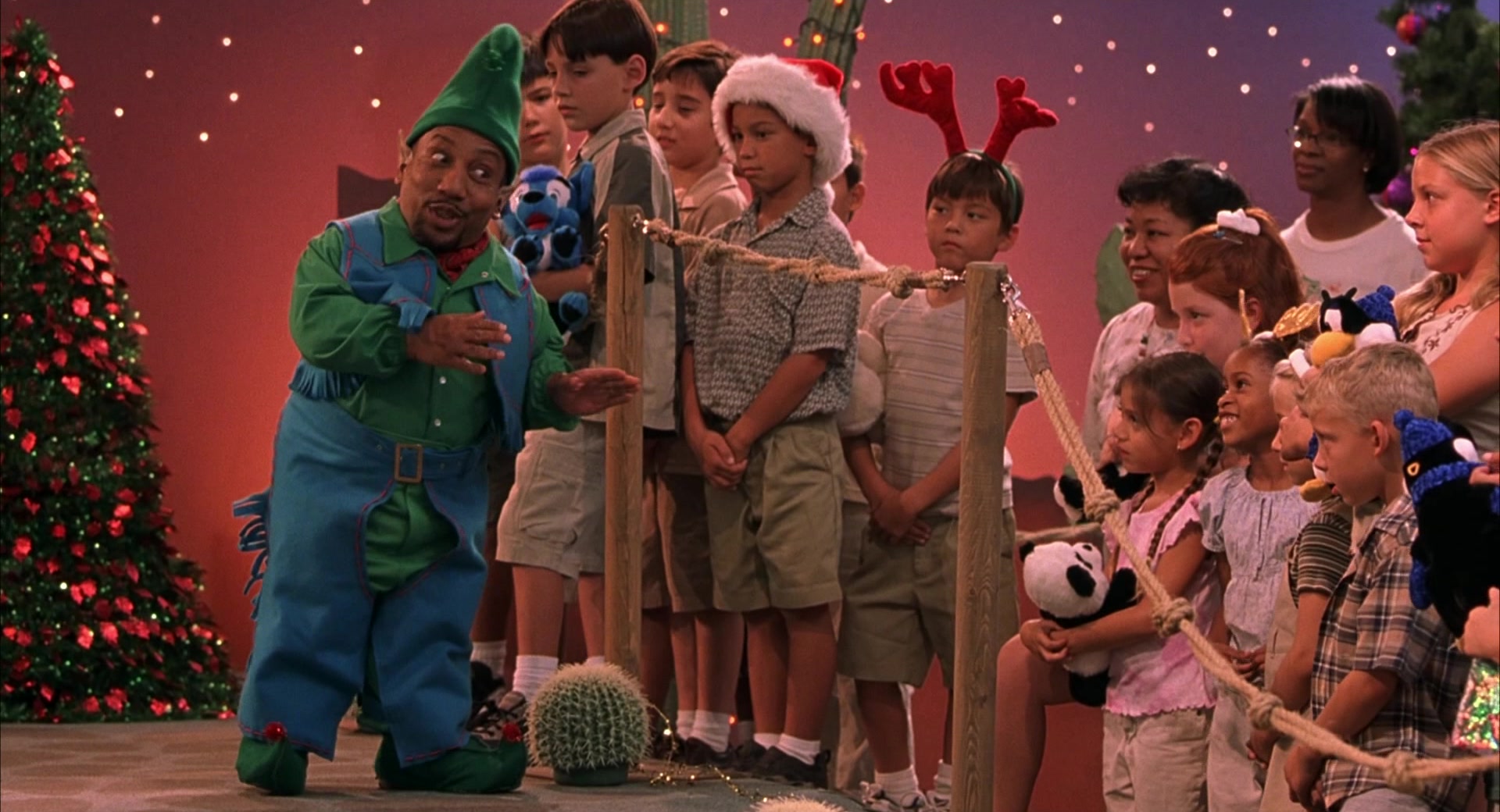}} & SMT-Best (ours) & Someone is a man, someone is a man. \\ 
 &S2VT & Someone looks at him, someone turns to someone.\\ 
 &Visual-Labels (ours) & Someone is standing in the crowd, \\
 &&a little man with a little smile. \\
 & Reference & Someone, back in elf guise, is trying to calm the kids. \\
 \\
\multirow{6}{*}{\includegraphics[width=3cm]{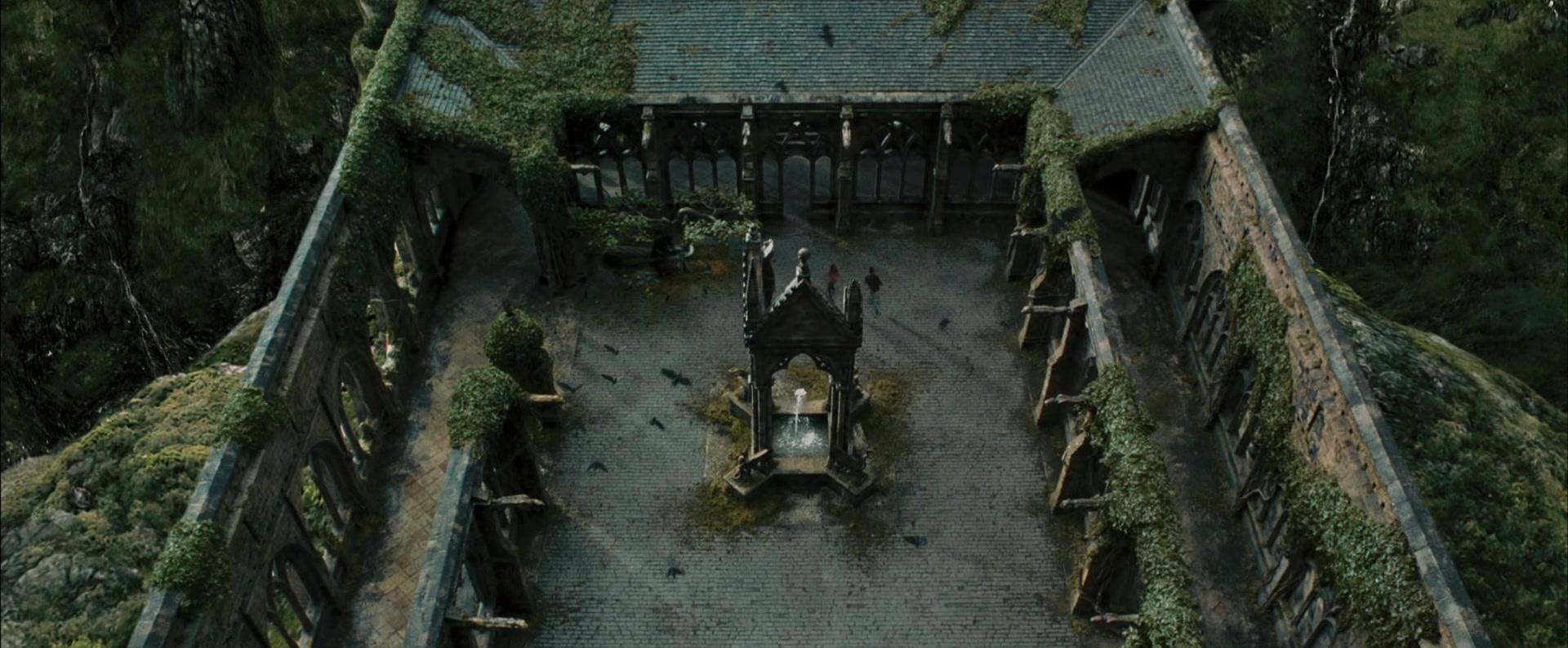}} &  \\
& SMT-Best (ours) &The car is a water of the water. \\
 &S2VT &  On the door, opens the door opens. \\
  &Visual-Labels (ours) &  The fellowship are in the courtyard. \\
  &Reference & They cross the quadrangle below and run along the cloister. \\
  \\
\multirow{6}{*}{\includegraphics[width=3cm]{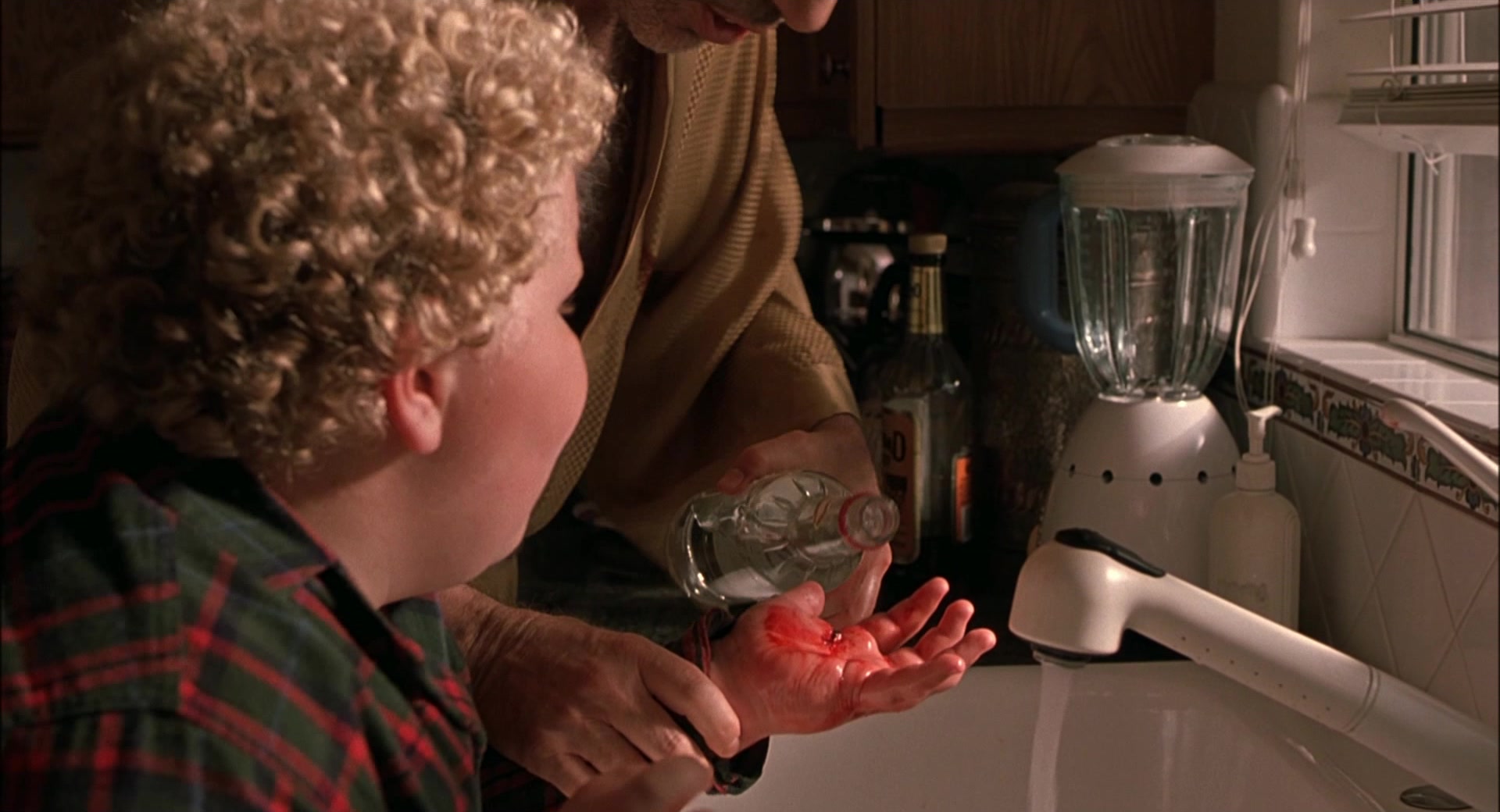}}& SMT-Best (ours) & Someone is down the door, \\
&&someone is a back of the door, and someone is a door. \\
 &S2VT &   Someone shakes his head and looks at someone.\\ 
 &Visual-Labels (ours) & Someone takes a drink and pours it into the water. \\
 &Reference& Someone grabs a vodka bottle standing open on the counter \\
 && and liberally pours some on the hand. \\
\bottomrule
\end{tabular}
\caption{Qualitative comparison of our proposed methods to prior work: S2VT \citep{venugopalan15arxiv1505.00487v2}. Examples from the test set of MPII-MD. Visual-Labels identifies activities, objects, and places better than the other two methods. See \secref{sec:results:mpiimd}.} 
\label{fig:qual_mpiimd}
\end{figure*}

\myparagraph{Results on MPII-MD.}
\label{sec:results:mpiimd}
\Tableref{tbl:testset}(a) summarizes the results on the test set of MPII-MD. 
Here we additionally include the results from a nearest neighbor baseline, \ie we retrieve the closest sentence from the training corpus using L1-normalized visual features and the intersection distance. When comparing the different features (introduced in Section \ref{subsec:visual_features}), we see that the pre-trained features (LSDA, PLACES, HYBRID) perform better than IDT, with HYBRID performing best. Our SMT-Best approach clearly improves over the nearest neighbor baselines. With our Visual-Labels approach we again significantly improve the performance, specifically by 1.44 METEOR points. Moreover, we improve over the recent approach of \citep{venugopalan15arxiv1505.00487v2}, which also uses an LSTM to generate video descriptions. Exploring different strategies to label selection and classifier training, as well as various LSTM configurations allows to obtain better result than prior on the MPII-MD dataset. Human evaluation mainly agrees with the automatic measure. Visual-Labels outperforms both other methods in terms of Correctness and Relevance, however it loses to S2VT in terms of Grammar. This is due to the fact that S2VT produces overall shorter (7.4 versus 8.7 words per sentence) and simpler sentences, while our system generates longer sentences and therefore has higher chances to make mistakes. We also propose a retrieval upperbound. For every test sentence we retrieve the closest training sentence according to the METEOR score. The rather low METEOR score of 19.43 reflects the difficulty of the dataset. We show some qualitative results in \Figref{fig:qual_mpiimd}.

\myparagraph{Results on \MVAD.}

\begin{figure*}[t]
\begin{center}
\includegraphics[width=\linewidth]{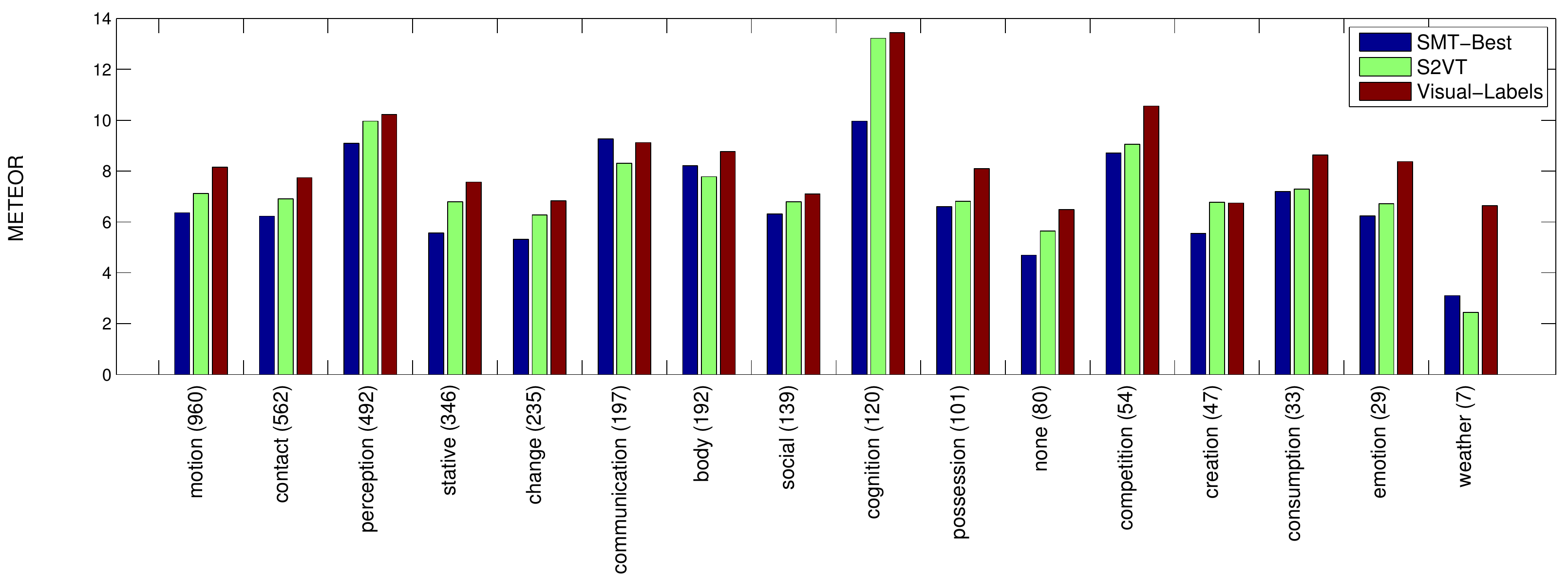}
\caption{Average METEOR score for WordNet verb Topics. Selected sentences with single verb, number of sentences in brackets. For discussion see \secref{sec:wordnetverbtopics}.}
\label{fig:topics}
\end{center}
\end{figure*}

\Tableref{tbl:testset}(b) shows the results on the test set of \MVAD dataset. Our Visual-Labels method outperforms S2VT \citep{venugopalan15arxiv1505.00487v2} and Temporal attention \citep{yao2015iccv} in METEOR score. As we see, the results agree with \tableref{tbl:testset}(a), but are consistently lower, suggesting that \MVAD is more challenging than MPII-MD. We attribute this to more precise manual alignments of the MPII-MD dataset.

\subsection{Movie description analysis}
\label{sec:analysis}

Despite the recent advances in the video description task, the performance on the movie description datasets (MPII-MD  and \MVAD) remains rather low. In this section we want to look closer at three methods, SMT-Best, S2VT and Visual-Labels, in order to understand where these methods succeed and where they fail. In the following we evaluate all three methods on the MPII-MD test set.

\begin{figure}
\center
\begin{tabular}{@{}p{4cm}p{4cm}}
\includegraphics[width=\linewidth]{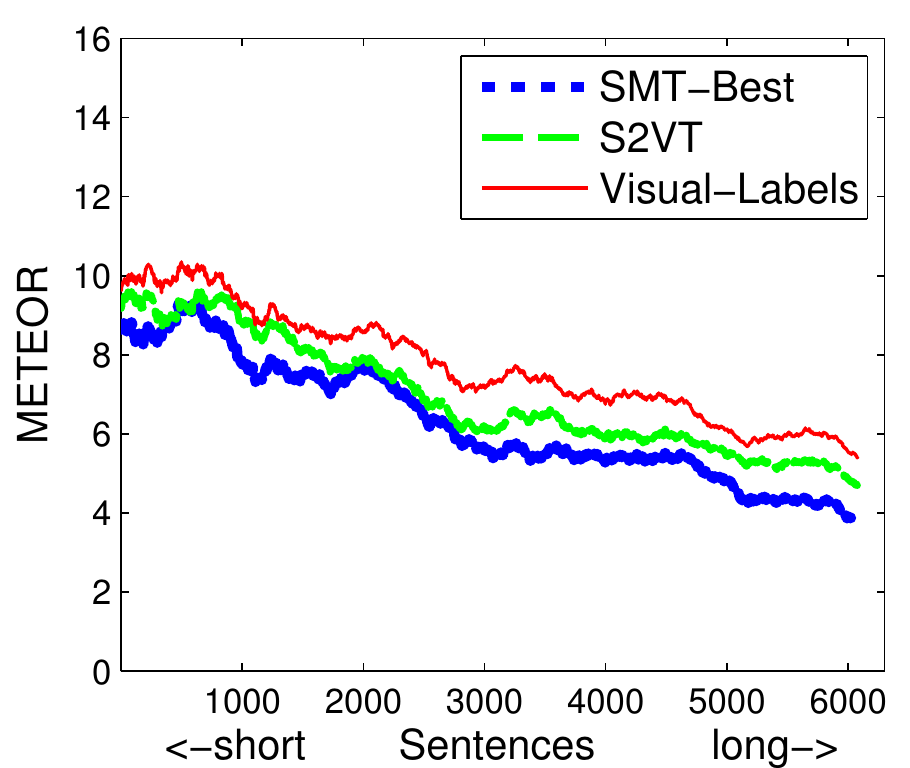}&
\includegraphics[width=\linewidth]{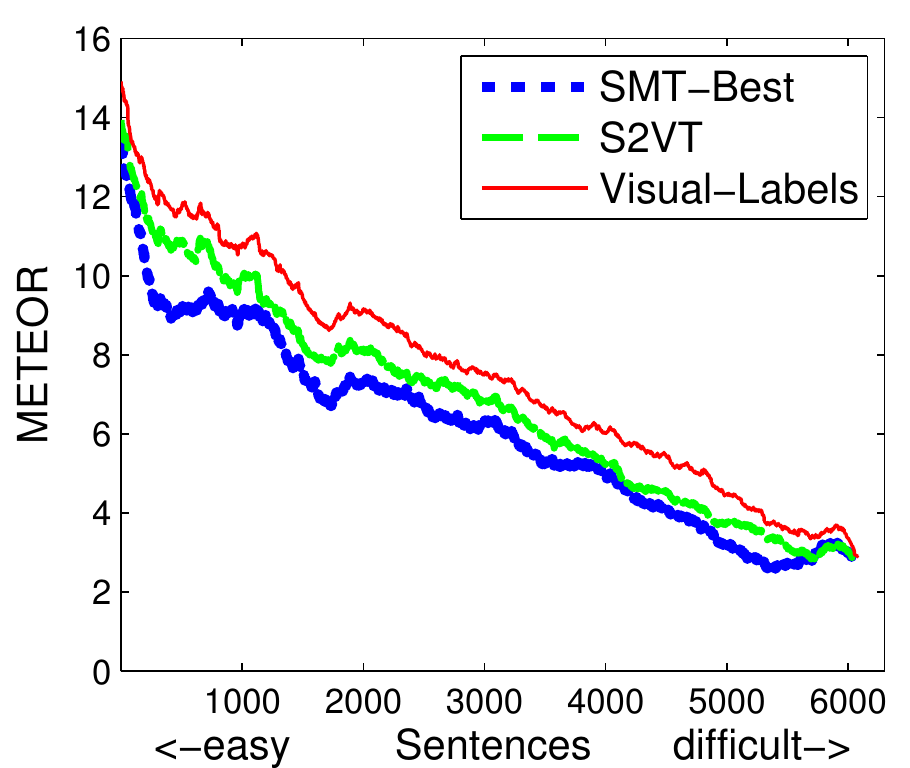}\\
(a) Sentence length
&(b) Word frequency\\
\end{tabular}
\caption{Y-axis: METEOR score per sentence. X-axis: MPII-MD test sentences 1 to 6,578 sorted by (a) length (increasing); (b) word frequency (decreasing). Shown values are smoothed with a mean filter of size 500. For discussion see \secref{sec:analysis:diffVsPerformance}.}
\label{fig:eval:difficulty}
\end{figure}

\subsubsection{Difficulty versus performance}
\label{sec:analysis:diffVsPerformance}
As the first study we suggest to sort the test reference sentences by difficulty, where difficulty is defined in multiple ways\textsuperscript{\ref{fn:supplemental}}.

\myparagraph{Sentence length and Word frequency.}
Some of the intuitive sentence difficulty measures are its length and average frequency of its words. When sorting the data by difficulty (increasing sentence length or decreasing average word frequency), we find that all three methods have the same tendency to obtain lower METEOR score as the difficulty increases. \Figref{fig:eval:difficulty}(a) shows the performance of compared methods w.r.t. the sentence length. For the word frequency the correlation is even stronger, see \Figref{fig:eval:difficulty}(b). Visual-Labels consistently outperforms the other two methods, most notable as the difficulty increases.

\subsubsection{Semantic analysis}

\begin{table}
\begin{center}
\scriptsize
\begin{tabular}{lllll}
Topic & Entropy & Top-1 & Top-2 & Top-3 \\
\toprule
\textbf{motion} & 7.05 & turn & walk & shake \\
\textbf{contact} & 7.10 & open & sit & stand \\
\textbf{perception} & 4.83 & look & stare & see \\
\textbf{stative} & 4.84 & be & follow & stop \\
\textbf{change} & 6.92 & reveal & start & emerge \\
\textbf{communication} & 6.73 & look up & nod & face \\
\textbf{body} & 5.04 & smile & wear & dress \\
\textbf{social} & 6.11 & watch & join & do  \\
\textbf{cognition} & 5.21 & look at & see & read \\
\textbf{possession} & 5.29 & give & take & have \\
\textbf{none} & 5.04 & throw & hold & fly \\
\textbf{creation} & 5.69 & hit & make & do \\
\textbf{competition} & 5.19 & drive & walk over & point \\
\textbf{consumption} & 4.52 & use & drink & eat \\
\textbf{emotion} & 6.19 & draw & startle & feel \\
\textbf{weather} & 3.93 & shine & blaze & light up \\
\bottomrule
\end{tabular}
\end{center}
\caption{Entropy and top 3 frequent verbs of each WordNet topic. For discussion see \secref{sec:wordnetverbtopics}.}
\label{tbl:topic_words}
\end{table}

\begin{table*}[t]
\begin{center}
\begin{tabular}{lrrrrrrr}
\toprule
 & \multicolumn{4}{c}{BLEU} & \multicolumn{1}{c}{METEOR} & \multicolumn{1}{c}{ROUGE} & \multicolumn{1}{c}{CIDEr} \\
Approach & 1 & 2 & 3 & 4 & \multicolumn{1}{c}{} & \multicolumn{1}{c}{} & \multicolumn{1}{c}{} \\
\midrule
Visual-Labels (ours) &	16.1 & 5.2 & \textbf{2.1} & \textbf{0.9} & \textbf{7.1} & \textbf{16.4} & \textbf{11.2} \\
S2VT \citep{venugopalan15iccv} & \textbf{17.4} & \textbf{5.3} & 1.8 & 0.7 & 7.0 & 16.1 & 9.1 \\
Frame-Video-Concept Fusion \citep{shetty15arxiv} & 11.0 & 3.4 & 1.3 & 0.6 & 6.1 & 15.6 & 9.0 \\
Temporal Attention \citep{yao2015iccv} & 5.6 & 1.5 & 0.6 & 0.3 & 5.2 & 13.4 & 6.2 \\
\bottomrule
\end{tabular}
\caption{Automatic evaluation on the blind test set of the LSMDC, in \%. For discussion see \secref{sec:lsmdc:qualresults}.}
\label{tbl:lsmdc_automatic}
\end{center}
\end{table*}

\begin{table*}[t]
\begin{center}
\begin{tabular}{lrrrr}
\toprule
Approach & Correctness & Grammar & Relevance & Helpful for blind \\
\midrule
Visual-Labels (ours) & 3.32 & 3.37 & 3.32 & \textbf{3.26} \\
S2VT \citep{venugopalan15iccv} & 3.55 & 3.09 & 3.53 & 3.42 \\
Frame-Video-Concept Fusion \citep{shetty15arxiv} & \textbf{3.10} & \textbf{2.70} & \textbf{3.29} & 3.29 \\
Temporal Attention \citep{yao2015iccv} & 3.14 & 2.71 & 3.31 & 3.36 \\
Reference & 1.88 & 3.13 & 1.56 & 1.57 \\
\bottomrule
\end{tabular}
\caption{Human evaluation on the blind test set of the LSMDC. Human eval ranked 1 to 5, lower is better. For discussion see \secref{sec:lsmdc:qualresults}.}
\label{tbl:lsmdc_human}
\end{center}
\end{table*}

\begin{table*}[t]
\begin{center}
\begin{tabular}{lrrrr}
\toprule
Approach & Avg. sentence & Vocabulary & Number of        &  \% Novel \\
         & length        & size       & unique sentences & sentences \\
\midrule
Visual-Labels (ours) & 7.47 & 525 & 4,320 & 67.63 \\
S2VT \citep{venugopalan15iccv} & 8.77 & 663 & 2,890 & 72.61 \\
Frame-Video-Concept Fusion \citep{shetty15arxiv} & 5.16 & 401 & 871 & 33.62 \\
Temporal Attention \citep{yao2015iccv} & 3.63 & 117 & 133 & 7.36 \\
Reference & 8.74 & 6,303 & 9,309 & 93.49 \\
\bottomrule
\end{tabular}
\caption{Description statistics for different methods and reference sentences on the blind test set of the LSMDC. For discussion see \secref{sec:lsmdc:qualresults}.}
\label{tbl:lsmdc_stat}
\end{center}
\end{table*}

\myparagraph{WordNet Verb Topics.}
\label{sec:wordnetverbtopics}
Next we analyze the test reference sentences w.r.t. verb semantics. We rely on WordNet Topics (high level entries in the WordNet ontology), e.g. ``motion'', ``perception'', defined for most synsets in WordNet \citep{Fellbaum1998}. Sense information comes from our automatic semantic parser, thus it might be noisy. We showcase the 3 most frequent verbs for each Topic in \Tableref{tbl:topic_words}. We select sentences with a single verb, group them according to the verb Topic and compute an average METEOR score for each Topic, see \Figref{fig:topics}. We find that Visual-Labels is best for all Topics except ``communication", where SMT-Best wins. The most frequent verbs there are ``look up'' and ``nod'', which are also frequent in the dataset and in the sentences produced by SMT-Best. The best performing Topic, ``cognition'', is highly biased to ``look at'' verb. The most frequent Topics, ``motion'' and ``contact'', which are also visual (e.g. ``turn'', ``walk'', ``sit''), are nevertheless quite challenging, which we attribute to their high diversity (see their entropy w.r.t. different verbs and their frequencies in \Tableref{tbl:topic_words}). Topics with more abstract verbs (e.g. ``be'', ``have'', ``start'') get lower scores.

\myparagraph{Top 100 best and worst sentences.}
We look at 100 test reference sentences, where Visual-Labels obtains highest and lowest METEOR scores. Out of 100 best sentences 44 contain the verb ``look'' (including phrases such as ``look at''). The other frequent verbs are ``walk'', ``turn", ``smile'', ``nod'', ``shake'', i.e. mainly visual verbs. Overall the sentences are simple. Among the worst 100 sentences we observe more diversity: 12 contain no verb, 10 mention unusual words (specific to the movie), 24 have no subject, 29 have a non-human subject. This leads to a lower performance, in particular, as most training sentences contain ``Someone'' as subject and generated sentences are biased towards it.

\myparagraph{Summary.} a) The test reference sentences that mention verbs like ``look'' get higher scores due to their high frequency in the dataset. b) The sentences with more ``visual'' verbs tend to get higher scores. c) The sentences without verbs (e.g. describing a scene), without subjects or with non-human subjects get lower scores, which can be explained by dataset biases.

\section{The \LaScMoDeCh 2015}
\label{sec:eval-lsmdc}

\begin{figure*}[t]
\scriptsize
\center
\begin{tabular}{l@{\ \ \ }l@{\ \ \ }l}
\toprule
& Approach &  Sentence\\
\midrule
\multirow{5}{*}{\includegraphics[width=3cm]{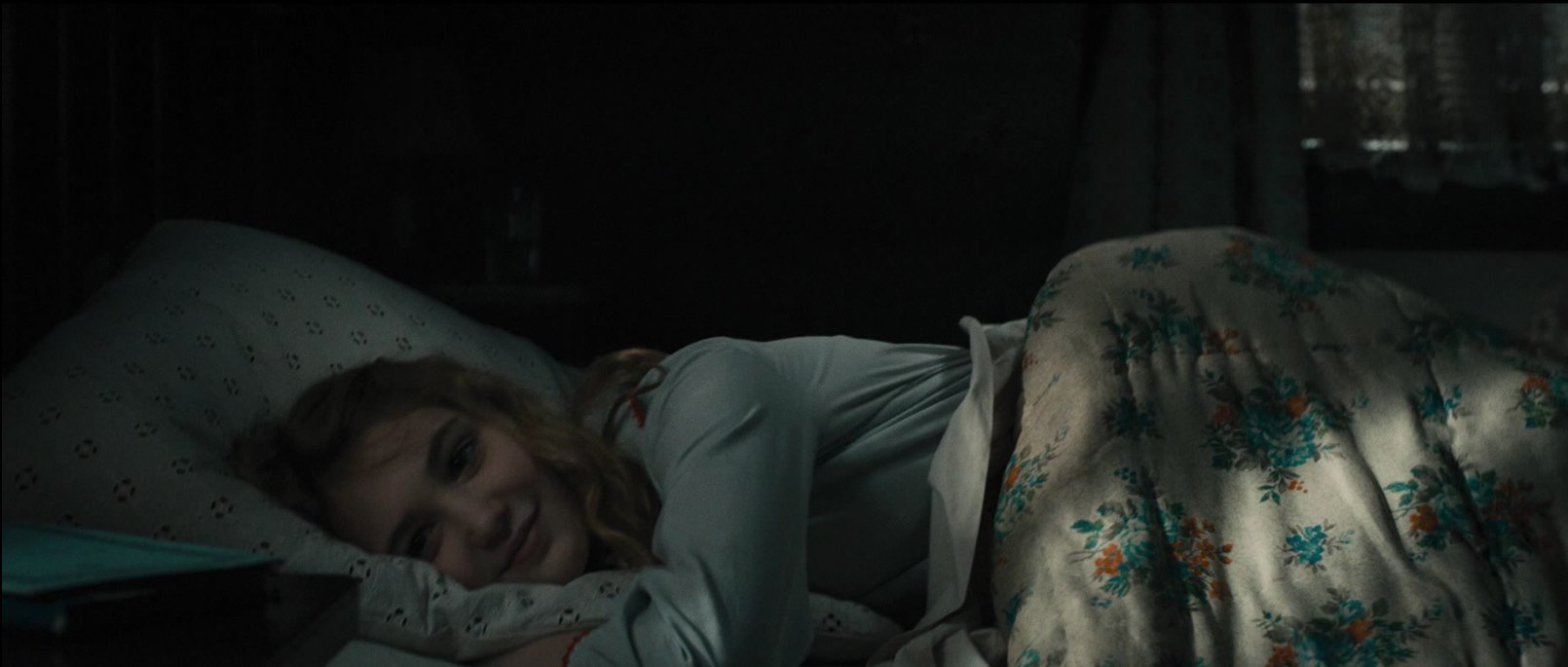}} & Visual-Labels (ours) & Someone lies on the bed. \\ 
 &S2VT & Someone lies asleep on his bed. \\ 
 &Frame-Video-Concept Fusion & Someone lies on the bed. \\
 &Temporal Attention & Someone lies in bed. \\
 & Reference & Someone lies on her side facing her new friend. \\
 \\
 \multirow{5}{*}{\includegraphics[width=3cm]{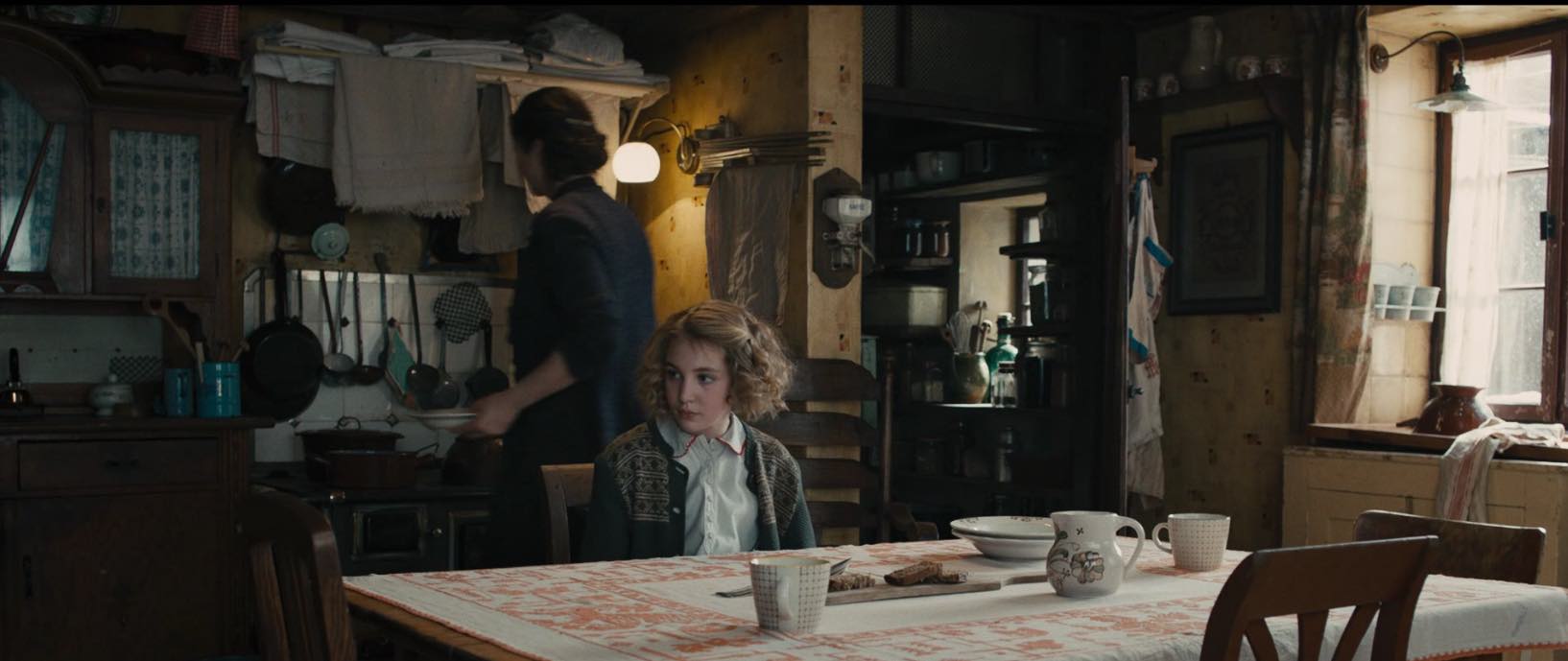}} & Visual-Labels (ours) & Someone sits down. \\ 
 &S2VT & Someone sits on the couch and looks at the tv. \\ 
 &Frame-Video-Concept Fusion & Someone sits at the table. \\
 &Temporal Attention & Someone looks at someone. \\
 & Reference & Someone takes a seat and someone moves to the stove. \\
 \\
 \multirow{5}{*}{\includegraphics[width=3cm]{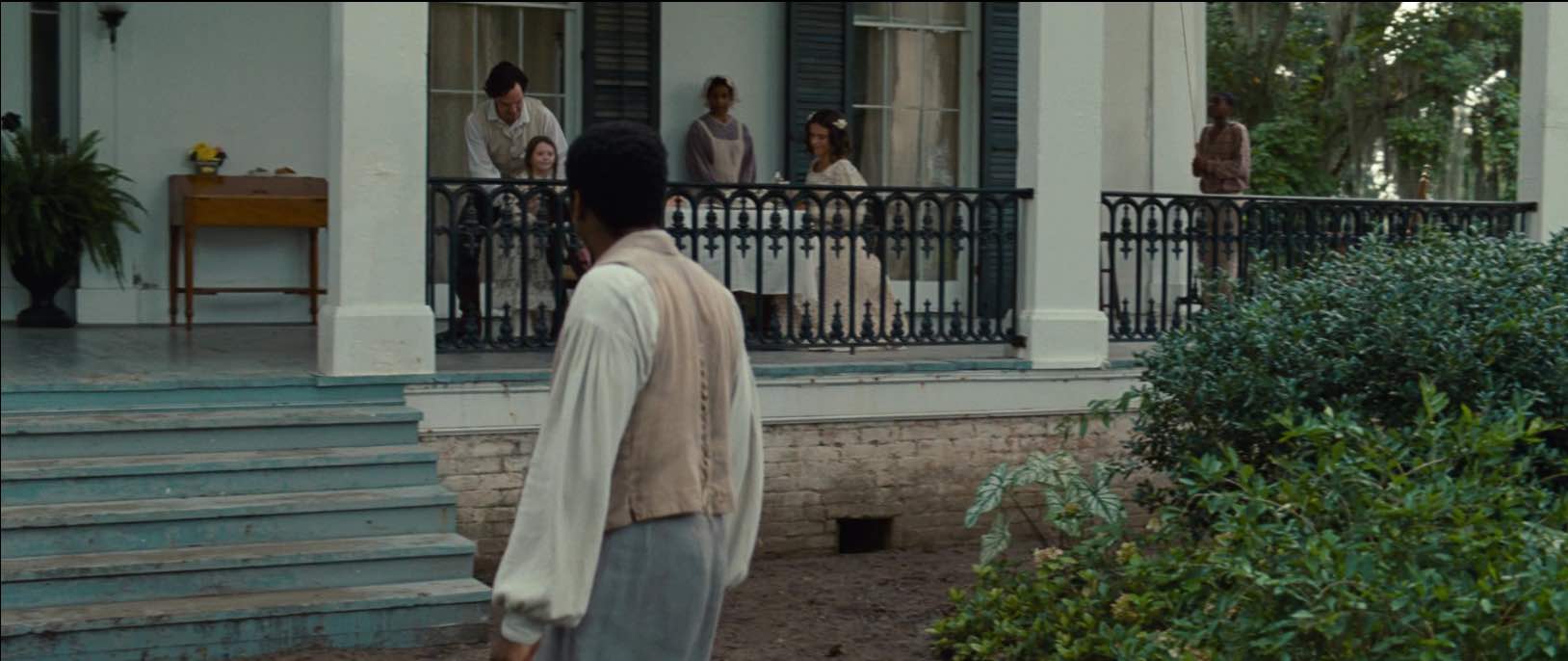}} & Visual-Labels (ours) & Someone walks to the front of the house. \\ 
 &S2VT & Someone looks at the house. \\ 
 &Frame-Video-Concept Fusion & Someone walks up to the house. \\
 &Temporal Attention & Someone looks at someone. \\
 & Reference & Someone sets down his young daughter then moves to a small wooden table. \\
 \\
 \multirow{6}{*}{\includegraphics[width=3cm]{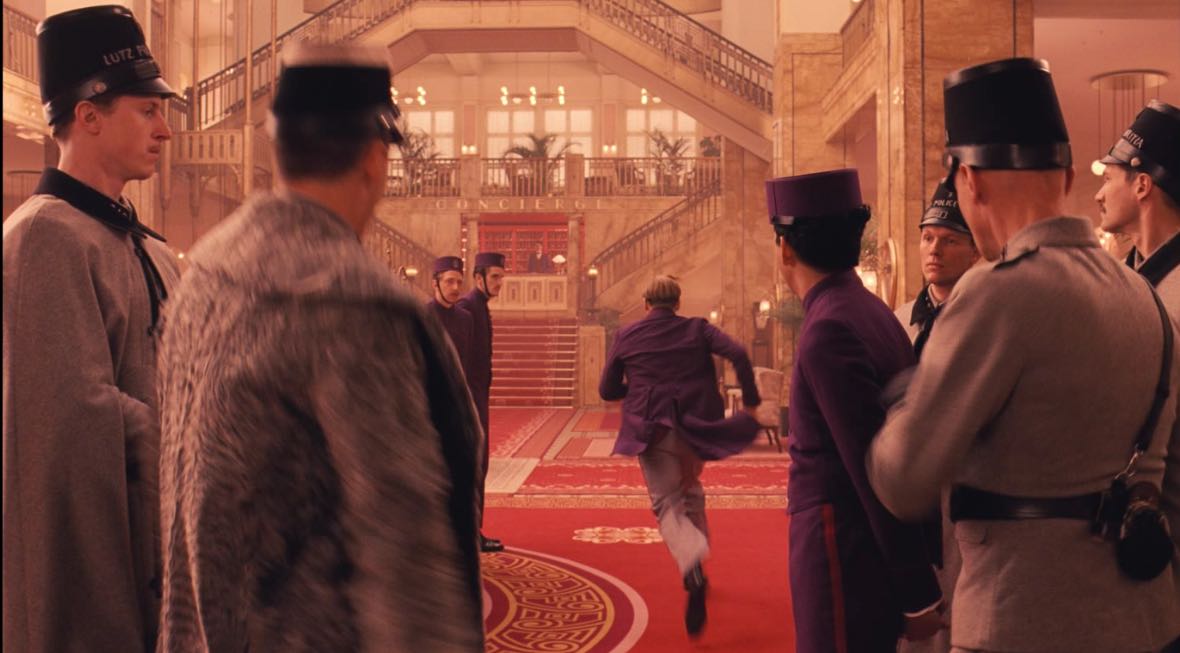}} & Visual-Labels (ours) & Someone turns to someone. \\ 
 &S2VT & Someone looks at someone. \\ 
 &Frame-Video-Concept Fusion & Someone turns to someone. \\
 &Temporal Attention & Someone stands alone. \\
 & Reference & Someone dashes for the staircase. \\
 \\
 \\
  \multirow{5}{*}{\includegraphics[width=3cm]{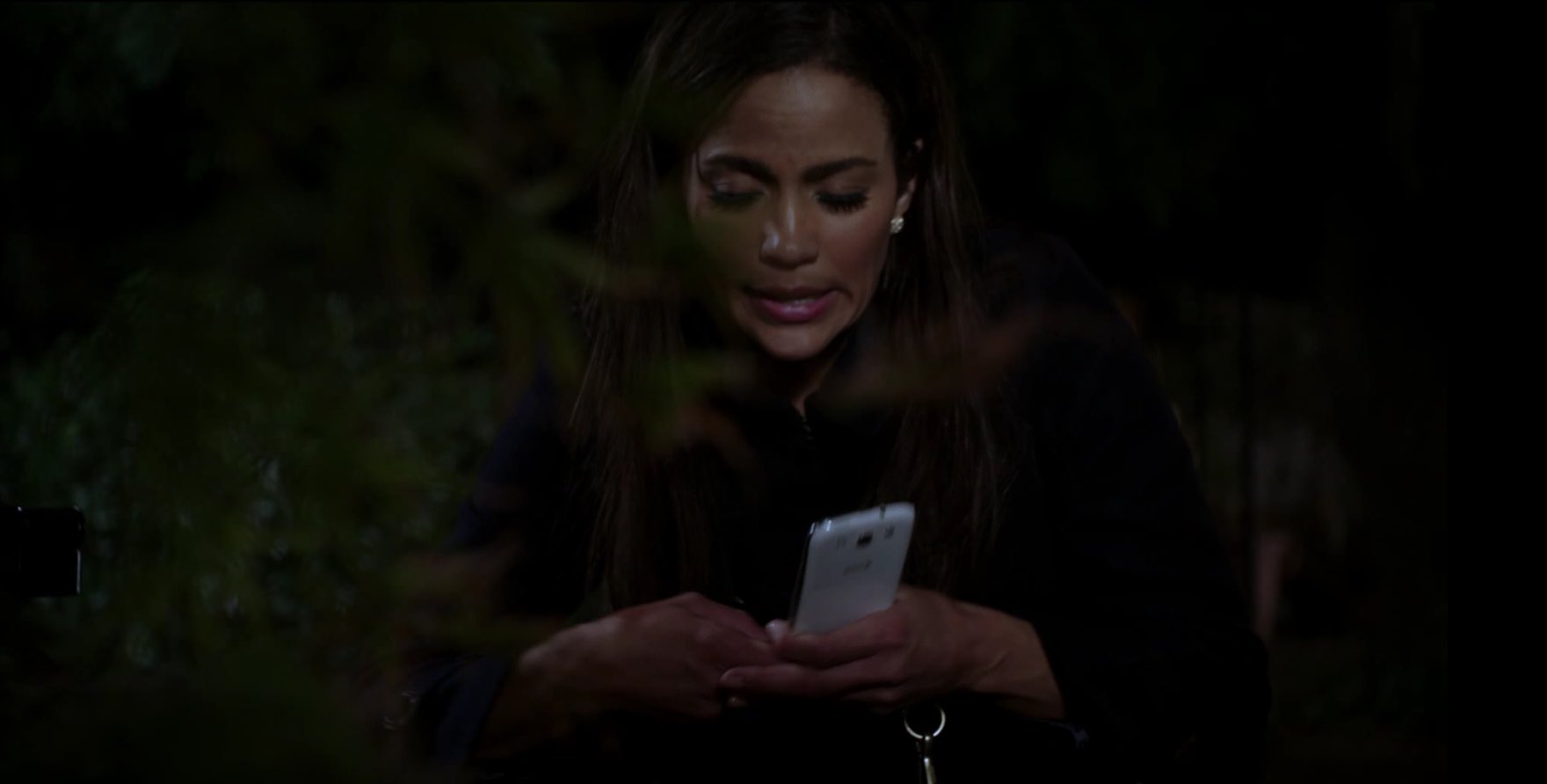}} & Visual-Labels (ours) & Someone takes a deep breath and takes a deep breath. \\ 
 &S2VT  & Someone looks at someone and looks at him. \\ 
 &Frame-Video-Concept Fusion  & Someone looks up at the ceiling. \\
 &Temporal Attention & Someone stares at someone. \\
 & Reference & Someone digs out her phone again, eyes the display, and answers the call. \\
 \\
\bottomrule
\end{tabular}
\caption{Qualitative comparison of our approach Visual-Labels, S2VT \citep{venugopalan15iccv}, Frame-Video-Concept Fusion \citep{shetty15arxiv} and Temporal Attention \citep{yao2015iccv} on the blind test set of the LSMDC. Discussion see \secref{sec:lsmdc:qual-results}.}
\label{fig:qual}
\end{figure*}

The Large Scale Movie Description Challenge (LSMDC) was held in conjunction with ICCV 2015. For the automatic evaluation we set up an evaluation server\textsuperscript{\ref{fn:codalab}}. During the first phase of the challenge the participants could evaluate the outputs of their system on the public test set of LSMDC dataset. In the second phase of the challenge the participants were provided with the videos from the blind test set (without textual descriptions). These were used for the final evaluation. To measure performance of the competing approaches we performed both automatic and human evaluation as described in \secref{sec:eval-metrics}. The submission format was similar to the MS COCO Challenge \citep{chen15arxiv1504.00325} and we also used the identical automatic evaluation protocol. 
 The challenge winner was determined based on the human evaluation. The human evaluation was performed over 1,200 randomly selected clips from the blind test set of LSMDC.

\begin{figure*}[t]
\scriptsize
\center
\begin{tabular}{l@{\ \ \ }l@{\ \ \ }l}
\toprule
& Approach &  Sentence\\
\midrule
\multirow{5}{*}{\includegraphics[width=3cm]{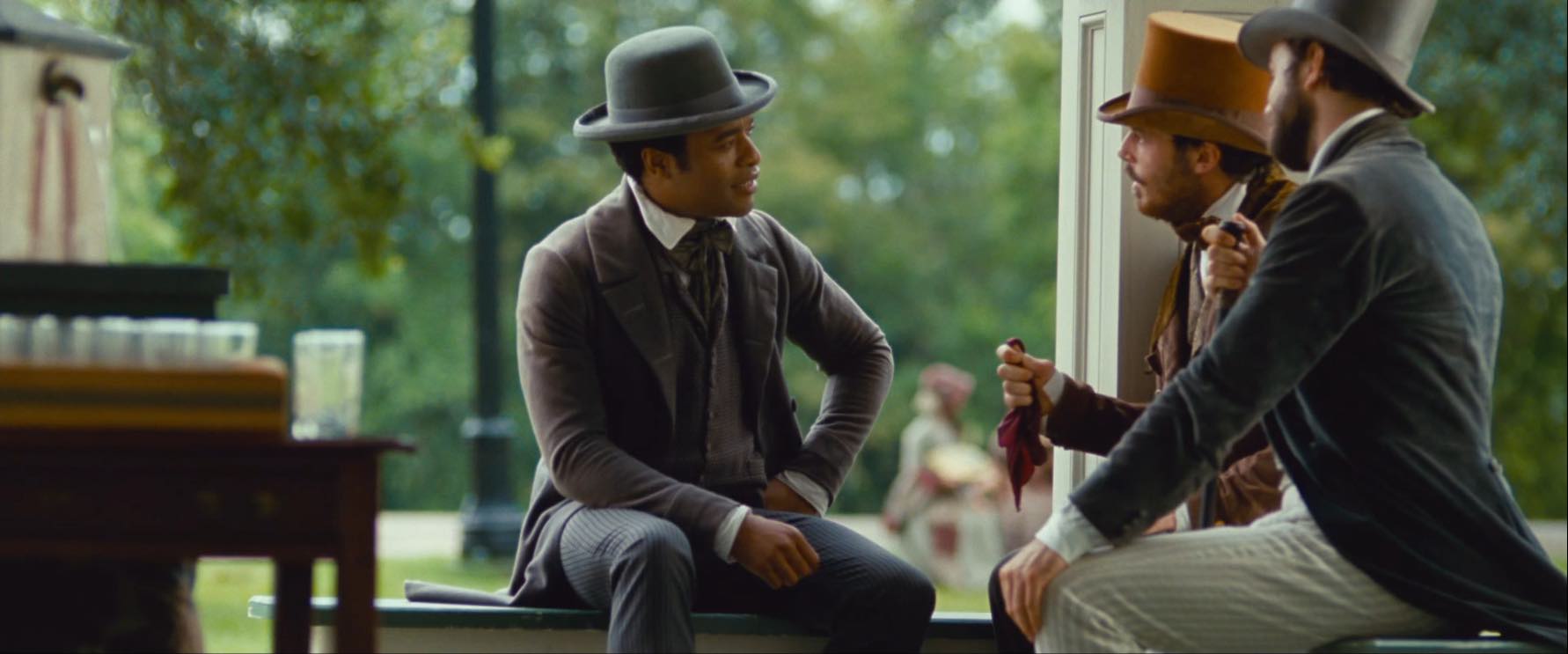}} & Visual-Labels (ours) & Someone takes a seat on the table and takes a seat on his desk. \\ 
 &S2VT & Someone looks at someone and smiles. \\ 
 &Frame-Video-Concept Fusion & Someone looks at someone. \\
 &Temporal Attention & Someone gets up. \\
 & Reference & Later, someone sits with someone and someone. \\
 \\
 \multirow{5}{*}{\includegraphics[width=3cm]{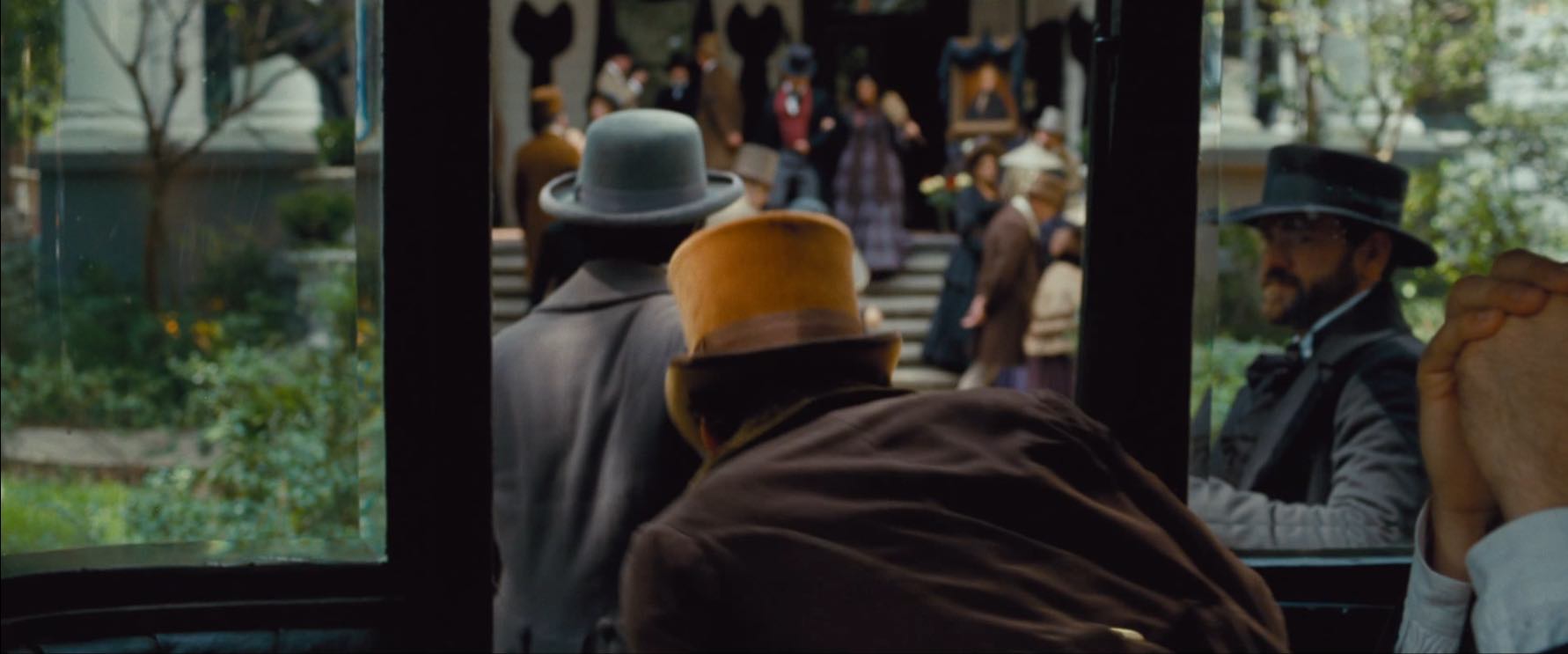}} & Visual-Labels (ours) & Someone gets out of the car and walks off. \\ 
 &S2VT & Someone walks up to the front of the house. \\ 
 &Frame-Video-Concept Fusion & Someone walks up to the front door. \\
 &Temporal Attention & Someone gets out of the car. \\
 & Reference & Now someone steps out of the carriage with his new employers. \\
 \\
 \multirow{5}{*}{\includegraphics[width=3cm]{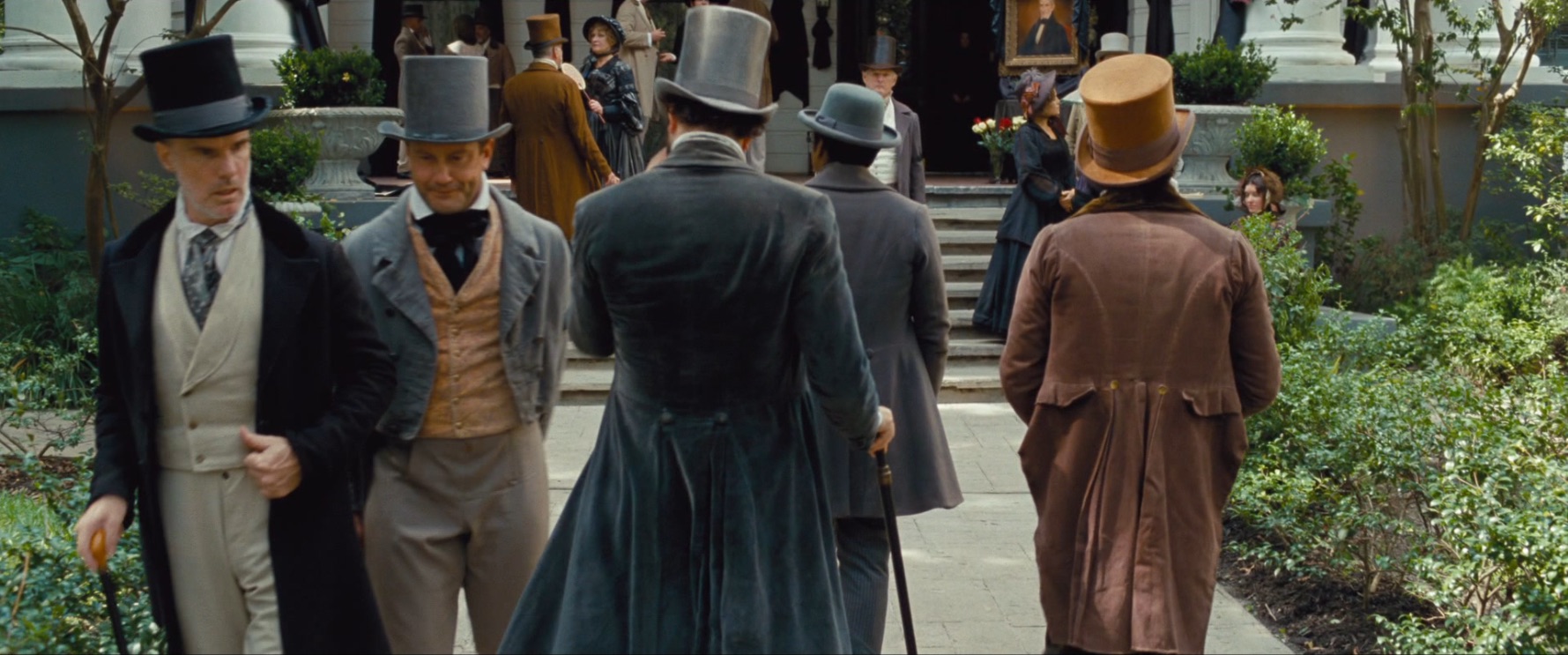}} & Visual-Labels (ours) & Someone walks up to the street, and someone is walking to the other side of. \\ 
 &S2VT & Someone walks over to the table and looks at the other side of the house. \\ 
 &Frame-Video-Concept Fusion & Someone walks away. \\
 &Temporal Attention & Someone gets out of the car. \\
 & Reference & The trio starts across a bustling courtyard. \\
 \\
 \multirow{5}{*}{\includegraphics[width=3cm]{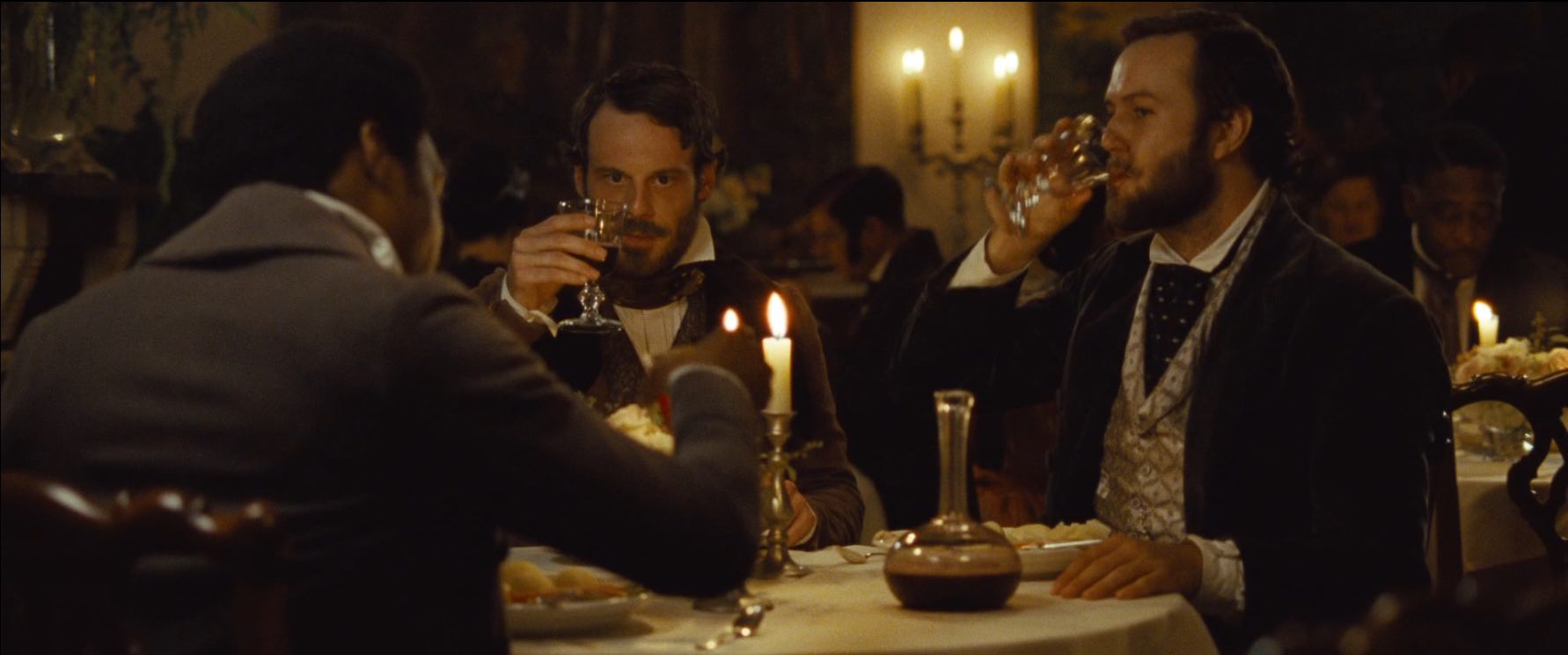}} & Visual-Labels (ours) & Someone sips his drink. \\ 
 &S2VT & Someone sits at the table and looks at someone. \\ 
 &Frame-Video-Concept Fusion & Someone sits up. \\
 &Temporal Attention & Someone looks at someone. \\
 & Reference & As the men drink red wine, someone and someone watch someone take a sip. \\
 \\
  \multirow{5}{*}{\includegraphics[width=3cm]{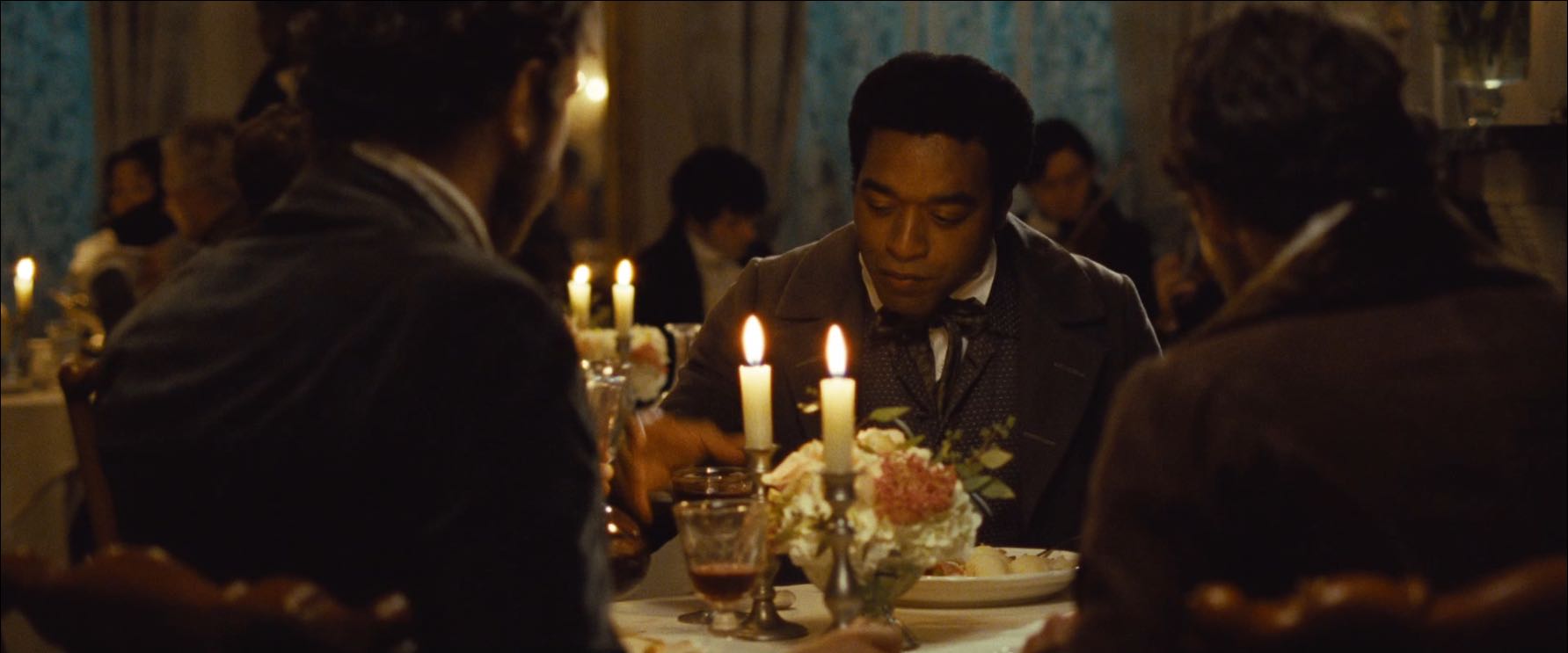}} & Visual-Labels (ours) & Someone takes a bite. \\ 
 &S2VT & Someone sits at the table. \\ 
 &Frame-Video-Concept Fusion & Someone looks at someone. \\
 &Temporal Attention & Someone looks at someone. \\
 & Reference & Someone tops off someone's glass. \\
 \\
\bottomrule
\end{tabular}
\caption{Qualitative comparison of our approach Visual-Labels, S2VT \citep{venugopalan15iccv}, Frame-Video-Concept Fusion \citep{shetty15arxiv} and Temporal Attention \citep{yao2015iccv} on 5 consecutive clips from the blind test set of the LSMDC. Discussion see \secref{sec:lsmdc:qual-results}.}
\label{fig:qual_sequence}
\end{figure*}

\subsection{\LSMDC quantitative results}
\label{sec:lsmdc:qualresults}
We received 4 submissions to the challenge, including our Visual-Labels approach. The other submissions are S2VT \citep{venugopalan15iccv}, Temporal Attention \citep{yao2015iccv} and Frame-Video-Concept Fusion \citep{shetty15arxiv}. We first look at the results of the automatic evaluation on the blind test set of LSMDC in Table \ref{tbl:lsmdc_automatic}. As we can see the Visual-Labels approach obtains highest scores in all evaluation measures except BLEU-1,-2 where S2VT gets highest score. While Visual-Labels get 7.1 METEOR and S2VT 7.0 METEOR, Frame-Video-Concept Fusion drops to 6.1 METEOR and Temporal Attention to 5.2 METEOR. One reason for this drop in the automatic measures is the sentence length, which is much smaller for Frame-Video-Concept Fusion and Temporal Attention compared to the reference sentences, as we discuss in more details below (see also  \tableref{tbl:lsmdc_stat}). %

Next, we look at the results of human evaluation in Table \ref{tbl:lsmdc_human}. As known from literature \citep{chen15arxiv1504.00325,elliott2013image,vedantam2014cider}, automatic evaluation measures do not always agree with the human evaluation. Here we see that human judges prefer the descriptions from Frame-Video-Concept Fusion approach in terms of correctness, grammar and relevance. In our alternative evaluation, in terms of being helpful for the blind, Visual-Labels wins. Possible explanation for it is that in this evaluation criteria human judges penalized less the errors in the descriptions but rather looked at their overall informativeness. In general, the gap between different approaches is not large. Based on the human evaluation the winner of the challenge is Frame-Video-Concept Fusion approach of \cite{shetty15arxiv}.

We closer analyze the outputs of the compared approaches in Table \ref{tbl:lsmdc_stat}, providing detailed statistics over the generated descriptions. With respect to the sentence length, \ApproachVisualLabels and S2VT demonstrate similar properties to the reference descriptions, while the approaches Frame-Video-Concept Fusion and Temporal Attention generate much shorter sentences (5.16 and 3.63 words on average vs. 8.74 of the references). In terms of vocabulary size all approaches fall far below the reference descriptions. This large gap indicates a problem in that all the compared approaches focus on a rather small set of visual and language concepts, ignoring a long tail in the distribution. The number of unique sentences confirms the previous finding, showing slightly higher numbers for \ApproachVisualLabels and S2VT, while the other two tend to frequently generate the same description for different clips. Finally, the percentage of novel sentences (not present among the training descriptions) highlights another aspect, namely the amount of novel vs. retrieved descriptions. As we see, all the methods ``retrieve'' some amount of descriptions from training data, while the approach Temporal Attention produces only 7.36\% novel sentences.

\subsection{\LSMDC qualitative results}
\label{sec:lsmdc:qual-results}
Figure \ref{fig:qual} shows qualitative results from the competing approaches. The first two examples are success cases, where most of the approaches are able to describe the video correctly. The third example is an interesting case where visually relevant descriptions, provided by most approaches, do not match the reference description, which focuses on an action happening in the background of the scene (``Someone sets down his young daughter then moves to a small wooden table.''). The last two rows contain partial and complete failures. In  one all approaches fail to recognize the person running away, only capturing the ``turning'' action which indeed happened before running. In the other one, all approaches fail to recognize that the woman interacts with the small object (phone).

Figure \ref{fig:qual_sequence} further compares all approaches on a sequence of 5 consecutive clips. We can make the following observations from these examples. Indeed Visual-Labels and S2VT produce longer and noisier descriptions. Sometimes, though, Visual-Labels is able to capture important details, such as ``Someone sips his drink'', which the other methods fail to recognize. Descriptions produced by Frame-Video-Concept Fusion and Temporal Attention are short and cleaner, also Temporal Attention tends to produce generally applicable sentences, \eg ``Someone looks at someone''. 

\subsection{\LSMDC summary}
We make the following conclusions from the presented results. (1) Visual-Labels and S2VT tend to generate longer, more diverse and novel descriptions, which leads to lower human rankings, due to higher error chances. (2) Temporal Attention and Frame-Video-Concept Fusion produce shorter and simpler descriptions and retrieve more from training data, and thus obtain better human rankings. (3) Frame-Video-Concept Fusion lands in the ``sweet spot'', in terms of sentence correctness and complexity. (4) Asking humans a different question, how helpful the descriptions would be for the blind, we see that the ranking changes, indicating that human judgments are sensitive to the question formulation. In the future we plan to experiment more with different evaluation criteria.

\section{Conclusion}
In this work we present the \LaScMoDeCh (\LSMDC), a novel dataset of movies with aligned descriptions sourced from movie scripts and ADs (audio descriptions for the blind, also referred to as DVS).
Altogether the dataset is based on \nLSMDCMovies movies and has \nLSMDCSentences sentences with aligned clips.
We compare AD with previously used script data and find that AD tends to be more correct and relevant to the movie than script sentences.

Our approach, \emph{\ApproachVisualLabels}, to automatic movie description trains visual classifiers and uses their scores as input to an LSTM. To handle the weak sentence annotations we rely on three ingredients. (1) We distinguish three semantic groups of labels (verbs, objects and places). (2) We train them separately, removing the noisy negatives. (3) We select only the most reliable classifiers. For sentence generation we show the benefits of exploring different LSTM architectures and learning configurations. %

To evaluate different approaches for movie description, we organized a challenge at ICCV 2015 where we evaluated submissions using automatic and human evaluation criteria. We found that the approaches S2VT and our \ApproachVisualLabels generate longer and more diverse description than the other submissions but are also more susceptible to content or grammatical errors. This consequently leads to worse human rankings with respect to correctness and grammar. In contrast, Frame-Video-Concept Fusion wins the challenge by predicting medium length sentences with intermediate diversity, which gets rated best in human evaluation for correctness, grammar, and relevance. When ranking sentences with respect to the criteria  ``helpful for the blind'', our  \ApproachVisualLabels is well received by human judges, likely because it includes important aspects provided by the strong visual labels. Overall all approaches have problems with the challenging long-tail distributions of our data. Additional training data cannot fully ameliorate this problem because a new movie might always contain novel parts. We expect new techniques, including relying on different modalities, see \eg \citep{hendricks16cvpr}, to overcome this challenge.

Our evaluation server will continue to be available for automatic evaluation and we will hold a second version of our challenge with human evaluation at ECCV 2016 as part of the \emph{Joint Workshop on Storytelling with Images and Videos and Large Scale Movie Description and Understanding Challenge}.
Our dataset has already been used beyond description, \eg for learning video-sentence embeddings or for movie question answering. Beyond our current challenge on single sentences, the dataset opens new possibilities to understand stories and plots across multiple sentences in an open domain scenario on large scale. %

\paragraph{Acknowledgements.}
Marcus Rohrbach was supported by a fellowship within the FITweltweit-Program of the German Academic Exchange Service (DAAD).

\bibliographystyle{plainnat}
\bibliography{biblioLong,rohrbach,related}
\end{document}